\documentclass[11pt]{article}

\usepackage[final]{acl}

\usepackage{times}
\usepackage{latexsym}
\usepackage[T1]{fontenc}
\usepackage[utf8]{inputenc}
\usepackage{microtype}

\usepackage{enumitem}
\usepackage{amsmath, amssymb}
\usepackage{booktabs}
\usepackage{graphicx}
\usepackage{xcolor}
\usepackage{longtable}
\usepackage{multirow}
\usepackage{placeins}
\usepackage{pifont}
\usepackage{soul}
\usepackage{float}
\usepackage{stfloats}  

\renewcommand{\topfraction}{0.92}
\renewcommand{\bottomfraction}{0.5}
\renewcommand{\textfraction}{0.07}
\renewcommand{\floatpagefraction}{0.85}
\newcommand{\cmark}{\ding{51}}
\newcommand{\xmark}{\ding{55}}
\newcommand{\EvaluationTarget}{\textsc{PACE}}
\newcommand{\BenchmarkName}{\textsc{PACEShop}}
\newcommand{\JudgeName}{\textsc{PACEJudge}}
\newcommand{\myparagraph}[1]{\smallskip\noindent\textbf{#1}}

\title{\BenchmarkName{}: Evaluating \underline{P}ersonalized, \underline{A}ctionable, \underline{C}ompositional, and \underline{E}vidence-grounded Shopping Assistants}

\author{\mdseries
Weimin Lyu\textsuperscript{\textnormal{1}}, 
Chen Luo\textsuperscript{\textnormal{1}}, 
Guangrui Li\textsuperscript{\textnormal{1}}, 
Yaochen Xie\textsuperscript{\textnormal{1}}, 
Dhineshkumar Ramasubbu\textsuperscript{\textnormal{1}}, \\ 
Arief Koesdwiady\textsuperscript{\textnormal{1}}, 
Wanqiu Long\textsuperscript{\textnormal{1}}, 
Hansu Gu\textsuperscript{\textnormal{1}}, 
Yutong Chen\textsuperscript{\textnormal{1}}, 
Zheshen Wang\textsuperscript{\textnormal{1}}, \\
Dakuo Wang\textsuperscript{\textnormal{2}}\thanks{Work done at Amazon.}, 
Yi Liu\textsuperscript{\textnormal{3}}\footnotemark[1] \\
\textsuperscript{1} Amazon 
\textsuperscript{2} Northeastern University 
\textsuperscript{3} Expedia
}

\vspace{-.2in}
\begin{document}
\maketitle

\begin{abstract}

Shopping assistants are shifting from ranked product lists toward structured decision support, where systems must synthesize shopper context, product evidence, and next-step guidance into a coherent recommendation experience. This changes the unit of evaluation: a fluent response can still fail by ignoring shopper context, contradicting itself across components, or leaving defects too vague to localize. Existing personalization, grounding, and LLM-as-a-judge benchmarks cover pieces of this problem, but they do not define a joint evaluation target for structured shopping-assistant responses. We formulate this missing evaluation target as \textbf{\EvaluationTarget}: \textbf{P}ersonalized, \textbf{A}ctionable, \textbf{C}ompositional, and \textbf{E}vidence-grounded evaluation. We instantiate \EvaluationTarget{} with two artifacts: \textbf{\BenchmarkName{}}, a benchmark dataset that makes the target measurable through 22,625 controlled records with structured personas, auditable evidence pools, GOOD/BAD labels, and gold defect family and location annotations; and \textbf{\JudgeName{}}, a training-free judging protocol that makes the target reportable through a structured output contract. Our experiments show that generic judges can recognize broad quality but fail to recover the diagnostic fields required for \EvaluationTarget{}; \BenchmarkName{} makes these failures verifiable, and \JudgeName{} improves persona-source, cross-component, grounding, and family/location closure without retraining, showing that realistic shopping-assistant evaluation requires a task-matched output contract rather than only a stronger backbone or scalar prompt.
\end{abstract}

\section{Introduction}

\begin{figure}[t]
\centering
\includegraphics[width=0.5\textwidth]{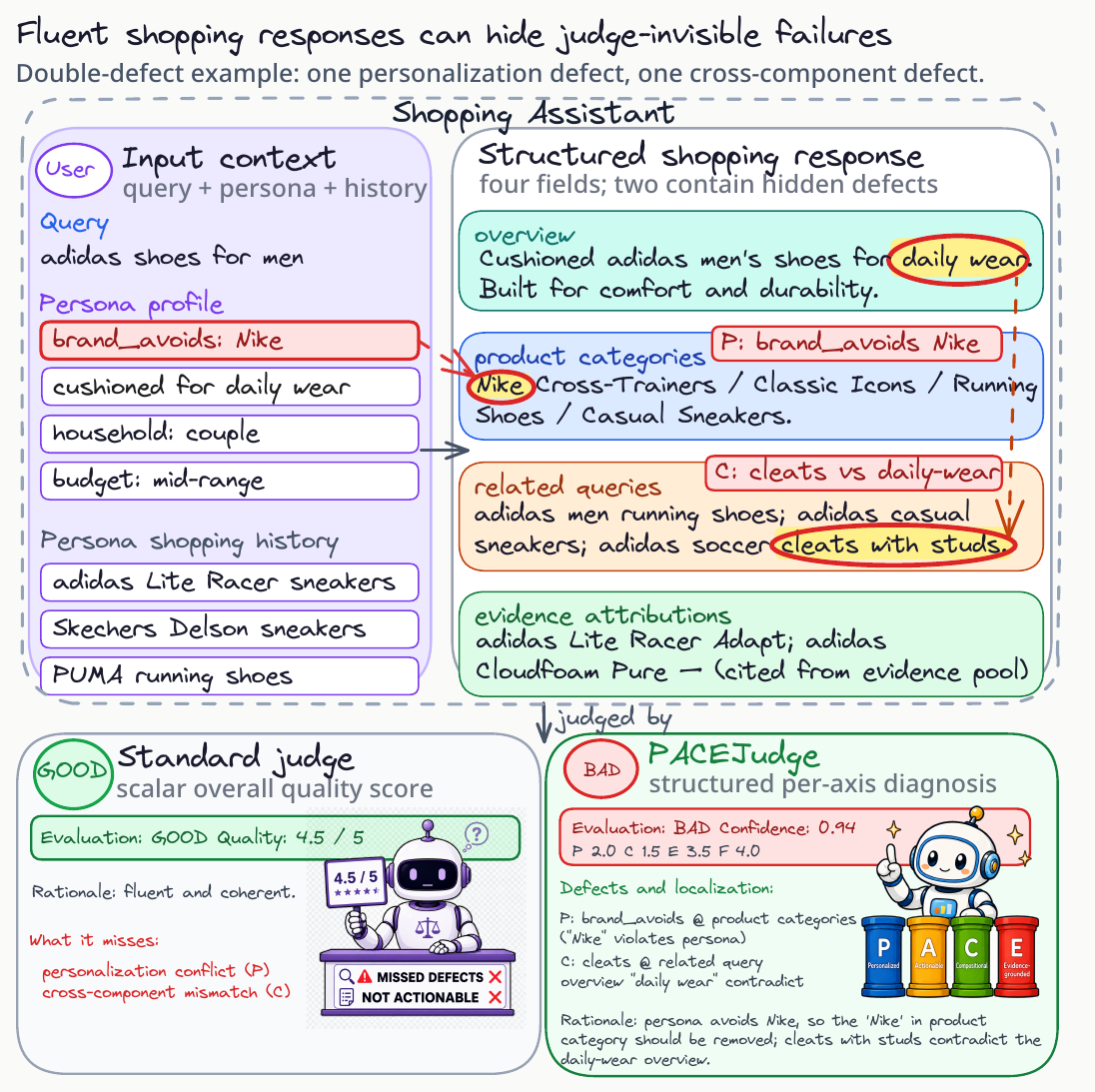}
\vspace{-.3in}
\caption{
\textbf{Generic judges can miss structured failures in shopping-assistant responses.}
For the same query, persona, and response, a standard judge returns GOOD, while \JudgeName{} returns BAD with defect family and field location. The example failures include a Nike preference conflict and a cross-field mismatch in related queries, exposing the need for \EvaluationTarget{} evaluation. 
}
\vspace{-.2in}
\label{fig:failure_teaser}
\end{figure}

Shopping assistants are moving beyond ranked product lists and single answers toward structured decision support: a response may summarize the shopper's need, organize product categories, suggest related query refinements, and cite evidence from a retrieved product pool \citep{miroyan2025searcharena, wang2025ecombench}. This changes the unit of evaluation. A response is no longer valid merely because one generated text is fluent or broadly relevant; it must be valid as a structured, persona-conditioned object. 
As Fig.~\ref{fig:failure_teaser} illustrates, a response can appear coherent while still failing in ways that a standard scalar judge may miss: it may recommend a brand the shopper avoids, contradict itself across response fields, or leave the defect unable to localize. Such failures are difficult to capture with an overall quality score, yet they are central to practical evaluation: a realistic judge must identify not only whether the response is wrong, but also what failed, where it failed, and which evidence supports the diagnosis.

These failures expose an under-specified evaluation target. Existing personalization benchmarks study whether systems adapt to user profiles, histories, preferences, or long-context behavioral signals \citep{salemi2023lamp,zhao2025personalens,zhao2025prefeval,jiang2025personamem,rahman2025likebench,hao2025etapp,dong2024personalizedjudge}. Grounding-focused evaluators test whether generated claims are supported by retrieved evidence \citep{saadfalcon2024ares}. Generic LLM-as-a-judge protocols score open-ended outputs with rubrics, scalar ratings, or pairwise preferences \citep{zheng2023judging, liu2023geval, thakur2024judging, li2024generationjudgment, chen2025distributional, feuer2025noise, enguehard2025lemaj}. These directions are complementary, but they do not define the joint target required for structured shopping-assistant responses: the judge must condition on shopper context, reason across response components, verify evidence support, and return a diagnosis that localizes the failure. The gap is therefore conceptual before it is empirical. A stronger judge backbone or a better scalar prompt cannot report defect fields that the evaluation task never asks it to produce.


We formulate this missing evaluation target as \textbf{\EvaluationTarget}: 
\textit{\underline{\textbf{P}}ersonalized},
\textit{\underline{\textbf{A}}ctionable},
\textit{\underline{\textbf{C}}ompositional}, and 
\textit{\underline{\textbf{E}}vidence-grounded}. 
\textit{\underline{\textbf{P}}ersonalized} means that validity depends on the shopper context; 
\textit{\underline{\textbf{A}}ctionable} means that the judge must identify what failed and where; 
\textit{\underline{\textbf{C}}ompositional} means that the response must be evaluated as a structured object rather than isolated text; 
\textit{\underline{\textbf{E}}vidence-grounded} means that claims must be supported by product evidence and persona history. 
Together, \EvaluationTarget{} clarifies what current evaluation lacks: records that make such failures observable, and a judging protocol that can report them. Tab.~\ref{tab:pace_operationalization} summarizes this design principle: each \EvaluationTarget{} target determines what the benchmark must contain, what the judge must output, and what scenario-level metric can test it.


To make \EvaluationTarget{} measurable, we introduce \BenchmarkName{}, a verifiable benchmark dataset for structured shopping-assistant evaluation. Existing datasets provide valuable personalization or judging signals, but they do not jointly expose persona-dependent validity, cross-component consistency, evidence support, and localizable defects in one controlled setting. \BenchmarkName{} fills this gap by pairing each shopping query with a structured persona, a candidate response, an auditable evidence pool, and a GOOD/BAD label; BAD records additionally include gold defect family and defect location annotations. This design turns the abstract \EvaluationTarget{} formulation into measurable evaluation cases.

To make \EvaluationTarget{} reportable, we propose \JudgeName{}, a training-free judging protocol. Existing scalar judges can estimate overall quality, but they do not require the judge to name the violated dimension, localize the broken response field, or ground the diagnosis in evidence. \JudgeName{} fills this gap through a judging protocol that evaluates \EvaluationTarget{} with a a structured output schema. The protocol is backbone-agnostic, allowing us to separate what comes from model capability from what comes from asking the judge to produce the right diagnostic fields.

Across multiple LLM backbones and judging configurations, we compare broad GOOD/BAD discrimination with scenario-level tests that require persona-source diagnosis, cross-component localization, evidence-grounding checks, and actionable defect labels. The results show that general discrimination is not the main bottleneck: scalar and generic judges can often recognize that a response is broadly good or bad. The gap appears when evaluation requires the diagnostic fields needed for debugging. On the same backbones, \JudgeName{} improves these contract-dependent metrics without retraining, supporting our central claim: realistic shopping-assistant evaluation requires diagnostic closure over persona, components, evidence, and defect location, not only broad quality discrimination. Our contributions are threefold:

\begin{enumerate}
    \vspace{-.08in}
    \item We identify a problem-setting gap in existing evaluation for structured shopping assistants, and formulate the missing target as \EvaluationTarget{}.
    \vspace{-.08in}

    \item We introduce \BenchmarkName{}, a benchmark dataset that makes \EvaluationTarget{} measurable through controlled persona-, component-, and evidence-grounded failure cases.
    \vspace{-.08in}
    
    \item We propose \JudgeName{}, a training-free judging protocol that makes \EvaluationTarget{} reportable and improves diagnostic metrics beyond scalar quality scores.
\vspace{-.1in}
\end{enumerate}

\begin{table*}[!t]
\vspace{-.2in}
\centering
\small
\resizebox{\textwidth}{!}{%
\begin{tabular}{l l l cccc l}
\toprule
\textbf{Benchmark / Resource} 
& \textbf{Domain} 
& \textbf{Size} 
& \textbf{P} 
& \textbf{A} 
& \textbf{C} 
& \textbf{E} 
& \textbf{Evaluation signal} \\
\midrule

LaMP \cite{salemi2023lamp} 
& General personalization 
& 7 tasks 
& \cmark & \xmark & \xmark & \xmark 
& Task metrics for user-conditioned generation \\

PersonaLens \cite{zhao2025personalens} 
& Task-oriented assistants 
& 1.5K profiles, 111 tasks 
& \cmark & \xmark & \xmark & \xmark 
& LLM user and judge agents \\

PrefEval \cite{zhao2025prefeval} 
& Long-context preferences 
& 3K pref-query pairs 
& \cmark & \xmark & \xmark & \xmark 
& Preference-following generation/classification \\

PersonaMem \cite{jiang2025personamem} 
& Dynamic personalization 
& 180 histories, $\leq 60$ sessions 
& \cmark & \xmark & \xmark & \xmark 
& Response selection over evolving user history \\

LikeBench \cite{rahman2025likebench} 
& Likability personalization 
& not publicly stated 
& \cmark & \xmark & \xmark & \xmark 
& Simulated-user likability diagnostics \\

EtaPP \cite{hao2025etapp} 
& Personalized tool use 
& 800 test cases 
& \cmark & \xmark & \xmark & \xmark 
& Key-point judging for tool-use trajectories \\

OPeRA \cite{wang2025opera} 
& Shopping behavior simulation 
& 692 sessions, 28.9K obs.-action pairs 
& \cmark & \xmark & \xmark & \xmark 
& Real-user observations, actions, and rationales \\

ECom-Bench \cite{wang2025ecombench} 
& E-commerce support 
& 53 tasks, hundreds of personas 
& \cmark & \xmark & \xmark & \xmark 
& Task-success trajectories \\

ESCI~\cite{reddy2022shopping} 
& Product search relevance 
& 97.3K US queries 
& \xmark & \xmark & \xmark & \xmark 
& Human product-relevance labels \\

Search Arena \cite{miroyan2025searcharena} 
& Search-augmented chat 
& 24K interactions, 12K votes 
& \xmark & \xmark & \xmark & \xmark 
& Pairwise human preferences \\

\midrule

\textbf{\BenchmarkName{} (ours)} 
& \textbf{Personalized shopping} 
& \textbf{1,132 $q$ $\times$ 1.2K $p$ = 22.6K records} 
& \textbf{\cmark} 
& \textbf{\cmark} 
& \textbf{\cmark} 
& \textbf{\cmark} 
& \textbf{Gold GOOD/BAD, defect family, and defect location labels} \\

\bottomrule
\end{tabular}%
}
\caption{
Existing benchmark datasets against the four \EvaluationTarget{} targets.
\BenchmarkName{} jointly covers all targets through structured personas, multi-component responses, auditable evidence pools, and gold defect family/location labels.
}
\vspace{-.2in}
\label{tab:related_work_pace}
\end{table*}

\section{Related Work}

\paragraph{Personalization, shopping, and search benchmarks.}
Personalized LLM benchmarks evaluate user-conditioned generation, preferences, memory, likability, and tool use \citep{salemi2023lamp, zhao2025personalens, zhao2025prefeval, jiang2025personamem, rahman2025likebench, hao2025etapp, dong2024personalizedjudge}. 
Commerce-oriented resources such as ECom-Bench~\citep{wang2025ecombench}, Search Arena~\citep{miroyan2025searcharena}, and OPeRA~\citep{wang2025opera} add realistic e-commerce support, search-augmented chat, or online-shopping behavior signals. 
However, they do not provide the joint evaluation unit needed and cannot directly test whether a response satisfies \EvaluationTarget{} targets: personalization, actionability, compositional consistency, and evidence grounding. 
Tab.~\ref{tab:related_work_pace} summarizes this benchmark-side gap.


\paragraph{LLM-as-a-judge and grounded evaluation.}
Generic LLM-as-a-judge methods such as G-Eval~\citep{liu2023geval} and MT-Bench~\citep{zheng2023judging} established rubric-based, scalar, and pairwise judging for open-ended model outputs. 
Subsequent work studies judge reliability, alignment, bias, distributional validity, and domain-specific judging protocols \citep{thakur2024judging, li2024generationjudgment, chen2025distributional, feuer2025noise, enguehard2025lemaj}. 
Grounding-focused evaluators such as ARES~\citep{saadfalcon2024ares} test whether generated claims are supported by retrieved evidence. 
These methods are complementary, but their evaluation contracts are not designed for structured shopping-assistant diagnosis: they do not jointly require persona conditioning, cross-component reasoning, evidence support, defect-family prediction, and field-level localization. 
\JudgeName{} addresses this protocol-side gap by making the four \EvaluationTarget{} targets reportable through structured diagnostic outputs.

\section{\BenchmarkName{} Dataset}
\label{sec:paceshop_dataset}

\begin{table*}[t]
\vspace{-.2in}
\centering
\small
\begin{tabular}{p{0.17\linewidth} p{0.27\linewidth} p{0.27\linewidth} p{0.21\linewidth}}
\toprule
\textbf{\EvaluationTarget{} target}
& \textbf{\BenchmarkName{}}
& \textbf{\JudgeName{} protocol}
& \textbf{Evaluation scenario} \\
\midrule

\textbf{Personalized} \newline
Validity depends on shopper context.
&
Pairs each query with structured personas; persona-swap records hold the query and response fixed while changing shopper context.
&
Requires persona-conditioned judging; reports persona-alignment score, rationale, and persona-conflict diagnosis.
&
\textbf{S1 Persona}: tests whether the judge detects when a swapped persona invalidates the same response. \\

\midrule

\textbf{Actionable} \newline
Evaluation must say what failed and where.
&
Provides BAD records with gold defect family and gold defect location from a 7-family $\times$ 4-field taxonomy.
&
Requires structured fault diagnosis; reports defect family, defect location, confidence, rationale, and supporting evidence IDs.
&
\textbf{S4 Actionable}: tests exact defect-family prediction and response-field localization. \\

\midrule

\textbf{Compositional} \newline
Responses must be consistent as structured objects.
&
Represents responses with multiple heterogeneous components; controlled BAD variants inject cross-component defects.
&
Requires judging the response as a structured object; reports compositional-consistency score and localized response component.
&
\textbf{S2 Compositional}: tests cross-component defect detection and localization. \\

\midrule

\textbf{Evidence-grounded} \newline
Claims must be supported by product and persona evidence.
&
Includes an auditable evidence pool; controlled BAD variants inject invalid product evidence and invented persona-history claims.
&
Requires grounding against product evidence and persona history; reports evidence-grounding score and evidence IDs constrained to the evidence pool.
&
\textbf{S3 Grounding}: tests product-evidence grounding, persona-history grounding, and hallucinated-ID control. \\

\bottomrule
\end{tabular}
\caption{
\textbf{Making \EvaluationTarget{} measurable and reportable.}
Each \EvaluationTarget{} target maps to what \BenchmarkName{} makes measurable, how the \JudgeName{} protocol makes it reportable, and which evaluation scenario tests the corresponding failure mode.
}
\vspace{-.2in}
\label{tab:pace_operationalization}
\end{table*}

\BenchmarkName{} is designed to make \EvaluationTarget{} measurable. 
The goal is to construct controlled and verifiable records in which personalization, actionability, compositional consistency, and evidence grounding can be observed and audited. 
Tab.~\ref{tab:pace_operationalization} previews how each \EvaluationTarget{} target is turned into measurable benchmark controls in \BenchmarkName{}.
The benchmark construction follows a valid-response-first, controlled-defect-second design: we first create and validate structured GOOD responses, then derive BAD variants through controlled edits with known defect family and location labels. 

\subsection{Benchmark Task}
\label{sec:benchmark_task}

\paragraph{Scope.}
\BenchmarkName{} focuses on the generated response layer of shopping assistants rather than the full ranking interface or end-to-end shopping session. 
This scope isolates the judging problem: given a shopping assistant user inputs and responses, can a judge determine whether the response is valid and diagnose any failure?

\paragraph{Task formulation.}
Each \BenchmarkName{} record defines a pointwise judging task over a tuple
\[
(q, p, y, E, \ell),
\]
where $q$ is a shopping query, $p$ is a structured persona, $y$ is a candidate shopping-assistant response, $E$ is an auditable evidence pool, and $\ell \in \{\textsc{GOOD}, \textsc{BAD}\}$ is the validity label. 
The judge receives $(q,p,y,E)$ and predicts
\[
\hat{\ell}=J(q,p,y,E), \qquad \hat{\ell}\in\{\textsc{GOOD},\textsc{BAD}\}.
\]
For BAD responses, the judge should additionally diagnose the failure by predicting a defect family $d \in \mathcal{D}$ and a defect location $r \in \mathcal{R}$. 
Thus, the benchmark measures not only whether a judge can separate GOOD from BAD responses, but also whether it can identify what failed and where the failure occurs.

\begin{figure*}[t]
\vspace{-.2in}
\centering
\includegraphics[width=\textwidth]{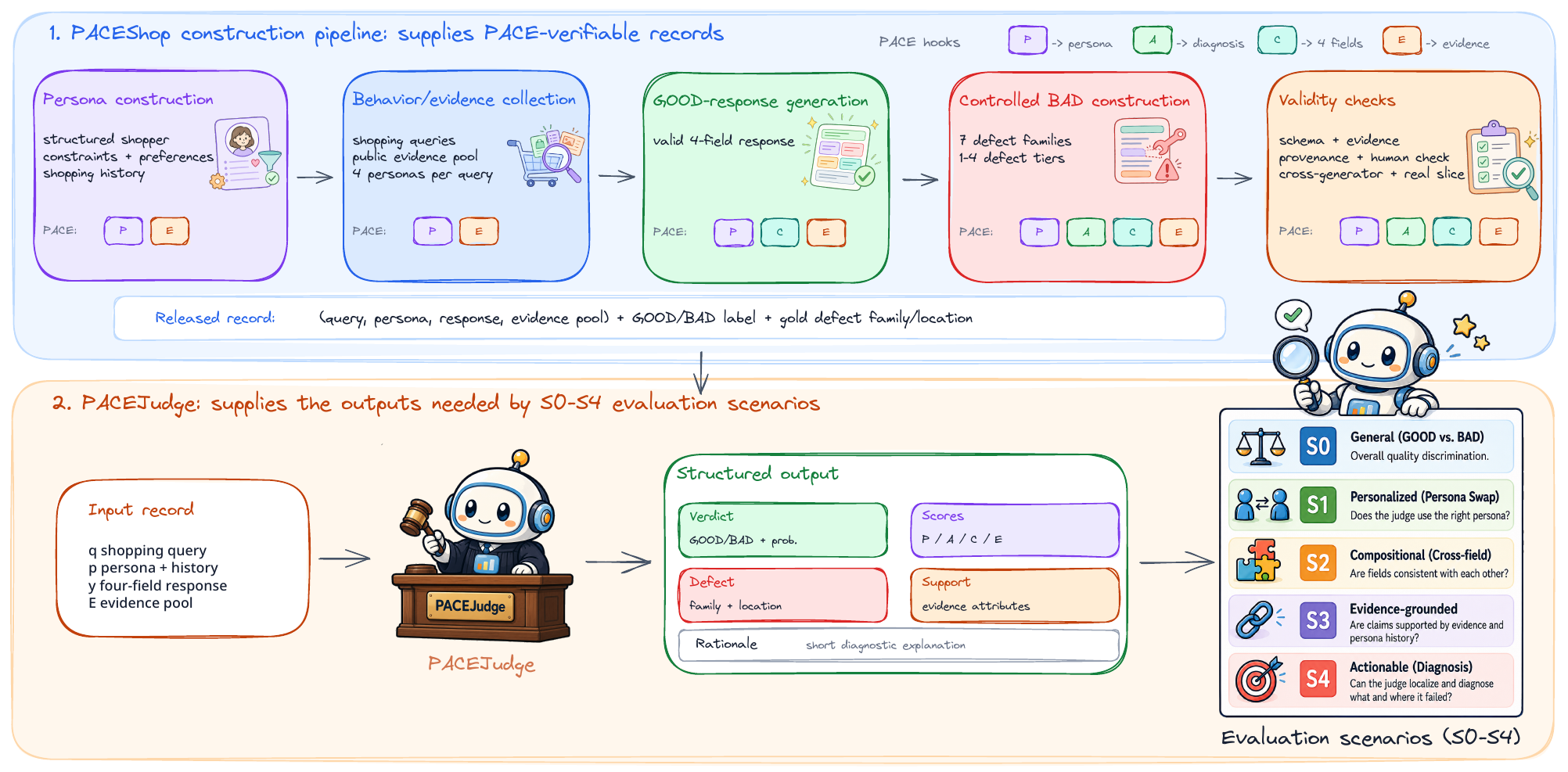}
\vspace{-.2in}
\caption{\BenchmarkName{} benchmark dataset makes \EvaluationTarget{} measurable; and \JudgeName{} evaluation protocal makes \EvaluationTarget{} reportable. 
\textbf{Top:} \BenchmarkName{} data construction carries persona, shopping behavior, query intent, and evidence through assignment, GOOD-response generation, controlled BAD construction, and validity checks. \textbf{Middle:} the resulting records make PACE properties measurable through S1--S4, with S0 as a reference GOOD/BAD check. \textbf{Bottom:} \JudgeName{} turns each record $(q,p,y,E)$ into a structured output whose fields support scenario-aligned evaluation.}
\vspace{-.2in}
\label{fig:generation_task_flow}
\end{figure*}

\subsection{\BenchmarkName{} Construction Pipeline}
\label{sec:benchmark_construction}

The PACEShop construction pipeline makes \EvaluationTarget{} failures observable and auditable. It isolates shopper-context effects through persona assignment, makes grounding checkable through evidence-backed query retention, uses validated GOOD responses as verified base cases, and derives controlled BAD variants with gold defect family and location labels. Full construction details, prompts, schema checks, and examples are provided in Appx.~\ref{app:paceshop_construction}.

\paragraph{Persona and evidence setup.}
We construct 1,200 personas as structured shopping contexts rather than free-form user descriptions. 
Each persona specifies stable shopper information, such as household type, budget, quality sensitivity, urgency, constraints, preferences, and brand affinities, together with concrete shopping behavior in the form of purchase history and recent searches. 
We pair these personas with evidence-backed shopping queries drawn from a public query and product-evidence pool. 
This setup lets the same query and evidence pool support different valid responses under different shopper contexts, making personalization and persona-history grounding testable.

\paragraph{Query--persona assignment.}
We begin with 97,227 normalized US shopping queries, 1,200 structured personas, and 4,460 public evidence records across 11 shopping domains.\footnote{Coverage is audited at the source level: queries are tagged by retail vertical and shopping mission, personas by shopper attributes and behavior-history fields, and evidence records by domain and provenance. Appx.~\ref{app:source_coverage} reports the detailed coverage breakdowns and representative examples.}
We retain 1,132 queries with sufficient evidence support and pair each retained query with distinct personas, yielding 4,525 validated query--persona pairs. 
The repeated-measures design holds the query and evidence pool fixed while varying shopper context, allowing persona-dependent failures to be isolated from changes in shopping intent.


\paragraph{GOOD-response generation.}
For each query--persona pair, a strong LLM generator produces one structured GOOD candidate under a constrained response contract. 
We do not treat these generations as valid merely because they come from a strong model: each candidate must pass deterministic normalization and validation, including schema, cardinality, attribution-ID, and parsing checks. 
Accepted GOOD responses therefore serve as contract-valid benchmark anchors for controlled defect construction.

\paragraph{Controlled BAD construction.}
From each validated GOOD response, we construct four BAD variants containing one, two, three, and four defects. 
Defects are drawn from seven families covering persona conflicts, intent drift, cross-component mismatch, redundancy, evidence mismatch, over-personalized hallucination, and unsupported claims. 
Because each BAD record is created by a controlled edit, its gold defect family and field location are known by construction.

\paragraph{Validity checks.}
\BenchmarkName{} is designed to make \EvaluationTarget{} failures measurable and verifiable, not to rely on synthetic generation as ground truth. 
Each released record carries an audit trail: deterministic validators check schema compliance and evidence-ID consistency, validated GOOD responses are used as base cases before any defect injection, controlled edits make BAD labels traceable to known defect families and response fields, and evidence-provenance flags expose the strength of the grounding signal. 
Together, these layers address the main dataset-quality risks: synthetic-on-synthetic circularity, thin grounding signal, persona-template artifacts, rigid schema effects, and limited transfer beyond one model family. 
Appx.~\ref{app:validity_stack} details the five verification layers; Appx.~\ref{app:defect_taxonomy} defines the defect taxonomy; Appx.~\ref{app:benchmark_prompt} gives GOOD/BAD record generation procedures.

\myparagraph{Dataset statistics.}
PACEShop contains 22,625 records: 4,525 GOOD records and 18,100 BAD records.
BAD records are evenly distributed across single-, double-, triple-, and quad-defect tiers, and every BAD record carries gold defect family and location annotations.
Tab.~\ref{tab:dataset_summary} in Appx.~\ref{app:paceshop_construction} summarizes the benchmark statistics, evidence coverage, defect structure, and validation assets.

\section{\JudgeName{} Evaluation Protocol}
\label{sec:percojudge}

\JudgeName{} is a training-free pointwise judging protocol that makes \EvaluationTarget{} reportable. 
In Tab.~\ref{tab:pace_operationalization}, \BenchmarkName{} makes the \EvaluationTarget{} targets measurable through controlled records, while \JudgeName{} specifies how a judge should report the diagnosis. 
Given a \BenchmarkName{} record, the protocol asks the judge to evaluate the candidate response as a structured, persona-conditioned, evidence-grounded object rather than as a single fluent text. 
This addresses the central limitation of scalar judging: an overall quality score does not identify whether the failure comes from shopper-context mismatch, cross-component inconsistency, unsupported evidence, or a defect that cannot be localized.

\subsection{Protocol Overview and Output Schema}


Given a \BenchmarkName{} record $(q,p,y,E)$, \JudgeName{} returns
\[
J(q,p,y,E) \rightarrow
(\hat{\ell}, \hat{p}_{\mathrm{bad}}, s_{\mathrm{P}}, a_{\mathrm{A}}, s_{\mathrm{C}}, s_{\mathrm{E}}, s_{\mathrm{F}}),
\]
where $\hat{\ell}\in\{\textsc{GOOD},\textsc{BAD}\}$ is the verdict, $\hat{p}_{\mathrm{bad}}\in[0,1]$ is the BAD probability, $s_{\mathrm{P}},s_{\mathrm{C}},s_{\mathrm{E}}\in[1,5]$ are the Personalized, Compositional, and Evidence-grounded scores, and $s_{\mathrm{F}}\in[1,5]$ is an auxiliary format/safety score. 
The Actionable target is represented as a diagnostic object rather than a scalar:
\[
a_{\mathrm{A}}=(\hat{d},\hat{r},\hat{S},\hat{c},\hat{z}),
\]
where $\hat{d}$ is the predicted defect family, $\hat{r}$ is the predicted defect location, $\hat{S}$ is the set of supporting evidence IDs, $\hat{c}$ is confidence, and $\hat{z}$ is a short rationale.
The diagnostic fields follow the benchmark schema:
\[
\hat d \in \mathcal{D}\cup\{\textsc{None}\}, \quad
\hat r \in \mathcal{R}\cup\{\textsc{None}\}, \quad
\hat S \subseteq E,
\]
where $\mathcal{D}$ is the seven-family defect taxonomy and $\mathcal{R}$ is the response-field set
(\textit{overview}, \textit{categories}, \textit{related queries}, and \textit{attributions}).
For GOOD verdicts, the defect family and location should be \textsc{None}; for BAD verdicts, the judge identifies the primary defect family and response field. Detailed protocol and examples can be found in Appx.~\ref{app:judge_prompt}.


\begin{table*}[t]
\vspace{-.2in}
\centering
\small
\setlength{\tabcolsep}{5pt}
\renewcommand{\arraystretch}{1.10}
\resizebox{\textwidth}{!}{%
\begin{tabular}{l cc cc c cc c c}
\toprule
& \textbf{S0} & \textbf{S1 P} & \multicolumn{2}{c}{\textbf{S2 C}} & \textbf{S3 E} & \multicolumn{2}{c}{\textbf{S4 A}} & \textbf{False-fire} & \textbf{PACE} \\
\cmidrule(lr){4-5} \cmidrule(lr){7-8}
\textbf{Method} & Gen. & P-source & detect & loc. & grounding & family & loc. & $\downarrow$ & avg. \\
\midrule
\multicolumn{10}{l}{\emph{Rubric ablations of the same PACE output contract (7-backbone mean)}} \\
G-Eval \cite{liu2023geval} & 0.73 & 0.17 & 0.75 & 0.70 & 0.40 & 0.48 & 0.68 & 0.52 & 0.45 \\
Scalar (MT-Bench-style) \cite{zheng2023judging} & 0.73 & 0.12 & 0.77 & 0.60 & 0.50 & 0.49 & 0.68 & 0.51 & 0.44 \\
Persona-blind & 0.74 & -- & 0.79 & 0.64 & 0.49 & 0.49 & 0.69 & 0.47 & -- \\
Persona-aware & 0.74 & 0.20 & 0.79 & 0.64 & 0.60 & 0.50 & 0.68 & 0.37 & 0.51 \\
\textbf{\JudgeName{} (ours)} & \textbf{0.74} & \textbf{0.32} & \textbf{0.93} & \textbf{0.72} & \textbf{0.62} & \textbf{0.59} & \textbf{0.78} & 0.47 & \textbf{0.56} \\
\midrule
\multicolumn{10}{l}{\emph{Honest-adapted single-target baselines (cells shown only where the native judging protocol supports the metric)}} \\
ARES-style \cite{saadfalcon2024ares} & 0.69 & -- & -- & -- & 0.39 & -- & -- & 0.26 & -- \\
PersonaLens-style \cite{zhao2025personalens} & 0.65 & 0.18 & -- & -- & -- & -- & -- & \textbf{0.07} & -- \\
EtaPP-style \cite{hao2025etapp} & 0.71 & -- & -- & -- & -- & 0.50 & 0.74 & 0.43 & -- \\
\bottomrule
\end{tabular}%
}
\caption{
\textbf{Main \EvaluationTarget{}-closure results averaged over seven LLM backbones.}
\textit{S0} reports reference GOOD/BAD balanced accuracy, while \textit{S1--S4} evaluate \EvaluationTarget{} diagnostic closure.
\JudgeName{} achieves the strongest overall performance among multi-target rubric and the single-target methods, showing that the main gap is not broad GOOD/BAD discrimination but recovering the diagnostic fields needed for \EvaluationTarget{} evaluation. 
\textit{False-fire} measures defect alarms on GOOD records, and \textit{\EvaluationTarget{} avg.} combines closure metrics with GOOD-record specificity. 
``--'' marks unsupported cells under a method's native judging protocol. 
Full per-backbone results (Tab.~\ref{tab:sota_failure_main_full}) and metric definitions are provided in Appx.~\ref{app:main_table_metrics}.
}
\label{tab:sota_failure_main}
\end{table*}

\subsection{Coverage of the \EvaluationTarget{} Targets}
\label{sec:judge_coverage}

The protocol is designed to cover all \EvaluationTarget{} targets.
For \textbf{P}, the persona score $s_{\mathrm{P}}$ makes shopper context part of the verdict. 
For \textbf{A}, the diagnostic object $a_{\mathrm{A}}$ requires the judge to report what failed, where it failed, and what evidence supports the diagnosis. 
For \textbf{C}, the compositional score $s_{\mathrm{C}}$ and defect-location field $\hat{r}$ require the judge to evaluate the response as a structured object rather than isolated text. 
For \textbf{E}, the evidence score $s_{\mathrm{E}}$ and evidence-ID constraint $\hat{S}\subseteq E$ require grounding in both the product evidence pool and persona history.

This design makes \JudgeName{} evaluable beyond GOOD/BAD accuracy. 
A scalar judge may call a response ``mostly helpful,'' but it has no required field for the violated target, the broken response component, the defect family, or supporting evidence. 
\JudgeName{} makes these diagnostic fields explicit, enabling the scenario-level metrics in Sec.~\ref{sec:experiments}.
A detailed per-target walkthrough of how \JudgeName{} reports the corresponding \EvaluationTarget{} target, together with the matching scenario-level metric, is provided in Appx.~\ref{app:judge_evaluation}.

\section{Experiments}
\label{sec:experiments}


\BenchmarkName{} provides controlled records where persona, compositional, grounding, and actionable-diagnosis failures are measurable; \JudgeName{} specifies the diagnostic fields a judge should return. 
We therefore ask a simple question: when evaluated on the same records and backbones, do scalar or generic judge protocols recover the diagnostic fields needed for \EvaluationTarget{} evaluation, or is an explicit structured protocol necessary?

\subsection{Experimental Setup}
We evaluate judge protocols on \BenchmarkName{} records using seven LLM backbones (Opus 4.7, Sonnet 4.6, Sonnet 4.5, Haiku 4.5, Qwen3 32B, GPT-OSS 20B, GPT-OSS 120B).
We compare \JudgeName{} with two baseline families on \BenchmarkName{} records. 
The first family consists of \emph{multi-target rubric ablations} that share the \JudgeName{} output schema but vary the judging instruction: G-Eval-style scalar judging~\citep{liu2023geval}, MT-Bench-style scalar judging~\citep{zheng2023judging}, persona-blind judging, and a persona-aware generic rubric. 
The second family consists of \emph{single-target adapted baselines} whose native protocols cover only partial \EvaluationTarget{} dimensions: ARES-style grounding~\citep{saadfalcon2024ares}, PersonaLens-style personalization~\citep{zhao2025personalens}, and EtaPP-style actionable diagnosis~\citep{hao2025etapp}.
Full backbone, prompt, and native-support details are in Appx.~\ref{app:experiments}.


\subsection{Evaluation Scenarios}
\label{sec:scenarios}

Building on the target-to-scenario mapping in Tab.~\ref{tab:pace_operationalization}, we report one reference scenario and four \EvaluationTarget{} closure tests. 
S0 measures broad GOOD/BAD discrimination and is included only to show whether records are broadly judgeable. 
S1--S4 test persona-source diagnosis, cross-component detection and localization, grounding over product evidence and persona history, and actionable defect-family/field-location prediction, respectively. 
Tab.~\ref{tab:scenario_metric_map} links these scenarios to metric columns; full definitions are provided in Appx.~\ref{app:eval_scenario}.

\begin{table*}[t]
\vspace{-.2in}
\centering
\scriptsize
\setlength{\tabcolsep}{4pt}
\renewcommand{\arraystretch}{1.10}
\resizebox{\textwidth}{!}{%
\begin{tabular}{ll cccccc}
\toprule
\textbf{Method} & \textbf{Comparison level} & \textbf{S0} & \textbf{S1$_{\text{P}}$} & \textbf{S2$_{\text{C}}$} & \textbf{S3$_{\text{E}}$} & \textbf{S4$_{\text{A}}$} & \textbf{PACE} \\
 & & ref. acc. & PJC & J-strict & J-harmonic & J-strict & avg.\ S1--S4 \\
\midrule
\multicolumn{8}{l}{\emph{Single-target adapted baselines on Sonnet 4.6 (each native judging protocol covers exactly one \EvaluationTarget{} target)}} \\
ARES-style \cite{saadfalcon2024ares} & native target only & 0.69 & -- & -- & 0.30 & -- & -- \\
PersonaLens-style \cite{zhao2025personalens} & native target only & 0.72 & 0.01 & -- & -- & -- & -- \\
EtaPP-style \cite{hao2025etapp} & native target only & 0.81 & -- & -- & -- & 0.00 & -- \\
\midrule
\multicolumn{8}{l}{\emph{Frontier control: 7-backbone self-consistency majority vote on the same rubric}} \\
G-Eval \cite{liu2023geval} + 7-vote SC & majority vote & 0.91 & 0.02 & 0.51 & 0.30 & 0.28 & 0.28 \\
Persona-aware + 7-vote SC & majority vote & 0.81 & 0.08 & 0.59 & 0.32 & 0.25 & 0.31 \\
\textbf{\JudgeName{}} + 7-vote SC & majority vote & 0.92 & 0.07 & 0.74 & 0.82 & 0.37 & 0.50 \\
\midrule
\multicolumn{8}{l}{\emph{PACE-compatible rubrics on the shared JSON contract (7-backbone mean)}} \\
G-Eval \cite{liu2023geval} & 7-backbone mean & 0.74 & 0.03 & 0.28 & 0.05 & 0.05 & 0.10 \\
Scalar (MT-Bench-style) \cite{zheng2023judging} & 7-backbone mean & 0.75 & 0.02 & 0.34 & 0.04 & 0.09 & 0.12 \\
Persona-blind & 7-backbone mean & 0.76 & -- & 0.33 & 0.05 & 0.09 & -- \\
Persona-aware & 7-backbone mean & 0.75 & 0.08 & 0.33 & 0.10 & 0.05 & 0.14 \\
\textbf{\JudgeName{}} & 7-backbone mean & 0.79 & 0.07 & 0.46 & 0.20 & 0.11 & 0.21 \\
\midrule
\textbf{Lift: \JudgeName{} mean vs.\ best PACE-compatible baseline mean} & per-target lift & $+0.04$ & $-0.00$ & $+0.13$ & $+0.11$ & $+0.02$ & $+0.07$ \\
\textbf{Lift: \JudgeName{} on Sonnet 4.6 vs.\ best Sonnet 4.6 baseline} & per-target lift & $+0.10$ & $+0.11$ & $+0.24$ & $+0.19$ & $+0.32$ & $+0.22$ \\
\bottomrule
\end{tabular}%
}
\caption{
\textbf{Strict paired-counterfactual \EvaluationTarget{}-closure results.}
Each S1--S4 score is a joint metric that requires more than detecting BAD examples: the judge must also satisfy target-specific conditions.
These gates prevent an always-BAD judge from receiving high credit through over-flagging. 
The last two rows summarize the lift from \JudgeName{} over the strongest comparable baseline, showing consistent gains under strict closure metrics at both the seven-backbone mean and best-backbone levels. 
Metric definitions are in Appx.~\ref{app:main_table_metrics}; native-protocol support and leakage audits for single-target adapted baselines are in Tabs.~\ref{tab:baseline_protocol_target_map} and~\ref{tab:evaluation_target_leakage_audit}.
}
\label{tab:sota_failure_main_v4}
\end{table*}

\subsection{Main Results}
\label{sec:main_results}

Tab.~\ref{tab:sota_failure_main} tells a simple story: broad GOOD/BAD judging is not where the main difficulty lies. 
Most judges/protocols can often decide whether a response is generally valid or flawed, but they break down when the metric asks for the fields needed to debug the failure: which \EvaluationTarget{} target was violated, which response field failed, and whether the diagnosis is grounded in the provided evidence. 
\JudgeName{} is strongest on this diagnostic closure setting among multi-target rubric ablations, while single-target adapted baselines leave most columns unsupported because their native protocols cover only one part of \EvaluationTarget{}. 
Tab.~\ref{tab:sota_failure_main_v4} then asks a stricter question: does the judge simply flag more responses as BAD? 
Under paired-counterfactual gates that require GOOD-record specificity, correct target/family/location, axis consistency, and evidence-ID containment, \JudgeName{} remains strongest. 
The improvement therefore reflects better diagnostic closure, not merely more aggressive defect detection.

\paragraph{Finding 1: broad accuracy and stronger backbones do not solve \EvaluationTarget{} closure.}
S0 in Tab.~\ref{tab:sota_failure_main} shows that scalar and generic judges can perform similarly to \JudgeName{} on broad GOOD/BAD discrimination. 
However, this binary signal does not translate into \EvaluationTarget{} closure: a judge may mark a response BAD while still failing to identify whether the failure comes from shopper-context mismatch, cross-component inconsistency, unsupported evidence, or a missing actionable diagnosis. 
Tab.~\ref{tab:sota_failure_main_v4} shows the same pattern in the stricter audit: even frontier self-consistency improves reference accuracy, but generic rubrics still trail \JudgeName{} on the joint S1--S4 closure metrics. 
This shows that stronger inference alone cannot recover fields that the protocol does not ask the judge to produce.

\begin{figure}[t]
\centering
\includegraphics[width=\columnwidth]{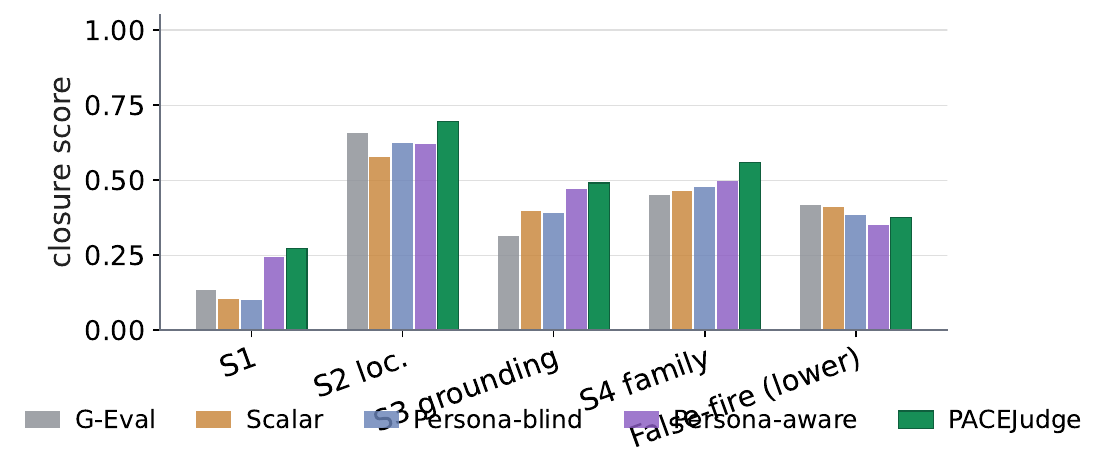}
\vspace{-.3in}
\caption{
\EvaluationTarget{}-closure scores averaged over seven backbones. 
\JudgeName{} is strongest on the S1--S4 metrics that require persona-source diagnosis, cross-component localization, grounding control, and actionable defect localization.
}
\label{fig:sota_failure_gap}
\end{figure}

\paragraph{Finding 2: \JudgeName{} improves the diagnostic fields required by \EvaluationTarget{}.}
Across the seven-backbone in Tab.~\ref{tab:sota_failure_main}, \JudgeName{} achieves the strongest overall \EvaluationTarget{} average among multi-target rubric ablations. 
The gain is diagnostic rather than scalar: \textbf{P} requires identifying persona conflict as the source of invalidity, \textbf{C} requires localizing the failed response component, \textbf{E} requires grounding against both product evidence and persona history, and \textbf{A} requires defect-family and field-location prediction. 
Fig.~\ref{fig:sota_failure_gap} summarizes this S1--S4 closure pattern, omitting S0 because the central claim is not broad classification but task-aligned diagnosis.

\paragraph{Finding 3: protocol design helps, but backbone capability sets the ceiling.}
The per-backbone breakdowns in Appx.~\ref{app:full_main_table}, \ref{app:persona_swap_full}, and \ref{app:component_consistency_full} show a common pattern: \JudgeName{} is most effective when the backbone can follow a structured contract, calibrate GOOD-record specificity, and reason over persona and field-level constraints. 
Frontier backbones such as Sonnet 4.6 and Opus 4.7 obtain the strongest \EvaluationTarget{} closure with relatively low false-fire rates, while smaller or open-weight backbones (Qwen3 32B, GPT-OSS 20B) often improve on individual diagnostic columns but remain unstable on persona-source recall, grounding, or false-fire control. 
Thus, \JudgeName{} supplies the missing output contract, but the absolute closure level is still bounded by the backbone's ability to execute that contract.

\paragraph{Finding 4: persona-source diagnosis and component localization expose hidden failures.}
Tabs.~\ref{tab:persona_swap_results} and~\ref{tab:component_consistency} unpack two failure modes that are easy to hide behind scalar quality scores. 
Tab.~\ref{tab:persona_swap_results} shows that personalization must be evaluated as a source of invalidity: the response is unchanged, but the swapped persona makes it wrong, so credit requires predicting \textsc{PREF\_CONFLICT} rather than merely lowering an overall score. 
Tab.~\ref{tab:component_consistency} separates detecting a cross-component defect from localizing the response field that must be fixed; this distinction is central to actionable debugging because a judge can know that a response is flawed while still failing to say where the repair is needed.

\begin{figure}[!htbp]
\centering
\includegraphics[width=\columnwidth]{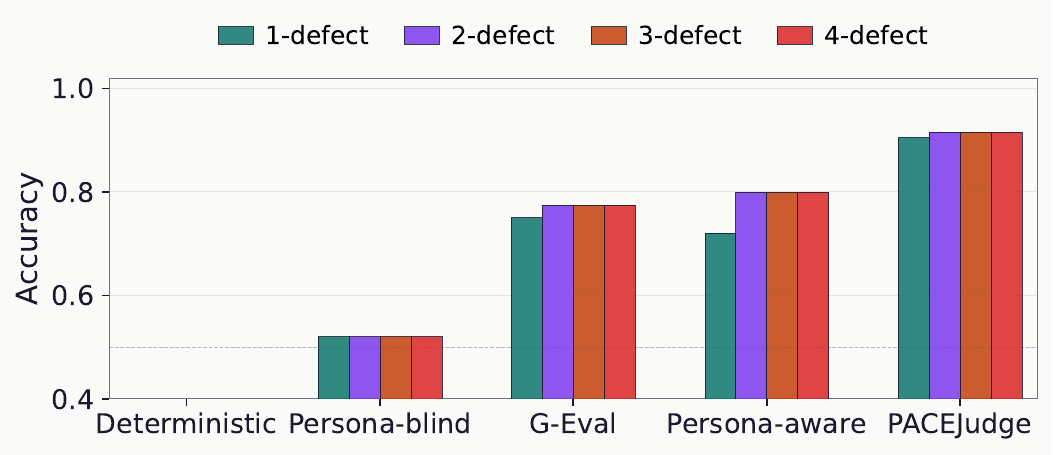}
\caption{
\textbf{GOOD/BAD accuracy by defect tier.}
BAD records are grouped by the number of injected defects. 
The 1-defect tier isolates the hardest setting for surface-level detection, while higher tiers expose whether methods mainly benefit from accumulating defects.
}
\label{fig:different_defect_level}
\vspace{-.1in}
\end{figure}

\paragraph{Finding 5: gains are most informative on low-signal, single-defect cases.}
The four-tier BAD construction (1-, 2-, 3-, and 4-defect BAD records) lets us test whether judges diagnose individual failures or merely accumulate surface alarms as more defects are added. 
Fig.~\ref{fig:different_defect_level} shows that the largest protocol gap appears on one-defect records, where only a single response field is wrong and surface signal is weakest. 
\JudgeName{} remains stable across the tiered difficulty gradient, suggesting that its advantage comes from structured diagnosis rather than simply detecting that heavily corrupted responses look bad.

\begin{table}[t]
\centering
\scriptsize
\renewcommand{\arraystretch}{1.10}

\begin{tabular}{l cccc}
\toprule
& \multicolumn{4}{c}{\textbf{Persona-swap accuracy by axis}} \\
\cmidrule(lr){2-5}
\textbf{Prompt} & budget & hard & brand & attr. \\
\midrule
G-Eval         & 0.24 & 0.16 & 0.16 & 0.12 \\
Persona-aware  & 0.30 & 0.25 & 0.14 & 0.11 \\
\textbf{\JudgeName{} (ours)} & \textbf{0.37} & \textbf{0.32} & \textbf{0.36} & \textbf{0.23} \\
\bottomrule
\end{tabular}
\caption{
\textbf{Persona-source diagnosis by conflict axis, averaged over backbones.}
A hit requires predicting \textsc{PREF\_CONFLICT} on a counterfactual persona swap. 
\JudgeName{} is strongest across all axes, showing that personalization failures require source-level diagnosis rather than broad BAD detection. 
Axis means correspond to S1 in Tab.~\ref{tab:sota_failure_main}; full per-backbone results are in Appx.~\ref{app:persona_swap_full}.
}
\label{tab:persona_swap_results}
\end{table}

\begin{table}[t]
\centering
\scriptsize
\renewcommand{\arraystretch}{1.10}
\begin{tabular}{l cc}
\toprule
\textbf{Prompt} & \textbf{detect} & \textbf{4-way loc.} \\
\midrule
G-Eval         & 0.75 & 0.70 \\
Persona-aware  & 0.79 & 0.64 \\
\textbf{\JudgeName{} (ours)} & \textbf{0.93} & \textbf{0.72} \\
\bottomrule
\end{tabular}
\caption{
\textbf{Cross-component detection and localization.}
Detection flags whether a compositional defect is present; 4-way localization requires identifying the failed response field. 
\JudgeName{} is strongest on both, showing that compositional failures require field-level diagnosis beyond broad BAD detection. 
Scores correspond to S2 detect / loc. in Tab.~\ref{tab:sota_failure_main}; full per-backbone results are in Appx.~\ref{app:component_consistency_full}.
}
\label{tab:component_consistency}
\end{table}


\section{Conclusion}

We argued that shopping-assistant evaluation should be treated as a joint \EvaluationTarget{} problem: responses must be personalized, actionable, compositional, and evidence-grounded. 
\BenchmarkName{} makes this setting measurable through controlled and verifiable records with structured personas, auditable evidence pools, and gold defect family/location labels. 
Across seven backbones and multiple judging configurations, scalar and generic rubric judges often recognize broad GOOD/BAD quality but fail to recover the diagnostic fields needed for debugging. 
\JudgeName{} improves these closure metrics without retraining. 
These results support our central claim: for structured shopping assistants, the key evaluation challenge is not broad quality scoring but producing a task-aligned diagnosis of what failed, where, and why.

\clearpage
\newpage
\section*{Limitations}

\BenchmarkName{} and \JudgeName{} have several bounded limitations.
\textbf{(1) Synthetic GOOD/BAD construction.}
GOOD responses come from a single generator (Claude Opus 4.6) and BAD responses from rule-based defect injection; the cross-generator and real-assistant slices audit but do not fully reproduce live traffic.
\textbf{(2) Evidence depth.}
Most evidence records carry product-listing fields only; 129/4{,}460 are review-augmented, bounding how deeply grounding can be audited per claim.
\textbf{(3) Holdout scope.}
Held-out persona bundles cover 345/1{,}200 personas across 8 families; this is narrower than real-user generalization across demographics, regions, and languages. The dataset is English-only.

\section*{Potential Risks}

\BenchmarkName{} and \JudgeName{} are intended for evaluating structured shopping-assistant responses, not for directly certifying production systems or replacing human review. 
A main risk is personalization. 
Although the personas are synthetic or structured benchmark artifacts, personalization evaluation can encourage systems to infer or exploit user traits if transferred carelessly to real deployments. 
Practical use should therefore avoid sensitive personal attributes, preserve user privacy, and treat persona-conditioned evaluation as a tool for detecting misalignment rather than for maximizing persuasion or user targeting.

\bibliography{custom}

\clearpage

\appendix

\raggedbottom

\setcounter{topnumber}{4}
\setcounter{bottomnumber}{2}
\setcounter{totalnumber}{6}
\renewcommand{\topfraction}{0.95}
\renewcommand{\bottomfraction}{0.5}
\renewcommand{\textfraction}{0.05}
\renewcommand{\floatpagefraction}{0.7}


\section{Existing Benchmark Datasets}
\label{app:benchmark_datasets}

\begin{table*}[!b]
\centering
\small
\setlength{\tabcolsep}{6pt}
\renewcommand{\arraystretch}{1.10}
\begin{tabular}{p{0.30\textwidth} p{0.62\textwidth}}
\toprule
\textbf{Native resource} & \textbf{Native evaluation protocol} \\
\midrule
LaMP \cite{salemi2023lamp} & Per-task accuracy or macro-F1 (classification tasks) and ROUGE-1/ROUGE-L (generation tasks) on user-conditioned generation. \\
PersonaLens \cite{zhao2025personalens} & LLM-judge agents score an assistant turn on three axes: personalization, response quality, and task success. \\
PrefEval \cite{zhao2025prefeval} & Preference-following accuracy (generation and classification variants) on long-context dialogues with stated user preferences. \\
PersonaMem \cite{jiang2025personamem} & Multiple-choice accuracy: pick the reply matching the user's evolving profile across $\leq\!60$ chat sessions. \\
LikeBench \cite{rahman2025likebench} & Per-turn likability score from a simulated user, decomposed into seven dimensions (emotion, formality, humor, callback, etc.). \\
EtaPP \cite{hao2025etapp} & Key-point-based LLM-as-judge against human-annotated key points on personalized tool-call trajectories. \\
Personalized judge \cite{dong2024personalizedjudge} & Agreement with human ground truth on a binary preference task (overall and high-certainty subset), with verbal confidence. \\
ECom-Bench \cite{wang2025ecombench} & Task-success trajectories in e-commerce support; agents simulate hundreds of personas across realistic commerce tasks. \\
Search Arena \cite{miroyan2025searcharena} & Pairwise human preference votes on full traces from two search-augmented chat systems (Bradley--Terry style). \\
ARES \cite{saadfalcon2024ares} & Three PPI-corrected fine-tuned LM judges scoring context relevance, answer faithfulness, and answer relevance for RAG. \\
\bottomrule
\end{tabular}
\caption{Native judging protocols of the prior personalization, evidence-grounding, and judging resources cited in the main text. None of these protocols defines a closure metric for the four \EvaluationTarget{} targets (Personalized, Compositional, Evidence-grounded, Actionable); they evaluate user-conditioned generation, MCQ over user history, RAG faithfulness, or pairwise human votes, but not the joint persona--compositional--grounded--actionable diagnosis that \BenchmarkName{} requires. We additionally implement honest single-target adapted baselines based on ARES (E target), PersonaLens (P target), and EtaPP (A target) and report them in the main experiment table (Table~\ref{tab:sota_failure_main}); the remaining seven resources are not run on \BenchmarkName{} because their native judging protocols cannot be honestly mapped to the four-target diagnosis without forcing inapplicable scores.}
\label{tab:native_prior_protocols}
\end{table*}

This section documents how the prior resources cited in
Tab.~\ref{tab:related_work_pace} relate to the PACEShop evaluation
setting. The goal is not to argue that prior personalization,
shopping, RAG, or LLM-as-judge benchmarks are weak. Rather, they were
designed for different native tasks: user-conditioned generation,
dialogue preference following, memory retrieval, task success,
pairwise search comparison, or RAG faithfulness. These protocols
measure important pieces of the problem, but they do not define the
same closure target as PACEShop: given a query, persona, four-field
shopping response, and evidence pool, the evaluator must diagnose
whether the response is personalized, cross-field consistent,
evidence-grounded, and actionable for debugging.

Tab.~\ref{tab:native_prior_protocols} therefore lists each resource
by its \emph{native evaluation protocol}. This distinction matters for
the experiment tables. A benchmark may be strongly related to one PACE
target while still lacking the output fields needed for the S1--S4
closure metrics. For example, personalization benchmarks can evaluate
whether a response follows user preferences, but usually do not define
a persona-swap defect family and response-field location. RAG
evaluators can measure faithfulness to retrieved context, but usually
do not check whether a response invents shopper history or whether the
judge itself cites evidence IDs outside the evidence pool. Shopping
task benchmarks can evaluate realistic user trajectories, but they do
not provide gold defect families and locations for four-field
shopping-assistant responses.

For this reason, the main experiment table
(Tab.~\ref{tab:sota_failure_main}) does not force every prior
resource into every PACEShop scenario. Instead, unsupported native
judging protocols are marked as ``--''. We implement honest single-target
adapted baselines only when the native judging protocol gives a defensible
mapping to one \EvaluationTarget{} target: \textbf{ARES-style} for evidence grounding
(E), \textbf{PersonaLens-style} for personalization (P), and
\textbf{EtaPP-style} for actionable key-point diagnosis (A). The other
resources remain comparison points for problem setting and protocol
coverage, but are not converted into artificial all-target baselines.
This keeps the appendix comparison auditable: ``--'' means the native
protocol does not define the scored PACEShop output field, not that
the resource is unimportant.

\section{\BenchmarkName{} Dataset Construction}
\label{app:paceshop_construction}

This appendix documents the construction-side material that supports
\BenchmarkName{}. We first describe the full construction pipeline
and the validity layers
(Secs.~\ref{app:benchmark_workflow}--\ref{app:defect_taxonomy});
we close with concrete records, JSON input examples, the
response-generation prompt, holdout-persona statistics, and
source-side coverage tables that make the benchmark auditable
(Secs.~\ref{app:worked_example}--\ref{app:source_coverage}).

\subsection{\BenchmarkName{} Dataset Statistics}

Tab.~\ref{tab:dataset_summary} summarizes the benchmark statistics, evidence coverage, defect structure, and validation assets.

\begin{table}[H]
\centering
\scriptsize
\setlength{\tabcolsep}{3pt}
\renewcommand{\arraystretch}{1.05}
\resizebox{\columnwidth}{!}{%
\begin{tabular}{@{}llr@{}}
\toprule
\textbf{Block} & \textbf{Statistic} & \textbf{Value} \\
\midrule
\multirow{4}{*}{\textsc{Source}}
  & Backbone queries (public) & 97{,}227 \\
  & Evidence-backed queries ($\geq 3$ products) & 1{,}132 \\
  & Persona pool (constraints+history+brand) & 1{,}200 \\
  & Query--persona pairs & 4{,}525 \\
\midrule
\multirow{3}{*}{\textsc{Evidence}}
  & Evidence records (11 domains) & 4{,}460 \\
  & Review-augmented records & 129 \\
  & Median evidence IDs / response & 10 \\
\midrule
\multirow{3}{*}{\textsc{Dataset}}
  & GOOD / BAD / total & 4{,}525 / 18{,}100 / 22{,}625 \\
  & Train / dev / test & 15{,}825 / 3{,}400 / 3{,}400 \\
  & Stealth-defect fraction & 35.1\% \\
\midrule
\multirow{3}{*}{\textsc{Defects}}
  & Families\textsuperscript{\textdagger} $\times$ field groups & 7 $\times$ 4 \\
  & Multi-defect tiers (1/2/3/4) & 4{,}525 each \\
  & Gold \{family, location\} on every BAD & \cmark \\
\midrule
\multirow{4}{*}{\textsc{Targets covered}}
  & \textbf{P}ersonalized (persona-swap slice) & \cmark \\
  & \textbf{A}ctionable (gold family+location) & \cmark \\
  & \textbf{C}ompositional (cross-field defects) & \cmark \\
  & \textbf{E}vidence-grounded (product + history) & \cmark \\
\midrule
\multirow{5}{*}{\textsc{Validity check}}
  & L1 deterministic checks (12/record) & 99.3\% pass \\
  & L2 evidence provenance flag & 4{,}331 / 129 \\
  & L3 human-calibration package & 250 items \\
  & L4 cross-generator slice (3 models) & 50 shared \\
  & L5 real-assistant slice & 100 outputs \\
\bottomrule
\end{tabular}%
}

\smallskip
{\scriptsize \textsuperscript{\textdagger}PREF\_CONFLICT, INTENT\_DRIFT, XCOMP\_MISMATCH, REDUNDANCY, EVIDENCE\_MISMATCH, OVER\_PERSONALIZATION\_HALLUCINATION, UNSUPPORTED\_CLAIM. Field groups: overview, categories, related queries, attributions.}

\caption{Summary of the \BenchmarkName{} benchmark. The ``Targets covered'' block names which of the four \EvaluationTarget{} targets each design choice supports; the ``Validity stack'' block enumerates the five verification layers that make the construction auditable.}
\label{tab:dataset_summary}
\end{table}

\subsection{Construction Pipeline Workflow}
\label{app:benchmark_workflow}

\BenchmarkName{} is built from public source artefacts in a
controlled, auditable pipeline.

\paragraph{Persona construction.}
We construct personas as structured shopping contexts rather than free-form user descriptions. 
The released persona pool contains 1,200 personas, including public seed personas and behavior-seeded synthetic personas normalized under a shared schema. 
Each persona specifies stable shopper information, such as household type, budget, quality sensitivity, urgency, constraints, preferences, and brand affinities, together with concrete shopping behavior in the form of purchase history and recent searches. 
This design supports two kinds of evaluation. 
First, shopper constraints and preferences make persona-conditioned correctness testable: a response can be valid for one shopper and invalid for another. 
Second, purchase and search histories make persona-history grounding auditable: a judge can check whether a response faithfully refers to the shopper's actual history or invents unsupported behavior.

\paragraph{Shopping-behavior and evidence collection.}
We begin from 97,227 normalized US shopping queries and 4,460 public evidence records across 11 shopping domains. 
Queries are tagged by coarse auditing metadata, including retail vertical and shopping mission, while evidence records expose product titles, brands, metadata, and available review snippets. 
We retain 1,132 queries whose evidence pool contains at least three products. 
This evidence-backed filtering is deliberate: PACEShop asks whether the same shopping intent can support different valid responses under different personas, which requires enough product alternatives for shopper constraints and preferences to matter. 
Each retained query is paired with four distinct personas using a coverage-balanced assignment procedure that minimizes persona-taxonomy reuse and prevents the same persona from repeating on a query. 
This yields 4,525 query--persona pairs after validation drops. 
The resulting repeated-measures design holds the shopping query and evidence pool fixed while varying shopper context, allowing personalization failures to be isolated from changes in shopping intent.

\paragraph{The full four-stage workflow.}

\begin{enumerate}[leftmargin=1.2em,itemsep=2pt,topsep=2pt]
\item \textbf{Query catalog.} Normalize 97{,}227 US ESCI queries; tag retail vertical (11 categories) and shopping mission (6 categories). Detailed breakdown can be found in Appx.~\ref{app:source_coverage}. 

\item \textbf{Evidence catalog.} Build 4{,}460 public evidence records across 11 shopping domains; augment 129 with review snippets. 

\item \textbf{Persona pool.} Merge 200 public v1 personas with 1{,}000 behavior-seeded synthetic personas; normalize under one structured schema (household, budget, quality, urgency, constraints, preferences, brand affinities, history).
\item \textbf{Coverage-balanced assignment.} Retain 1{,}132 queries with $\geq$3 evidence candidates; reserve 345 holdout personas across 8 bundles for dev/test; assign 4 personas per retained query.
\item \textbf{GOOD generation.} One structured response per assignment, Claude Opus 4.6, constrained JSON output.
\item \textbf{Normalization.} Schema validation and deterministic post-processing.
\item \textbf{BAD construction.} Four tiers (1/2/3/4 defects) per GOOD, defects drawn from distinct field groups.
\item \textbf{Packaging.} Export packaged / explicit / clean JSONL views plus 250-item human-calibration package, 50-input cross-generator slice, 100-output real-assistant slice.
\end{enumerate}

\noindent
The four substantive stages of this workflow are summarized in
Tab.~\ref{tab:construction_pipeline}, which makes explicit how each
stage's design choice advances one or more PACE properties.

\begin{table*}[t]
\centering
\small
\begin{tabular}{p{0.14\linewidth} p{0.27\linewidth} p{0.27\linewidth} p{0.22\linewidth}}
\toprule
\textbf{Stage} & \textbf{Motivation} & \textbf{Operation} & \textbf{PACE role} \\
\midrule
Source profiling &
Ensure broad, auditable coverage over shopping intents, personas, and evidence. &
Tag queries, personas, and evidence records by retail vertical, mission, household, budget, constraints, preferences, and history. &
Creates controlled coverage for P and E. \\
\midrule
Evidence-backed retention and persona assignment &
Make persona differences meaningful while holding query intent fixed. &
Retain queries with at least three evidence-backed products; pair each query with four distinct personas under coverage-balanced assignment. &
Supports persona-swap and repeated-measures evaluation for P. \\
\midrule
GOOD response generation &
Create valid structured responses under the target response contract. &
Generate overview, categories, related queries, and attributions; normalize and validate schema and evidence IDs. &
Defines valid multi-field objects for C and E. \\
\midrule
BAD defect injection &
Create known, localizable failures rather than relying on vague quality differences. &
Inject one to four controlled defects from a 7-family $\times$ 4-field taxonomy. &
Provides gold defect family/location for A and scenario-specific tests for P/C/E. \\
\bottomrule
\end{tabular}
\caption{
\BenchmarkName{} construction pipeline. Each stage is designed to make one or more PACE properties observable and verifiable.
}
\label{tab:construction_pipeline}
\end{table*}

\subsection{\BenchmarkName{} Coverage Details}
\label{app:source_coverage}

The released persona pool, evidence catalog, and query catalog are
documented below with representative examples. The query catalog
covers 11 retail verticals and 6 shopping missions; the persona pool
covers 12 household types and 8 holdout bundle families; the evidence
catalog spans 11 shopping domains and includes 129 review-augmented
records. 

\begin{table}[h]
\centering
\footnotesize
\begin{tabular}{lrr}
\toprule
\textbf{Retail vertical} & \textbf{Backbone} & \textbf{Retained} \\
\midrule
Toys \& Games         & 17{,}279 & 235 \\
Electronics            &  8{,}044 & 186 \\
Tools \& Automotive    & 13{,}429 & 157 \\
Home \& Kitchen        & 12{,}439 & 144 \\
Apparel \& Accessories & 13{,}874 & 104 \\
Pet \& Baby            &  6{,}859 & 76 \\
General Merchandise    &  7{,}216 & 66 \\
Sports \& Outdoors     &  6{,}196 & 62 \\
Office \& School       &  5{,}009 & 46 \\
Beauty \& Health       &  2{,}901 & 33 \\
Books \& Media         &  3{,}981 & 23 \\
\midrule
\textbf{Total}         & \textbf{97{,}227} & \textbf{1{,}132} \\
\bottomrule
\end{tabular}
\caption{Query distribution across 11 retail verticals. \emph{Backbone} counts all unique US normalized ESCI queries; \emph{Retained} counts the 1{,}132 evidence-backed queries that pass the evidence-availability filter ($\geq$3 evidence-backed candidates).}
\label{tab:query_vertical_coverage}
\end{table}

\begin{table}[h]
\centering
\footnotesize
\begin{tabular}{lrr}
\toprule
\textbf{Shopping mission type} & \textbf{Backbone} & \textbf{Retained} \\
\midrule
Broad Browse                 & 32{,}889 & 312 \\
Replenishment \& Repeat      & 20{,}166 & 241 \\
Problem Solution             & 14{,}533 & 217 \\
Compatibility \& Replacement &  9{,}807 & 144 \\
Gift \& Event                & 12{,}386 & 129 \\
Feature Constrained          &  7{,}446 &  89 \\
\midrule
\textbf{Total}               & \textbf{97{,}227} & \textbf{1{,}132} \\
\bottomrule
\end{tabular}
\caption{Query distribution across 6 shopping mission types. Mission type is assigned by intent-pattern matching against query text and top product titles/brands.}
\label{tab:query_mission_coverage}
\end{table}

\clearpage
\onecolumn
\raggedbottom

\subsubsection{Query Coverage}
\label{app:query_catalog}

\noindent The public backbone contains 97{,}227 unique US-locale normalized ESCI queries categorized along two dimensions.

\myparagraph{Retail verticals (11 categories).} These follow standard e-commerce product departments (e.g., Amazon Browse Nodes). Each query is assigned to one vertical by a keyword-hit counting rule (\texttt{\_pick\_rule()} in the released code): the assignment function concatenates the normalized query text with the top candidate product titles and brands, then counts the number of keyword hits for each vertical's keyword list. The vertical with the most hits wins; ties are broken by list order, and if no keyword matches the query falls back to General Merchandise. Representative keyword$\to$vertical mappings include:
\begin{itemize}[leftmargin=1.2em, itemsep=0pt, topsep=2pt]
    \item ``shirt'', ``shoes'', ``backpack'' $\to$ Apparel \& Accessories
    \item ``usb'', ``charger'', ``headphones'' $\to$ Electronics
    \item ``kitchen'', ``vacuum'', ``air fryer'' $\to$ Home \& Kitchen
    \item ``toy'', ``lego'', ``puzzle'' $\to$ Toys \& Games
    \item ``dog'', ``baby'', ``stroller'' $\to$ Pet \& Baby
    \item ``book'', ``vinyl'', ``dvd'' $\to$ Books \& Media
\end{itemize}
The full keyword lists (10--19 terms per vertical) are provided in the released \texttt{v2\_taxonomy.py}. The lists are intentionally broad rather than exhaustive: common category-indicative terms are sufficient for coarse coverage reporting, since the taxonomy serves as an auditing tool rather than a fine-grained classifier.

\myparagraph{Shopping mission types (6 categories).} These capture the shopper's task framing and follow well-known shopping-behavior distinctions in the e-commerce literature: undirected browsing, repeat/replenishment purchases, problem-driven search, gift/event shopping, compatibility/replacement needs, and feature-constrained filtering. Assignment uses the same \texttt{\_pick\_rule()} keyword-hit counting mechanism as retail verticals, applied to intent-indicative keyword lists. Representative keyword$\to$mission mappings include:
\begin{itemize}[leftmargin=1.2em, itemsep=0pt, topsep=2pt]
    \item ``gift'', ``birthday'', ``wedding'' $\to$ Gift \& Event
    \item ``best'', ``quiet'', ``waterproof'' $\to$ Problem Solution
    \item ``wireless'', ``organic'', ``compact'' $\to$ Feature Constrained
    \item ``compatible'', ``replacement'', ``adapter'' $\to$ Compatibility \& Replacement
    \item ``bulk'', ``refill'', ``pods'' $\to$ Replenishment \& Repeat
\end{itemize}
The fallback category (no keyword hits) is Broad Browse. Full keyword lists (7--12 terms per mission type) are in \texttt{v2\_taxonomy.py}.

Table~\ref{tab:query_vertical_coverage} shows the backbone distribution across the 11 retail verticals together with the number of evidence-backed queries retained after evidence filtering. Table~\ref{tab:query_mission_coverage} shows the corresponding distribution across the 6 shopping mission types.

\noindent The retention rate varies across verticals and missions because it depends on evidence availability in the public ESCI product metadata rather than on intentional stratification. For example, Beauty \& Health retains 33 queries (1.14\%) while Toys \& Games retains 235 (1.36\%), reflecting differences in product-metadata completeness across categories.

\clearpage
\twocolumn

\subsubsection{Persona Coverage}
\label{app:persona_catalog}

\noindent The full released persona pool contains 1{,}200 personas. 
Table~\ref{tab:released_persona_coverage} summarizes the aggregate coverage across the major taxonomy axes, including household type, budget, quality preference, urgency, constraints, shopping preferences, behavioral-history density, and holdout attributes. 
Table~\ref{tab:representative_released_personas} then provides a compact set of representative released personas, selected to illustrate the major coverage patterns discussed in the main text without listing the full persona catalog.

\begin{table*}[t]
\centering
\footnotesize
\setlength{\tabcolsep}{4pt}
\renewcommand{\arraystretch}{0.95}
\resizebox{\textwidth}{!}{%
\begin{tabular}{p{0.14\textwidth} p{0.78\textwidth}}
\toprule
\textbf{Axis} & \textbf{Released coverage summary} \\
\midrule
Household & couple (117); creator hobbyist (38); home office (74); parents school age (121); parents young children (132); pet owner (97); roommates (81); senior caregiver (81); single professional (136); small space (104); student (88); travel outdoor (131) \\
Budget & budget conscious (380); premium (401); value balanced (419) \\
Quality & balanced (406); functional (416); high (378) \\
Urgency & fast (380); standard (420); urgent (400) \\
Constraints & allergen ingredient (406); compatibility (162); durability (135); maintenance noise (334); material avoidance (514); safety age (148); size space (206) \\
Preferences & aesthetics (467); convenience (665); giftability (224); performance (282); portability (619); premium finish (314); sustainability (672) \\
History & moderate (566); rich (588); sparse (46) \\
Holdout flags & eco-conscious materials (115); sensitive skin (115); smart-home compatibility priority (115) \\
\bottomrule
\end{tabular}%
}
\caption{Released persona coverage summary used in \BenchmarkName{}. Counts are over the full 1{,}200-persona pool, while the rows below show representative released personas rather than the full catalog.}
\label{tab:released_persona_coverage}
\end{table*}

\subsubsection{Evidence Coverage}
\label{app:evidence_catalog}
This subsection expands the evidence side of \BenchmarkName{} in two
complementary views. Tab.~\ref{tab:evidence_domain_coverage}
summarizes the domain-level coverage of the released evidence catalog,
while Tab.~\ref{tab:representative_evidence_view} shows representative
evidence records from each domain so readers can inspect the concrete
product information available to the response generator and judge.

\noindent The full released evidence catalog contains 4{,}460 public evidence records spanning 11 shopping domains. Each record exposes human-readable product information extracted from public ESCI product listings: product title, brand, color, bullet points, and description. Of these, 4{,}331 records (97.1\%) are \emph{product-listing only}, grounded solely in these catalog fields. The remaining 129 records (2.9\%) are \emph{review-augmented}, carrying the same product-listing fields plus public customer review snippets that provide richer grounding signals (e.g., real-user opinions on durability, fit, or usability). Every record carries an explicit provenance flag so that downstream judges and analyses can condition on evidence strength. This design keeps the grounding signal lighter than fully review-grounded RAG benchmarks, but more transparent than a generic ``grounded'' claim because evidence quality is measurable per record rather than assumed uniform. Table~\ref{tab:evidence_domain_coverage} gives the aggregate domain distribution of the evidence catalog, while Table~\ref{tab:representative_evidence_view} provides a representative released evidence record from each domain. Together, the two tables show both the breadth of evidence coverage and the concrete product-level information available in each per-query evidence pool.

\begin{table}[h]
\centering
\footnotesize
\begin{tabular}{lrr}
\toprule
\textbf{Shopping domain} & \textbf{Records} & \textbf{\% of total} \\
\midrule
General Merchandise    & 1{,}232 & 27.6\% \\
Toys \& Games          &    623 & 14.0\% \\
Apparel \& Accessories &    610 & 13.7\% \\
Home \& Kitchen        &    418 &  9.4\% \\
Tools \& Automotive    &    375 &  8.4\% \\
Electronics            &    335 &  7.5\% \\
Pet \& Baby            &    267 &  6.0\% \\
Sports \& Outdoors     &    225 &  5.0\% \\
Office \& School       &    191 &  4.3\% \\
Books \& Media         &    104 &  2.3\% \\
Beauty \& Health       &     80 &  1.8\% \\
\midrule
\textbf{Total}         & \textbf{4{,}460} & \textbf{100\%} \\
\bottomrule
\end{tabular}
\caption{Evidence record distribution across 11 shopping domains. Of these, 129 records (2.9\%) are review-augmented; the remainder are grounded in product-listing fields only (title, brand, color, bullets, description).}
\label{tab:evidence_domain_coverage}
\end{table}

\begin{table*}[t]
\centering
\scriptsize
\setlength{\tabcolsep}{3pt}
\renewcommand{\arraystretch}{0.92}
\resizebox{\textwidth}{!}{%
\begin{tabular}{p{0.15\textwidth} p{0.27\textwidth} p{0.52\textwidth}}
\toprule
\textbf{Persona ID} & \textbf{Coverage highlights} & \textbf{Representative shopper sketch} \\
\midrule
\texttt{v2-persona-0276} & household: parents school age; budget: value balanced; quality: high; urgency: urgent; constraint: durability/maintenance noise; preference: aesthetics/convenience; history: moderate; holdout: eco-conscious materials & family with school-age kids; mid-range, quality first, needs it urgently; constraints: quiet operation required, durable enough for daily use; attrs: eco-conscious materials, travel friendly \\
\texttt{v2-persona-0208} & household: home office; budget: premium; quality: balanced; urgency: fast; constraint: allergen ingredient/material avoidance; preference: convenience/premium finish; history: rich; holdout: smart-home compatibility priority & home office household; premium, balanced, needs it this week; constraints: fragrance-free only, kid-safe materials only; attrs: smart-home compatibility priority, ergonomic support \\
\texttt{v2-persona-1130} & household: senior caregiver; budget: budget conscious; quality: functional; urgency: standard; constraint: compatibility/size space; preference: convenience/performance; history: moderate & senior caregiver household; budget, value oriented, standard shipping; constraints: must fit existing device compatibility, small-space friendly; attrs: organic cotton, ergonomic support \\
\texttt{persona-0027} & household: single professional; budget: premium; quality: high; urgency: urgent; constraint: material avoidance/size space; preference: performance/portability; history: sparse; holdout: sensitive skin & single professional; premium, quality first, needs it urgently; constraints: small-space friendly, avoid wool; attrs: sensitive skin, compact storage \\
\texttt{persona-0030} & household: travel outdoor; budget: value balanced; quality: high; urgency: standard; constraint: allergen ingredient/material avoidance; preference: convenience/performance; history: moderate; holdout: sensitive skin & active outdoor household; mid-range, quality first, standard shipping; constraints: fragrance-free only, latex-free materials; attrs: sensitive skin, eco-conscious materials \\
\texttt{persona-0009} & household: pet owner; budget: value balanced; quality: functional; urgency: standard; constraint: allergen ingredient; preference: convenience/premium finish; history: rich; holdout: sensitive skin & active outdoor household; mid-range, value oriented, standard shipping; constraints: fragrance-free only; attrs: sensitive skin, low-maintenance finishes \\
\texttt{v2-persona-0289} & household: student; budget: premium; quality: balanced; urgency: standard; constraint: allergen ingredient; preference: aesthetics/convenience; history: rich; holdout: smart-home compatibility priority & student; premium, balanced, standard shipping; constraints: avoid peanuts in ingredients; attrs: smart-home compatibility priority, compact storage \\
\texttt{v2-persona-0467} & household: parents young children; budget: budget conscious; quality: functional; urgency: standard; constraint: allergen ingredient; preference: giftability/performance; history: moderate; holdout: sensitive skin & family with toddler; budget, value oriented, standard shipping; constraints: avoid peanuts in ingredients; attrs: sensitive skin, compact storage \\
\texttt{persona-0022} & household: couple; budget: budget conscious; quality: balanced; urgency: fast; constraint: allergen ingredient/material avoidance; preference: convenience/portability; history: rich; holdout: eco-conscious materials & couple in a smart home; budget, balanced, needs it this week; constraints: avoid peanuts in ingredients, latex-free materials; attrs: eco-conscious materials, travel friendly \\
\texttt{v2-persona-0234} & household: small space; budget: value balanced; quality: balanced; urgency: standard; constraint: compatibility/size space; preference: portability/sustainability; history: moderate; holdout: eco-conscious materials & travel and outdoor household; mid-range, balanced, standard shipping; constraints: small-space friendly, must fit existing device compatibility; attrs: eco-conscious materials, travel friendly \\
\texttt{v2-persona-0226} & household: roommates; budget: budget conscious; quality: functional; urgency: fast; constraint: material avoidance; preference: aesthetics/convenience; history: rich; holdout: smart-home compatibility priority & roommates; budget, value oriented, needs it this week; constraints: avoid wool, latex-free materials; attrs: smart-home compatibility priority, compact storage \\
\texttt{v2-persona-0223} & household: creator hobbyist; budget: budget conscious; quality: high; urgency: urgent; constraint: maintenance noise/material avoidance; preference: giftability/performance; history: moderate; holdout: smart-home compatibility priority & creator and hobbyist household; budget, quality first, needs it urgently; constraints: quiet operation required, latex-free materials; attrs: smart-home compatibility priority, eco-conscious materials \\
\bottomrule
\end{tabular}%
}
\caption{Representative released personas covering the major taxonomy values emphasized in the paper.}
\label{tab:representative_released_personas}
\end{table*}


\FloatBarrier

\begin{table*}[t]
\centering
\footnotesize
\setlength{\tabcolsep}{3pt}
\renewcommand{\arraystretch}{0.95}
\resizebox{\textwidth}{!}{%
\begin{tabular}{p{0.14\textwidth} p{0.07\textwidth} p{0.13\textwidth} p{0.16\textwidth} p{0.44\textwidth}}
\toprule
\textbf{Domain} & \textbf{Count} & \textbf{Evidence ID} & \textbf{Brand} & \textbf{Representative title} \\
\midrule
apparel accessories & 610 & \texttt{B07T3DJ4TF} & DC Comics & Youth Boys' Classic Batman Costume Hoodie Sweatshirt-Small \\
beauty health & 80 & \texttt{B07YNJGVX8} & Respected Roots & Respected Roots Shaving Cream All Natural Shave Cream for Men and Body - Travel Shaving Cream with Tree Tea Oil to Shave Sensitive Skin, Mens Pre-Shave lotion. \\
books media & 104 & \texttt{0679767398} & Vintage & South of the Border, West of the Sun: A Novel \\
electronics & 335 & \texttt{B07HK6K5CS} & LENRUE & Portable Bluetooth Speakers,Dual-Driver Wireless Speaker with Surround Stereo Sound and More Bass,for iPhone and Samsung Android ... (Black) \\
general merchandise & 1232 & \texttt{B00MG5FZWA} & Embroidex & Embroidex Sewing Kit for Home, Travel \& Emergencies - Filled with Quality Notions Scissor \& Thread - Great Gift \\
home kitchen & 418 & \texttt{B07SNPJSB4} & SHINEURI & SHINEURI 2 1/2 qt Copper Saucepan with Lid, Mini Saute Pan with Stainless Steel Handle - Cooking for Soup, Stew, Sauce, Pasta \& Reheat Food, Compatible for Induction, Gas, Electric \& Stovetops \\
office school & 191 & \texttt{1939814456} & Handwriting Without Tears & Learning Without Tears - My Printing Book Student Workbook, Current Edition - Handwriting Without Tears Series - 1st Grade Writing Book - Letters, Language Arts Lessons - for School or Home Use \\
pet baby & 267 & \texttt{B07WV78D6S} & M JJYPET & M JJYPET Interactive Cat Toys,Rechargeable 3 in 1 Cat Red Dot Cat Kitten Dog Toy \\
sports outdoors & 225 & \texttt{B00339C3PK} & Coleman & Coleman Folding Camp Chair | Woodsman II Portable Outdoor Chair, 17" x 17.5" \\
tools automotive & 375 & \texttt{B0000AXFST} & Northern Tool and Equipment & Fast Framer Universal Storage Shed Framing Kit \\
toys games & 623 & \texttt{B0742MSVH8} & The Treemendous Ornament Decorator & TreeMendous Christmas Tree Ornament Decorating Kit for Kids Ages 3 and up - Top Rated Craft Activity Game, Holiday Toy DIY Ornament Maker \\
\bottomrule
\end{tabular}%
}
\caption{Representative evidence view for \BenchmarkName{}. Each row reports one domain count and one representative released evidence record from that domain.}
\label{tab:representative_evidence_view}
\end{table*}

\subsection{Validity Check}
\label{app:validity_stack}

We do not treat the dataset as self-validating. Five independent
layers make the construction auditable: (i) a 12-check deterministic
record-level validator that every record passes through, (ii) explicit
evidence-provenance flags (metadata vs.\ review-augmented), (iii) a
prepared 250-item human-calibration package, (iv) a cross-generator
slice on 50 shared inputs to confirm that GOOD generation does not
overfit a single backbone, and (v) a normalized real-assistant slice
of 100 external outputs that map cleanly into the same response
contract. Of the 4{,}525 released GOOD records, 99.3\% pass all 12
deterministic checks; the remaining 0.7\% fail only a case-sensitive
category-duplicate check and are flagged but retained.
Tab.~\ref{tab:benchmark_validity_mitigations} maps each validity
layer to the specific construction risk it mitigates.

\begin{table*}[t]
\centering
\small
\setlength{\tabcolsep}{4pt}
\begin{tabular}{p{0.19\linewidth} p{0.36\linewidth} p{0.37\linewidth}}
\toprule
\textbf{Addressed dimensions} & \textbf{Mitigation in benchmark construction} & \textbf{Current measurable support} \\
\midrule
Synthetic-on-synthetic circularity & Keep the main 10K release public and fully auditable, then add manual-calibration and transfer assets rather than treating synthetic data as self-validating. & Prepared stratified \textbf{250}-item human-calibration package with \textbf{110} GOOD and \textbf{140} BAD examples; BAD slice balanced at \textbf{20} items for each of the 7 defect families. \\
Thin grounding signal & Expose evidence strength explicitly and augment metadata with review snippets wherever public coverage exists instead of implying uniformly rich grounding. & \textbf{129} review-augmented evidence records, \textbf{254} review snippets, and exact metadata-vs-review provenance on every released record. \\
Template-like persona realism & Distinguish raw-v1 personas from behavior-seeded synthetic personas and reserve held-out persona bundles for evaluation. & Persona pool of \textbf{1200}: \textbf{200} raw-v1 plus \textbf{1000} behavior-seeded synthetic personas; \textbf{345} holdouts across \textbf{8} bundles. \\
Rigid schema versus ecological validity & Treat the fixed schema as a judge-facing normalization layer and add a relaxed-schema import path for external outputs. & \textbf{100} normalized real-assistant outputs stored with raw originals preserved; persona claims withheld for this slice when persona context is unavailable. \\
Lack of transfer beyond Claude-generated outputs & Add shared-input transfer slices rather than evaluating only on benchmark-generation outputs from one model family. & \textbf{50} shared held-out inputs across \textbf{3} generator families (\emph{completed}); \textbf{100} additional real-assistant outputs for ecological-validity checks. \\
\bottomrule
\end{tabular}
\caption{Threats and mitigations for benchmark validity. The key design choice is to make each concern measurable through released artifacts rather than acknowledging it only qualitatively.}
\label{tab:benchmark_validity_mitigations}
\end{table*}

\subsection{Defect Taxonomy}
\label{app:defect_taxonomy}

BAD records are constructed from seven defect families distributed
across four response-field groups (overview, categories, related
queries, evidence attributions). Each defect family is paired with
the response field whose contract it violates, so localization in
the main experiment table is well posed.
Tab.~\ref{tab:defect_taxonomy_generated} lists the seven families,
their target field groups, and the operational definition used to
inject each defect during BAD construction.

\begin{table*}[!b]
\centering
\small
\resizebox{\textwidth}{!}{%
\begin{tabular}{l l l p{0.38\textwidth}}
\toprule
\textbf{Defect family} & \textbf{Component group} & \textbf{Gold location} & \textbf{Why it matters} \\
\midrule
PREF\_CONFLICT & category & product categories & Introduces an option that directly violates persona constraints or preferences. \\
INTENT\_DRIFT & category & product categories & Shifts the answer away from the original shopping intent while remaining superficially plausible. \\
XCOMP\_MISMATCH & rqList & related queries & Breaks consistency between the overview, categories, and refinement suggestions. \\
REDUNDANCY & rqList & related queries & Produces repetitive or low-diversity refinements that hurt utility. \\
EVIDENCE\_MISMATCH & attribution & evidence attributions & Cites evidence that does not support the claims made in the response. \\
OVER\_PERS\_HALLUC. & overview & personalized overview & Adds user-specific claims that are unsupported by the available evidence. \\
UNSUPPORTED\_CLAIM & overview & personalized overview & Adds a generic but unverifiable claim that lacks support in the evidence store. \\
\midrule
\multicolumn{4}{l}{\textbf{Multi-defect tier design} (defects within the same component group conflict and are never combined):} \\
\midrule
Single-defect & --- & 1 location & 7 single-defect variants, one per family. \\
Double-defect & --- & 2 locations & 18 valid cross-group pairs (e.g., category + attribution). \\
Triple-defect & --- & 3 locations & 20 valid cross-group triples. \\
Quad-defect & --- & 4 locations & 4 curated quads covering all 4 component groups. \\
\bottomrule
\end{tabular}
}
\caption{Release-level defect taxonomy and multi-tier difficulty design used in the full \BenchmarkName{} benchmark. Each BAD example preserves the original query and persona while injecting one to four defects from distinct component groups. The tiered design creates a measurable difficulty gradient---single-defect variants test basic fault detection, while quad-defect variants require simultaneous identification of failures across all four response components.}
\label{tab:defect_taxonomy_generated}
\end{table*}

\subsection{An Example Record of \JudgeName{} Dataset}
\label{app:worked_example}

\paragraph{Benchmark objects.}
In \BenchmarkName{}, the candidate response is instantiated as a structured object
\[
y=(y^{\mathrm{ov}}, y^{\mathrm{cat}}, y^{\mathrm{rq}}, y^{\mathrm{attr}}),
\]
where $y^{\mathrm{ov}}$ is the overview, $y^{\mathrm{cat}}$ is the product-category list, $y^{\mathrm{rq}}$ is the related-query list, and $y^{\mathrm{attr}}$ is the evidence list. 
The persona $p$ contains shopper context, including household profile, budget, quality sensitivity, urgency, constraints, preferences, brand affinities, purchase history, recent searches, and a short context summary. 
The evidence pool $E$ contains product-level evidence snippets such as titles, brands, bullets, descriptions, and, when available, review snippets; each snippet has an evidence ID so that attribution claims can be checked against the released pool. 
For paper-facing presentation, hard constraints and soft preferences are merged into a single preference view, while the released JSON preserves the original fields for reproducibility. 
Fig.~\ref{fig:benchmark_worked_example} in Appx.~\ref{app:worked_example} illustrates one concrete PACEShop record; full persona and evidence JSON examples are provided in the same appendix section.

\begin{figure*}[h]
\centering
\fbox{%
\begin{minipage}{0.44\textwidth}
\small
\textbf{Input} \\[0.5em]
\textbf{Query:} \textit{tactical duffle gray} \\[0.6em]
\textbf{Persona} (structured JSON object): \\[0.4em]
\textit{profile:} household = family with school-age kids \\[0.3em]
\textit{preferences:} avoid peanuts; quiet operation; \\
\quad easy to clean; durable for daily use \\[0.3em]
\textit{brand\_likes:} [Ecco Press, Lansinoh] \\[0.3em]
\textit{shopping:} budget = mid-range; quality = balanced; \\
\quad shipping = standard \\[0.3em]
\textit{history:} purchases = [Barack Before Obama\ldots, \\
\quad\quad Lansinoh Storage Bags\ldots] \\[0.2em]
\quad recent\_searches = [\ldots alternatives, \ldots upgrades] \\[0.3em]
\textit{context:} \textit{``family with school-age kids; mid-range budget; balanced quality; recent purchases include Barack Before Obama, Lansinoh Storage Bags.''}
\end{minipage}}
\hfill
\fbox{%
\begin{minipage}{0.50\textwidth}
\small
\textbf{Output} \\[0.5em]
\textbf{Overview:} ``For a durable, travel-friendly gray tactical duffle that handles daily family use, explore \textbf{Tactical}, \textbf{Gray Gym Bags}, \textbf{Rolling Duffles}, \textbf{Travel Duffles}, and \textbf{Range Bags}\ldots'' \\[0.5em]
\textbf{Category} (4--5): Tactical; Gray Gym Bags; \\
\quad Rolling Duffles; Travel Duffles; Range Bags \\[0.5em]
\textbf{Related queries} (5): \\
\quad tactical duffle with MOLLE webbing; \\
\quad gray duffel bag water resistant; \\
\quad wheeled tactical duffel carry-on; \\
\quad packable lightweight gray duffle; \\
\quad tactical range bag gunmetal grey \\[0.5em]
\textbf{Attributes} (8 cited products): \\
\quad Nexpak 26'' Tactical Duffle Military Molle; \\
\quad East West U.S.A Tactical Outdoor Duffel; \\
\quad Under Armour Contain 4.0 Duffle Bag; \\
\quad Gonex 80L Packable Travel Duffle; \\
\quad Samsonite Andante 2 Wheeled Rolling Duffel; \\
\quad OSAGE RIVER Tactical Range Bag; \\
\quad Amazon Basics Large Travel Duffel; \\
\quad Vertx Contingency Duffel 85L
\end{minipage}}
\caption{Worked example of a \BenchmarkName{} record (\texttt{full-good-00000}). Each example pairs a shopping query with a
structured persona, an auditable evidence pool, and a four-field
shopping response. The input pairs a shopping query with a structured persona object; the output is a heterogeneous response with four evaluation-facing components. Correctness depends jointly on the persona, all components, and the cited evidence. Attributes are shown as short product descriptions; the underlying JSON carries evidence IDs. BAD records additionally expose gold defect family
and location labels, enabling diagnostic evaluation rather than only scalar scoring.}
\label{fig:benchmark_worked_example}
\end{figure*}

\paragraph{Worked Example of a Dataset Record.}
The boxed schema below summarizes the abstract \BenchmarkName{} record
schema: every record pairs a query with a structured persona, an
auditable evidence pool, and a four-field shopping response, and BAD
records additionally expose gold defect family and location labels.
Fig.~\ref{fig:benchmark_worked_example} then shows a single concrete
record end to end: the persona-conditioned query on the left, the
structured four-field response on the right. Correctness on this
record requires joint judgements -- persona consistency, cross-field
coherence, evidence support, and schema validity -- which the four
\EvaluationTarget{} targets operationalize.


\onecolumn
\subsection{Concrete Persona-JSON Input Example}
\label{app:concrete_persona}

The \texttt{[PERSONA\_JSON]} slot of the generation prompt
(Sec.~\ref{app:benchmark_prompt}) receives the full persona record,
including household profile, preferences (merged hard + soft), brand
affinities, shopping context, purchase history, recent searches, and
a natural-language context summary. Below is a representative example
(truncated for space; the published JSON preserves
\texttt{hard\_constraints} and \texttt{soft\_preferences} as separate
lists for reproducibility):

{\footnotesize
\begin{verbatim}
{
  "persona_id": "persona-0045",
  "profile": {
    "household": "family with school-age kids",
    "demographics": "balances durability and value"
  },
  "preferences": {
    "hard_constraints": ["avoid peanuts in ingredients",
                         "quiet operation required"],
    "soft_preferences": ["easy to clean",
                         "durable enough for daily use"],
    "brand_likes": ["Ecco Press", "Lansinoh"],
    "brand_avoids": [],
    "attribute_preferences": ["travel friendly",
                              "low-maintenance finishes"]
  },
  "shopping_context": {
    "budget": "mid-range",
    "quality_sensitivity": "balanced",
    "shipping_urgency": "standard shipping"
  },
  "history": {
    "purchase_history": [
      {"asin": "0063028743",
       "product_name": "Barack Before Obama: Life Before
                        the Presidency",
       "brand": "Ecco Press"},
      {"asin": "B006XISCNA",
       "product_name": "Lansinoh Breastmilk Storage Bags,
                        100 Count",
       "brand": "Lansinoh"}
    ],
    "recent_searches": [
      "Barack Before Obama alternatives",
      "Lansinoh upgrades"
    ]
  },
  "context_summary": "This shopper is part of a family with
    school-age kids who balances durability and value on a
    mid-range budget. They prioritize travel-friendly,
    low-maintenance products..."
}
\end{verbatim}
}
\twocolumn

\onecolumn
\subsection{Concrete Evidence-JSON Input Example}
\label{app:concrete_evidence}

The \texttt{[EVIDENCE\_JSON]} slot receives up to 24 evidence snippets
drawn from the public evidence catalog. Each snippet carries a product
identifier, title, brand, a unique evidence ID, the text snippet, and
the source field. Below is a representative excerpt; paper-facing
tables show these as short product descriptions (``Nexpak 26'' Tactical
Duffle'') rather than raw IDs.

{\footnotesize
\begin{verbatim}
[
  {"canonical_product_id": "B079YYGMR3",
   "title": "26\" Tactical Duffle Military Molle Gear
             Shoulder Strap Range Bag TF126 GMG
             Gunmetal Grey",
   "brand": "Nexpak",
   "evidence_id": "M:B079YYGMR3:0",
   "snippet": "26\" Tactical Duffle Military Molle Gear
               Shoulder Strap Range Bag TF126 GMG
               Gunmetal Grey",
   "source_field": "product_title"},
  {"canonical_product_id": "B079YYGMR3",
   "title": "26\" Tactical Duffle Military Molle Gear ...",
   "brand": "Nexpak",
   "evidence_id": "M:B079YYGMR3:1",
   "snippet": "Nexpak",
   "source_field": "product_brand"},
  {"canonical_product_id": "B077KKMFFN",
   "title": "Under Armour Adult Contain 4.0 Duffle Bag,
             Graphite Medium Heat (040)/Black,
             One Size Fits All",
   "brand": "Under Armour",
   "evidence_id": "M:B077KKMFFN:0",
   "snippet": "Under Armour Adult Contain 4.0 Duffle Bag,
               Graphite Medium Heat (040)/Black,
               One Size Fits All",
   "source_field": "product_title"}
]
\end{verbatim}
}
\twocolumn

\onecolumn
\subsection{\BenchmarkName{} Benchmark record Generation Prompt}
\label{app:benchmark_prompt}

\BenchmarkName{} construction has two stages, and only the first stage uses an LLM. \emph{GOOD records} are generated by a fixed LLM (Claude Opus 4.6) under the constrained response-generation prompt reproduced below. \emph{BAD records} are then derived from each accepted GOOD record by deterministic rule-based defect injection in Python rather than by any second LLM prompt; this design choice is what guarantees that every BAD record carries an auditable gold defect family and gold defect location label by construction. We document both stages in this subsection so the appendix is self-contained.

\paragraph{GOOD generation: LLM under a constrained JSON contract.}
For each query--persona pair, the generator is invoked once with a constrained JSON contract. The prompt enforces the four-field response schema (overview, bolded category list, related queries, evidence attributions), the cardinality and length budgets used in main-text \S\ref{sec:benchmark_construction}, and a strict no-extras rule on the list fields. Crucially, the generator is never told what defect families exist: it only sees the persona and the candidate evidence pool, so accepted GOOD responses are contract-valid anchors rather than negative examples for the judge to learn against. The prompt's slot syntax (\texttt{[ASSIGNMENT\_ID]}, \texttt{[QUERY]}, \texttt{[PERSONA\_JSON]}, \texttt{[GROUP\_JSON]}, \texttt{[EVIDENCE\_JSON]}) matches the released runner.

{\footnotesize
\begin{verbatim}
You are generating a PACEShop V2 benchmark response.
Return JSON only with keys: overview_text, bold_categories,
productCategoryList, category_search_queries, rqList, attribution_list.

Requirements:
- Overview <= 350 chars.
- Include exactly 4 or 5 bold categories using **...** in overview_text.
- productCategoryList must be an array of plain strings aligned with
  bold categories.
- Each category label <= 15 chars.
- rqList must contain exactly 5 short related shopping refinements.
- attribution_list must be an array of plain evidence ID strings only.
- category_search_queries must be an array of plain strings aligned 1:1
  with productCategoryList.
- Do not return products, scores, or citation objects inside list fields.
- Keep the response personalized to the provided persona and compatible
  with the query intent.

Assignment: [ASSIGNMENT_ID]
Query: [QUERY]
Persona JSON: [PERSONA_JSON]
Support groups: [GROUP_JSON]
Candidate evidence: [EVIDENCE_JSON]
\end{verbatim}
}

\noindent
Generated GOOD candidates pass through deterministic normalization and the L1 12-check validator (\S\ref{app:validity_stack}) before any BAD variant is derived; candidates that fail hard schema, evidence-ID, or XML round-trip checks are rejected.

\paragraph{BAD construction: deterministic rule-based defect injection.}
We deliberately do \emph{not} use an LLM prompt to generate BAD records. Each BAD record is produced by a deterministic Python edit on a validated GOOD record drawn from the seven-family defect taxonomy in Tab.~\ref{tab:defect_taxonomy_generated} (Sec.~\ref{app:defect_taxonomy}). For every defect family $d \in \mathcal{D}$ the edit is fixed in advance, so the gold defect family $d$ and gold defect location $r \in \mathcal{R}$ are known by construction rather than being predicted by another model:
\begin{itemize}[leftmargin=1.2em,itemsep=2pt,topsep=2pt]
\item \textsc{PREF\_CONFLICT}: replace the first product category with a label that contradicts the persona's first hard constraint (e.g., \emph{leather handbags} for a persona with \texttt{avoid\_leather}). Location: \texttt{productCategoryList[0]}.
\item \textsc{OVER\_PERSONALIZATION\_HALLUCINATION}: prepend a fabricated persona-history reference (e.g., ``Given their Dyson Airwrap, \ldots'') to the overview. Location: \texttt{overview\_text}.
\item \textsc{XCOMP\_MISMATCH}: replace the related-query list with the related-query list of an unrelated GOOD record. Location: \texttt{rqList}.
\item \textsc{INTENT\_DRIFT}: replace the first product category with an unrelated category drawn from a fixed off-intent pool. Location: \texttt{productCategoryList[0]}.
\item \textsc{REDUNDANCY}: duplicate the first related query into the second slot. Location: \texttt{rqList[1]}.
\item \textsc{EVIDENCE\_MISMATCH}: replace the attribution list with the attribution list of an unrelated GOOD record. Location: \texttt{attribution\_list}.
\item \textsc{UNSUPPORTED\_CLAIM}: append a fabricated product-feature claim (e.g., ``It also highlights solar charging support.'') to the overview. Location: \texttt{overview\_text}.
\end{itemize}

\noindent
Multi-defect variants (tiers 2--4) are produced by composing edits from distinct response-field groups (\emph{categories}, \emph{related queries}, \emph{attributions}, \emph{overview}) under the canonical group order \emph{categories $\rightarrow$ related queries $\rightarrow$ attributions $\rightarrow$ overview}, which avoids interactions between overview-modifying edits and category-modifying edits (e.g., bolded-category alignment). Combinations spanning two defects from the same field group are disallowed; the full set of admissible single- and multi-defect combinations is enumerated in \texttt{pmsb.faults} of the released benchmark code.

Because every BAD edit is a deterministic function of (i)~the source GOOD record, (ii)~the assigned persona, and (iii)~the defect family, BAD records do not require a separate LLM-judging-style validator on top of the L1 12-check audit. The injection produces both the modified four-field response and the gold location string in a single step, so any disagreement between a judge's predicted (defect family, defect location) and the gold pair in Tabs.~\ref{tab:sota_failure_main} and~\ref{tab:sota_failure_main_v4} is unambiguously a judge error rather than a labeling error.
\twocolumn

\section{Prior Evaluation Protocols and Baseline Adaptation}
\label{app:prior_protocols_adaptation}

This appendix documents how prior personalization, RAG, and judging
protocols relate to the \BenchmarkName{} evaluation schema.
We first describe, in a single unified support table, which scenarios each prior judging protocol natively covers and which scenarios each running baseline scores or abstains on (Sec.~\ref{app:eval_baselines}). We then audit target leakage for the three single-target adapted baselines on the same held-out test set (Sec.~\ref{app:target_leakage_audit}). Finally, we list the baseline prompt templates run on the shared backbone pool (Sec.~\ref{app:baseline_prompts}). Together these subsections explain why the ``--'' cells in the main result tables are deliberate and measurement-backed, not notational laziness.

\subsection{Native and Adapted Scenario Support}
\label{app:eval_baselines}

Prior personalization, RAG, and judging protocols differ in what they natively evaluate.
A personalization benchmark may test whether a model adapts to user history without defining defect-family labels; a RAG evaluator may check product-evidence faithfulness without evaluating invented shopper history; and a scalar judge may estimate overall quality without localizing the failed response field.
Tab.~\ref{tab:baseline_scenario_support} summarizes this in two blocks: the upper block lists prior protocols that we do \emph{not} directly run as \BenchmarkName{} judges and marks the targets they natively cover with ``$\partial$'' or ``--''; the lower block lists the baselines we \emph{do} run on the shared backbone pool used in Tab.~\ref{tab:sota_failure_main} -- four multi-target rubric ablations of the \JudgeName{} output schema and three single-target adapted baselines (ARES-/PersonaLens-/EtaPP-style) that score only their native target and report ``--'' elsewhere.
Unsupported scenarios are reported as ``--'' rather than forced into the \BenchmarkName{} output schema.

\begin{table*}[t]
\centering
\scriptsize
\setlength{\tabcolsep}{4pt}
\renewcommand{\arraystretch}{1.10}
\resizebox{\textwidth}{!}{%
\begin{tabular}{l l c c c c l}
\toprule
\textbf{Method / native protocol} & \textbf{Native signal or role in study} & \textbf{S1 P-source} & \textbf{S2 C detect+loc} & \textbf{S3 E product/history} & \textbf{S4 A family+loc} & \textbf{Why this row scores or abstains} \\
\midrule
\multicolumn{7}{l}{\emph{Prior protocols not directly run as \BenchmarkName{} judges (reference)}} \\
LaMP \cite{salemi2023lamp} & Task metrics for personalized generation & $\partial$ & -- & -- & -- & No unchanged-response persona swap or defect-source label. \\
PersonaLens \cite{zhao2025personalens} & LLM user and judging agents for assistants & $\partial$ & -- & -- & -- & Evaluates user-conditioned success, not four-field response diagnosis. \\
PrefEval \cite{zhao2025prefeval} & Preference following/classification & $\partial$ & -- & -- & -- & Persona signal is present, but no evidence pool or response-field location. \\
PersonaMem \cite{jiang2025personamem} & Response selection over user histories & $\partial$ & -- & -- & -- & History use is measured, not defect family and location. \\
LikeBench \cite{rahman2025likebench} & Simulated-user likability diagnostics & $\partial$ & -- & -- & -- & Subjective user-fit signal; no \BenchmarkName{} grounding or localization contract. \\
EtaPP \cite{hao2025etapp} & Key-point judging for personalized tool use & $\partial$ & -- & -- & $\partial$ & Checks personalized task points, but not \BenchmarkName{} source-family/location labels. \\
Personalized judge \cite{dong2024personalizedjudge} & User-conditioned judging preference & $\partial$ & -- & -- & -- & Personalizes a judging preference, but does not define multi-field defect localization. \\
ECom-Bench \cite{wang2025ecombench} & Task-success trajectories in e-commerce support & $\partial$ & -- & -- & -- & Realistic commerce tasks, but not fixed-response judge diagnosis. \\
Search Arena \cite{miroyan2025searcharena} & Human preferences for search-augmented chat & -- & -- & $\partial$ & -- & Search/citation quality is relevant, but no persona-history grounding label. \\
ARES \cite{saadfalcon2024ares} & RAG faithfulness to retrieved evidence & -- & -- & $\partial$ & -- & Covers product-evidence grounding only, not persona or multi-field diagnosis. \\
G-Eval / MT-Bench \cite{liu2023geval,zheng2023judging} & Scalar or pairwise quality judgments & -- & -- & -- & -- & Runnable only after adapting the prompt to emit the \BenchmarkName{} schema (see below). \\
\midrule
\multicolumn{7}{l}{\emph{Running baselines on the shared backbone pool (used in Tab.~\ref{tab:sota_failure_main})}} \\
G-Eval-style rubric \cite{liu2023geval} & Multi-target rubric ablation, persona-aware & \cmark & \cmark & \cmark & \cmark & Adapted to the \JudgeName{} output schema; differs only in instruction. \\
MT-Bench-style scalar \cite{zheng2023judging} & Multi-target rubric ablation, scalar-first & \cmark & \cmark & \cmark & \cmark & Adapted to the \JudgeName{} output schema; back-fills structured fields after a scalar judgment. \\
Persona-blind & Multi-target rubric ablation, persona withheld & -- & \cmark & \cmark & \cmark & Persona payload replaced by \texttt{\{"omitted": true\}}; isolates the marginal value of persona conditioning. \\
Persona-aware generic & Multi-target rubric ablation, persona-aware & \cmark & \cmark & \cmark & \cmark & Shared protocol for the backbone comparison without explicit counterfactual reasoning. \\
ARES-style & Single-target adapted: evidence (E) only & -- & -- & \cmark & -- & Native protocol scores RAG faithfulness only; non-E targets reported as ``--''. \\
PersonaLens-style & Single-target adapted: persona (P) only & \cmark & -- & -- & -- & Native protocol scores personalization only; non-P targets reported as ``--''. \\
EtaPP-style & Single-target adapted: actionable (A) only & -- & -- & -- & \cmark & Native protocol scores key-point matches only; non-A targets reported as ``--''. \\
\midrule
\textbf{\JudgeName{} (ours)} & \textbf{Structured \EvaluationTarget{} judging contract} & \textbf{\cmark} & \textbf{\cmark} & \textbf{\cmark} & \textbf{\cmark} & \textbf{Verdict, target scores, defect family, location, and evidence IDs.} \\
\bottomrule
\end{tabular}%
}
\caption{Native and adapted support for the four \EvaluationTarget{} target scenarios. The first block lists prior protocols that are \emph{not directly run} as \BenchmarkName{} judges; ``$\partial$'' marks a related signal that does not define the \BenchmarkName{} closure metric or the required diagnostic output fields. The second block lists the baselines we \emph{do run} on the shared backbone pool in Tab.~\ref{tab:sota_failure_main}: four multi-target rubric ablations of the \JudgeName{} output schema and three single-target adapted baselines (ARES-/PersonaLens-/EtaPP-style) that score only their native target and report ``--'' on the others. Tab.~\ref{tab:baseline_protocol_target_map} expands the native protocol mapping for the single-target adapted baselines, and Tab.~\ref{tab:evaluation_target_leakage_audit} audits their off-target leakage.}
\label{tab:baseline_scenario_support}
\end{table*}

\subsection{Native Judging-Protocol Map and Evaluation-Target Leakage Audit}
\label{app:target_leakage_audit}

The strict paired-counterfactual main table
(Tab.~\ref{tab:sota_failure_main_v4}) is supported by two audit
tables kept here rather than in the main text. The native
judging-protocol map (Tab.~\ref{tab:baseline_protocol_target_map})
records, for each adapted single-target baseline, which
\EvaluationTarget{} target its native judging protocol covers and why the
other targets are reported as ``--'' rather than over-adapted. The
target-leakage audit (Tab.~\ref{tab:evaluation_target_leakage_audit}) then
measures, on the same 3{,}400-record held-out test set, the rate at which each adapted
baseline emits a defect-family label outside its native target
(e.g.,~ARES-style firing \textsc{PREF\_CONFLICT} on a record whose
gold defect is compositional). ARES-style and PersonaLens-style stay
under $1$--$2\%$ on the \emph{Any off-target} aggregate, validating
their honest restriction in Tab.~\ref{tab:sota_failure_main_v4} even
though they show non-trivial target leakage on individual P/C cells;
EtaPP-style leaks heavily (99\% any-off-target), and we therefore do
not interpret its S4 cell as evidence that EtaPP-style covers more
than the A target. Together these two tables protect
Tab.~\ref{tab:sota_failure_main_v4} from the failure mode of
``apply a prompt, get a number on every cell'': the ``--''
restrictions are deliberate and supported by measurement, not
notational laziness.

\begin{table*}[t]
\centering
\scriptsize
\setlength{\tabcolsep}{4pt}
\renewcommand{\arraystretch}{1.12}
\resizebox{\textwidth}{!}{%
\begin{tabular}{p{0.18\linewidth} p{0.30\linewidth} c c p{0.30\linewidth}}
\toprule
\textbf{Baseline} & \textbf{Native judging protocol} & \textbf{\EvaluationTarget{} target} & \textbf{Supported scenario cell} & \textbf{Why other \EvaluationTarget{} targets are ``--''} \\
\midrule
ARES-style \cite{saadfalcon2024ares} & RAG faithfulness: 3 LM-judges (context relevance, answer faithfulness, answer relevance) & E & S3$_{\text{E}}$ only & Native judging protocol explicitly does not score persona axes or cross-field consistency; non-E defect families are not emitted. \\
PersonaLens-style \cite{zhao2025personalens} & 3-axis judge: personalization, response quality, task success & P & S1$_{\text{P}}$ only & Native rubric has no slot for cross-field consistency or evidence-pool faithfulness; family vocabulary covers only persona conflicts. \\
EtaPP-style \cite{hao2025etapp} & Key-point judging: $\{$defect family, defect location$\}$ as the gold key-points & A & S4$_{\text{A}}$ only & Target scores are fixed at 3.0; consequently the target-triage conjunct is rarely satisfied even on the native target. \\
Personalized judge \cite{dong2024personalizedjudge} & Binary preference judge measured by judge-vs-human agreement & P (binary) & -- & No defect family / location / target-score output; the judging protocol is judge-vs-human pairwise agreement, not pointwise diagnosis. \\
Search Arena \cite{miroyan2025searcharena} & Pairwise human votes on search-augmented chat & E (partial) & -- & Captures search/citation quality, but no persona-history label and no defect family. \\
PrefEval \cite{zhao2025prefeval} & Preference-following accuracy / classification across long context & P & -- & No persona-swap counterfactual label, and no multi-field response to localize. \\
PersonaMem \cite{jiang2025personamem} & MCQ accuracy over an evolving user history & P & -- & No structured response, and no native defect-family taxonomy. \\
LikeBench \cite{rahman2025likebench} & Per-turn likability dimensions from simulated users & P & -- & Subjective likability dimensions; no structured response and no grounding contract. \\
LaMP \cite{salemi2023lamp} & Task metrics (accuracy / ROUGE) on user-conditioned generation & P & -- & Single-text generation; no four-field response and no defect labels. \\
\bottomrule
\end{tabular}%
}
\caption{\textbf{Native judging-protocol to \EvaluationTarget{} target mapping.}  Each adapted baseline is restricted to the \EvaluationTarget{} target that its native judging protocol scores; unsupported targets are reported as ``--''.  Table~\ref{tab:sota_failure_main_v4} uses this mapping when reporting scenario scores: prior protocols cover at most one target each, so the joint S1--S4 PACE average is undefined for them.  Only \JudgeName{} has an output contract supporting all four \EvaluationTarget{} targets.}
\label{tab:baseline_protocol_target_map}
\end{table*}

\begin{table*}[h]
\centering
\scriptsize
\setlength{\tabcolsep}{4pt}
\renewcommand{\arraystretch}{1.10}
\begin{tabular}{l cccc}
\toprule
\textbf{Adapted baseline} & \textbf{P off-target} & \textbf{C off-target} & \textbf{E off-target} & \textbf{Any off-target} \\
\midrule
\multicolumn{5}{l}{\emph{Off-target firing rate: P/C/E columns are computed on records whose gold target is not the baseline's native target.  Lower is better; 0 means perfect abstention.}} \\
ARES-style \cite{saadfalcon2024ares} (native: E) & 23\% & 22\% & \emph{native} & 2\% \\
PersonaLens-style \cite{zhao2025personalens} (native: P) & \emph{native} & 20\% & 1\% & 0\% \\
EtaPP-style \cite{hao2025etapp} (native: A) & 0\% & 0\% & 0\% & 99\% \\
\midrule
\multicolumn{5}{l}{\emph{Reference: \JudgeName{} supports all \EvaluationTarget{} targets natively (so the relevant rate is family misclassification, not leakage)}} \\
\textbf{\JudgeName{}} (native: P+A+C+E) & 7\% & 14\% & 27\% & n/a \\
\bottomrule
\end{tabular}
\caption{\textbf{Evaluation-target leakage audit.}  For each adapted single-target baseline (Sonnet 4.6 on the 3{,}400-record held-out test set), this table reports the rate at which the baseline emits a defect-family label outside its native \EvaluationTarget{} target.  A genuine single-target baseline should score near 0\% on every non-native target; high leakage means the prompt is over-adapted.  The \JudgeName{} reference row reports family misclassification rather than leakage, since \JudgeName{} supports all \EvaluationTarget{} targets.  This audit is the safeguard against ``apply a prompt, get a number on every cell'': cells in Table~\ref{tab:sota_failure_main_v4} are non-zero for a single-target baseline only on its native target.}
\label{tab:evaluation_target_leakage_audit}
\end{table*}

\subsection{Baseline Prompt Templates}
\label{app:baseline_prompts}

We organize the baselines into two groups. The first group is a set of \emph{multi-target rubric ablations} that share the \JudgeName{} input bundle (query, persona JSON, response XML, structured response JSON, candidate evidence IDs, candidate evidence preview, deterministic checks) and the same structured output contract, differing only in \emph{instruction} and \emph{persona visibility}. The second group is a set of \emph{single-target adapted baselines} (ARES-style, PersonaLens-style, EtaPP-style) whose native judging protocols cover only one \EvaluationTarget{} target each; we adapt them faithfully on their native target and abstain (``--'') on the other targets, as documented in Tab.~\ref{tab:baseline_protocol_target_map} and audited in Tab.~\ref{tab:evaluation_target_leakage_audit}.

\subsubsection{Multi-target rubric ablations}
\label{app:baseline_prompts_rubric}

The rubric ablations all emit the full \JudgeName{} output schema (verdict, P/C/E/F axis scores, defect family, defect location, supporting evidence IDs, confidence, rationale) so that metrics across rows of Tab.~\ref{tab:sota_failure_main} differ only by the prompt instruction.

\myparagraph{G-Eval-style rubric judge \cite{liu2023geval}.}
``You are a rubric-guided judge in the style of a generic high-quality LLM evaluator. Score the response using a compact rubric over relevance, helpfulness, coherence, grounding, and structure. Then translate that rubric judgment into the required output schema.''

\myparagraph{Scalar-first judge (MT-Bench style) \cite{zheng2023judging}.}
``You are a scalar-first judge. Base the decision mainly on one overall quality impression, then back-fill the required structured fields in a best-effort way.''

\myparagraph{Persona-blind ablation.}
``You are a generic shopping-response judge. The persona is intentionally withheld for this baseline. Judge only whether the response seems broadly useful, coherent, grounded, and structurally valid.'' The persona payload is serialized as \texttt{\{"omitted": true\}}.

\myparagraph{Persona-aware generic rubric.}
``You are a persona-aware but otherwise generic shopping-response judge. Use the persona and query to judge whether the response is suitable for the user. Do not use explicit counterfactual swap reasoning and do not invent additional defect classes beyond the provided labels.''

\subsubsection{Single-target adapted baselines}
\label{app:baseline_prompts_adapted}

ARES-style, PersonaLens-style, and EtaPP-style do not share the \JudgeName{} four-target output schema in their native form. We therefore adapt each one only on the \EvaluationTarget{} target that its native judging protocol scores, leaving cells outside the native target as ``--'' in Tab.~\ref{tab:sota_failure_main} and Tab.~\ref{tab:sota_failure_main_v4}. To keep the adaptation auditable, all three adapted baselines run on the same 3{,}400-record held-out test set and the same shared backbone pool as the rubric ablations; only the \emph{native-target} cells are scored against the \JudgeName{} schema, and an evaluation-target leakage audit (Tab.~\ref{tab:evaluation_target_leakage_audit}) measures the rate at which each baseline emits a defect family outside its native target.

\myparagraph{ARES-style grounding judge \cite{saadfalcon2024ares}.}
The native ARES protocol is a RAG-faithfulness judge built from three lightweight LM scorers: \emph{context relevance}, \emph{answer faithfulness}, and \emph{answer relevance}. 
We instantiate this protocol on \BenchmarkName{} by mapping the three ARES axes onto $(q, y, E)$: context relevance asks whether the candidate evidence IDs in $E$ are related to $q$; answer faithfulness asks whether claims in $y$ are supported by $E$, including no fabricated evidence IDs and no invented persona-history claims; answer relevance asks whether $y$ answers $q$. 
The judge averages the three sub-judgments into the evidence-grounding axis $s_{\mathrm{E}} \in [1, 5]$ and emits $\hat{\ell}=\textsc{BAD}$ only when $s_{\mathrm{E}} < 3.0$. 
When BAD, the predicted defect family $\hat{d}$ is restricted to evidence-target labels
\{\texttt{\footnotesize EVIDENCE\_\allowbreak MISMATCH}, 
\texttt{\footnotesize OVER\_\allowbreak PERSONALIZATION\_\allowbreak HALLUCINATION}, 
\texttt{\footnotesize UNSUPPORTED\_\allowbreak CLAIM}\}, 
and $\hat{r}$ is the response field carrying the unsupported claim or invalid attribution. 
For any non-evidence defect, the prompt explicitly returns \texttt{\footnotesize NONE}/\texttt{\footnotesize NONE} rather than fabricating a label. 
The persona, compositional, and format/safety axes $\{s_{\mathrm{P}}, s_{\mathrm{C}}, s_{\mathrm{F}}\}$ are pinned to a neutral $3.0$ because ARES does not natively score them, and supporting evidence IDs are constrained to $\hat{S} \subseteq E$. 
Consequently, this baseline reports a measured number on the S3 column of Tab.~\ref{tab:sota_failure_main} and S3$_{\text{E}}$ in Tab.~\ref{tab:sota_failure_main_v4}, and reports ``--'' on S1, S2, and S4. 
The leakage audit in Tab.~\ref{tab:evaluation_target_leakage_audit} confirms that this restriction is honored in practice: ARES-style fires a non-evidence defect family on under $2\%$ of off-target records on the \emph{Any off-target} aggregate.

\myparagraph{PersonaLens-style personalization judge \cite{zhao2025personalens}.}
The native PersonaLens judging protocol is a three-axis judge that scores \emph{personalization}, \emph{response quality}, and \emph{task success} for user-conditioned generation. 
We adapt it to \BenchmarkName{} by walking through the structured persona one slot at a time---hard constraints, brand likes/avoids, attribute preferences, and budget---and asking the judge to mark each slot as honored, contradicted, or unconstrained by the response fields; this drives the persona-alignment axis $s_{\mathrm{P}} \in [1, 5]$. 
Response quality scores the format/safety axis $s_{\mathrm{F}}$, and task success is folded into the verdict via the threshold $s_{\mathrm{P}} < 3.0$. 
When BAD, the predicted defect family $\hat{d}$ is restricted to persona-target labels: 
\texttt{\footnotesize PREF\_\allowbreak CONFLICT} for an explicit persona conflict, e.g., the response bolds a brand the persona avoids, or 
\texttt{\footnotesize OVER\_\allowbreak PERSONALIZATION\_\allowbreak HALLUCINATION} when the response invents a purchase or search not in the persona history; $\hat{r}$ is the response field carrying the conflict. 
The compositional and evidence axes $\{s_{\mathrm{C}}, s_{\mathrm{E}}\}$ are pinned at $3.0$, $\hat{S}$ may be empty, and any non-persona defect is reported as \texttt{\footnotesize NONE}/\texttt{\footnotesize NONE}. 
This baseline therefore contributes a measured number on the S1 column, S1$_{\text{P}}$ in Tab.~\ref{tab:sota_failure_main_v4}, and abstains elsewhere. 
The evaluation-target leakage audit in Tab.~\ref{tab:evaluation_target_leakage_audit} shows that PersonaLens-style fires a non-persona family on under $1\%$ of off-target records on the \emph{Any off-target} aggregate.

\myparagraph{EtaPP-style actionable key-point judge \cite{hao2025etapp}.}
The native EtaPP judging protocol judges a personalized action trace against human-annotated \emph{key-points} that the trace must satisfy. We adapt this to \BenchmarkName{} by treating the gold defect family $d$ and the gold defect location $r$ as the two key-points the judge must name: the candidate response $y$ replaces the action trace, the deterministic checks plus a brief read of the four response fields drive the GOOD/BAD verdict, and when BAD the judge proposes both $\hat{d} \in \mathcal{D}$ and $\hat{r} \in \mathcal{R}$ from the full label space. Following the EtaPP convention, the target score on the predicted target is fixed at $3.0$ rather than emitted by the judge; $s_{\mathrm{P}}$, $s_{\mathrm{C}}$, $s_{\mathrm{E}}$ are likewise pinned at $3.0$, while $s_{\mathrm{F}}$ is scored honestly from schema validity and $\hat{S}$ may be empty. Because EtaPP's native judging protocol does not score persona, compositional, or evidence-grounding targets separately, this baseline reports measured S4 family/location numbers in Tab.~\ref{tab:sota_failure_main} (and S4$_{\text{A}}$ in Tab.~\ref{tab:sota_failure_main_v4}) and ``--'' on S1--S3. Two caveats follow from the audit tables. First, the fixed $3.0$ target score is why EtaPP-style fails the strict target-triage conjunct in Tab.~\ref{tab:sota_failure_main_v4} more often than its S4 family number alone would suggest. Second, the evaluation-target leakage audit (Tab.~\ref{tab:evaluation_target_leakage_audit}) shows that EtaPP-style emits \emph{some} defect family on $99\%$ of off-target records, so we do not interpret its S4 cell as evidence that EtaPP-style covers more than the A target.

The full Markdown prompt files for all baselines, including the three adapted single-target baselines, are shipped with the code release.

\FloatBarrier

\section{\JudgeName{} Evaluation}
\label{app:judge_evaluation}

This appendix expands the main-paper subsection on \emph{Coverage of the \EvaluationTarget{} Targets} (Sec.~\ref{sec:judge_coverage}) and the structured output contract used by \JudgeName{}.
It is organized as follows.
Sec.~\ref{app:pace_targets_closure} gives a per-target walkthrough mapping each \JudgeName{} output field to one of the four \EvaluationTarget{} targets and to the corresponding evaluation scenario.
Sec.~\ref{app:judge_dimensions} expands the four main-body axes into the fuller six-dimension rubric used during judge development.
Sec.~\ref{app:fault_localization} documents the structured fault-localization protocol that turns the judge from a scalar scorer into a diagnostic tool.
Readers who only need the headline mapping can rely on Tab.~\ref{tab:pace_operationalization} and the main-paper subsection; the subsections below are intended for readers who want the full coverage argument and the dimension-by-dimension protocol.

\subsection{How the Protocol Closes the Four \EvaluationTarget{} Targets}
\label{app:pace_targets_closure}

Tab.~\ref{tab:pace_operationalization} gives the high-level mapping from \EvaluationTarget{} targets to \BenchmarkName{} design choices, \JudgeName{} output fields, and scenario-level metrics. 
The role of \JudgeName{} is to make the corresponding diagnosis reportable: \BenchmarkName{} makes each target observable, while \JudgeName{} gives the judge explicit fields for reporting it.

\begin{itemize}[leftmargin=1.2em,itemsep=2pt,topsep=2pt]
\item \textbf{Personalized} $(s_{\mathrm{P}} \rightarrow S1)$. 
The persona-alignment score checks whether $y$ respects the shopper context in $p$, including constraints, preferences, brand affinities, budget, purchase history, and recent searches. 
This closes the Personalized target by forcing the verdict to condition on $p$ rather than treating persona as optional context. 
It directly supports the S1 persona-swap evaluation.

\item \textbf{Actionable} $(a_{\mathrm{A}} \rightarrow S4)$. 
The actionable diagnosis object requires the judge to report what failed and where: $\hat{d}$ identifies the defect family, $\hat{r}$ identifies the response field, $\hat{S}$ provides evidence support, $\hat{c}$ indicates confidence, and $\hat{z}$ explains the decision. 
This closes the Actionable target by turning the judge from a scalar scorer into a diagnostic tool. 
It directly supports the S4 defect-family and field-location evaluation.

\item \textbf{Compositional} $(s_{\mathrm{C}}, \hat{r} \rightarrow S2)$. 
The compositional-consistency score evaluates $y$ as a four-field object rather than as independent text. 
The overview should agree with the category list, related queries should refine the represented product facets rather than drift away from the intent, and attributions should support the claims they are attached to. 
Together with the location field $\hat{r}$, this closes the Compositional target and supports S2 cross-field detection and localization.

\item \textbf{Evidence-grounded} $(s_{\mathrm{E}}, \hat{S} \rightarrow S3)$. 
The evidence-grounding score verifies both product grounding and persona-history grounding. 
Product grounding requires attribution IDs to exist in $E$ and support the relevant claims. 
Persona-history grounding requires the response not to invent purchases, searches, or user-specific facts absent from $p$. 
Together with the supporting-evidence field $\hat{S}$, this closes the Evidence-grounded target and supports S3 product- and history-grounding evaluation.
\end{itemize}

The auxiliary format/safety score $s_{\mathrm{F}}$ checks whether $y$ satisfies the response contract, including count constraints, length limits, banned-string filters, evidence-ID formatting, and XML round-trip validity. 
This separates structural compliance from the four \EvaluationTarget{} targets: a response may be well-formed but semantically invalid, or semantically plausible but structurally unusable.

This output contract explains where scalar judges fail and where \JudgeName{} can be evaluated more precisely. 
A scalar judge may call a response ``mostly helpful,'' but it has no required slot for naming the broken field, the defect type, or the evidence supporting the diagnosis. 
\JudgeName{} exposes these missing fields directly: the P/C/E target scores identify which quality dimension degraded, the Actionable fields identify what failed and where, and the evidence-ID constraint prevents the judge from inventing its own support. 
As a result, the protocol can be evaluated not only by GOOD/BAD accuracy, but also by the scenario-aligned metrics in Sec.~\ref{sec:experiments}: persona-swap accuracy, cross-field localization, grounding detection, hallucinated-ID rate, and defect-family/location accuracy.

\subsection{Detailed \JudgeName{} Evaluation Dimensions}
\label{app:judge_dimensions}

The four main-body axes (\S\ref{sec:percojudge}) collapse a richer six-dimension protocol that we used during the judge development. Each dimension below states (i)~the challenge that motivates it, (ii)~how \JudgeName{} operationalizes it, and (iii)~the metrics used to measure it. Every dimension admits both deterministic checks and LLM-scored assessments; deterministic checks provide a cost-free lower bound, while LLM scoring captures semantic nuances that rules cannot.

\myparagraph{D1: Persona Alignment.}
\emph{Challenge.} The central gap in personalization evaluation is distinguishing responses that are \emph{generically helpful} from those that are \emph{correct for a specific user} \cite{zhao2025personalens,zhao2025prefeval,jiang2025personamem,rahman2025likebench}. A response can score highly on fluency and relevance yet still ignore the user's budget constraint, violate a hard preference, or recommend products misaligned with the user's shopping journey stage.
\emph{Approach.} \JudgeName{} evaluates persona alignment through \emph{counterfactual persona swaps} (S1 in \S\ref{sec:scenarios}): if swapping the persona changes which response dimensions should differ, the judge must detect that change; dimensions that should remain invariant under the swap must stay stable. Formally, let $A_{\mathrm{rel}}(q,p)$ denote dimensions that should change under a persona swap and $A_{\mathrm{inv}}(q,p)$ denote dimensions that should remain invariant. The persona-swap score is $S_{\mathrm{swap}} = \sum_{a \in A_{\mathrm{rel}}} w_a \Delta_a - \lambda \sum_{a \in A_{\mathrm{inv}}} w_a \Delta_a$, where $\Delta_a$ measures the distance between axis signatures of paired responses. We also define the alignment gap $G(q,p) = H_{\mathrm{gen}}(f(q,\varnothing)) - H_{\mathrm{pers}}(f(q,p))$, which quantifies the extent to which persona conditioning changes the evaluation outcome.
\emph{Metrics.} Persona-alignment score (1--5, LLM-scored); GOOD--BAD score gap on D1 (larger gap $=$ stronger persona discrimination). A \emph{negative} gap indicates the judge cannot distinguish persona-conditioned quality---the signature of a persona-blind evaluation.

\myparagraph{D2: Relevancy \& Helpfulness.}
\emph{Challenge.} A shopping assistant must match both explicit requirements (brand, price, features) and implicit expectations (use case, quality tier, journey stage) \cite{hao2025etapp,miroyan2025searcharena}. Generic ``helpfulness'' ratings conflate these with fluency.
\emph{Approach.} The judge scores each component on product relevancy, intent matching (explicit + implicit), and actionable decision support---whether the overview provides real buying advice rather than generic category listing.
\emph{Metrics.} Per-section helpfulness score (1--5); binary criteria for product relevancy, intent matching, and helpfulness.

\myparagraph{D3: Diversity \& Distinctiveness.}
\emph{Challenge.} Responses that are relevant but homogeneous---e.g., recommending five near-identical products or generating overlapping category groups---fail to help users compare options.
\emph{Approach.} The judge checks whether products, categories, and related queries offer sufficient variety in brands, price points, and features; whether category groups are non-overlapping and organized by customer need; and whether related queries span multiple types (comparison, recommendation).
\emph{Metrics.} Binary criteria for product diversity, type diversity, query specificity, and category distinctness; deterministic duplicate-detection checks.

\myparagraph{D4: Compositional Consistency.}
\emph{Challenge.} A structured four-field response is correct only if its fields are \emph{mutually coherent}: the overview should preview the categories, the categories should be refinable by the related queries, and the attributions should support the claims. Existing judges evaluate each field independently and miss cross-field failures.
\emph{Approach.} \JudgeName{} reasons over explicit cross-field constraints: overview--category alignment (bolded terms match category names), category--query coverage (queries refine the categories), evidence--claim grounding, diversity/non-redundancy, and intent preservation across all four fields. The objective is to judge the response as a structured object rather than scoring each field in isolation.
\emph{Metrics.} Component-consistency score (1--5, LLM-scored); deterministic highlight--category overlap check; GOOD--BAD score gap on D4.

\myparagraph{D5: Evidence Grounding.}
\emph{Challenge.} Shopping assistants must ground their claims in real product evidence. Hallucinated product features or invalid attribute IDs erode user trust. Evidence-only evaluators \cite{saadfalcon2024ares} address this but ignore whether the \emph{selected} evidence is appropriate for the specific user.
\emph{Approach.} The judge verifies that claims in the overview and categories are supported by the referenced evidence records, that attribute IDs resolve to valid entries in the evidence pool, and that no hallucinated evidence appears. Evidence-grounding assessment combines deterministic ID validation with LLM-based claim--evidence alignment.
\emph{Metrics.} Evidence-grounding score (1--5, LLM-scored); attribute-ID validity (deterministic); hallucinated-evidence rate (fraction of cited IDs absent from the evidence pool); GOOD--BAD score gap on D5.

\myparagraph{D6: Format \& Safety.}
\emph{Challenge.} Structural compliance (correct counts, lengths, valid XML) and trust/safety requirements (neutral tone, appropriate disclaimers, no banned content) are prerequisites for deployment but are often treated as afterthoughts in evaluation.
\emph{Approach.} Twelve deterministic validation checks (overview length, bold--category match, category and related-query counts and uniqueness, evidence-ID validity, XML round-trip, banned-string screening) form a cost-free baseline. LLM judges additionally assess language tone and score a format-safety axis.
\emph{Metrics.} Format-safety score (1--5, LLM-scored); 12-check deterministic pass rate; parse success rate; GOOD--BAD score gap on D6.

\myparagraph{Mapping D1--D6 to the four main-body axes.}
In the main-body table (\S\ref{sec:main_results}) we report four axes rather than six: D2 and D3 both contribute to the \emph{component\_consistency} axis (relevancy is assessed within each component, and diversity is checked across components), which keeps the output contract compact without losing the underlying signals. The full six-dimension rubric is used by the LLM in its free-text reasoning and is reflected back into the four axis scores.

\subsection{Fault Localization Protocol}
\label{app:fault_localization}

Beyond the quality dimensions, \JudgeName{} addresses a capability missing from existing evaluation frameworks: \emph{structured fault localization}. Rather than reducing evaluation to a scalar score, the judge predicts (i)~the \emph{defect family} (which of 7 failure modes occurred) and (ii)~the \emph{defect location} (which of 4 field groups contains the fault). This transforms the judge from a classifier into a \emph{diagnostic tool}: developers can identify not just that a response failed, but \emph{what} went wrong and \emph{where} to fix it.

The multi-tier defect design in \BenchmarkName{} creates a particularly demanding test: for multi-defect records, the judge must detect and localize faults across up to four field groups simultaneously. We measure localization via defect-type macro-F1 (7 families), defect-location exact match, and field-group match. For single-defect records, exact match is reported directly. For multi-defect records the gold label is a set of locations while the current protocol outputs one prediction, so we report a \emph{hit rate}---whether the predicted location appears in the gold set. Multi-location prediction is a natural extension (a single inference emitting a list rather than a scalar location).

\onecolumn
\subsection{\JudgeName{} Evaluation Protocol and Prompt}
\label{app:judge_prompt}

The \JudgeName{} evaluation protocol is implemented as a single training-free judging prompt that takes a \BenchmarkName{} record $(q, p, y, E)$ as typed input and returns the structured output defined in \S\ref{sec:percojudge}. To make the protocol fully reproducible, we reproduce the prompt verbatim in this subsection; this is the symmetric counterpart of the dataset-side response-generation prompt in Appx.~\ref{app:benchmark_prompt}, and it is the prompt that all \JudgeName{} rows in Tabs.~\ref{tab:sota_failure_main} and~\ref{tab:sota_failure_main_v4} use across every backbone.

\paragraph{Prompt structure.}
The prompt is parameterized by four slot groups, each contributing a different piece of the protocol:
\begin{enumerate}[leftmargin=1.2em,itemsep=2pt,topsep=2pt]
\item \textbf{Typed input bundle.} Ten named slots bind the record to evaluation-facing fields: \texttt{record\_id}, \texttt{split}, \texttt{keyword} (the query $q$), \texttt{persona\_json} (the structured persona $p$), \texttt{response\_xml} and \texttt{response\_struct\_json} (the candidate response $y$ in two views), \texttt{candidate\_evidence\_ids\_json} and \texttt{candidate\_evidence\_json} (the evidence pool $E$ as IDs and previews), \texttt{used\_evidence\_ids\_json} (the evidence subset cited during \BenchmarkName{} GOOD-response construction, used as a soft anchor rather than a label), and \texttt{deterministic\_checks\_json} (the L1 12-check validator output, exposed to the judge as objective anchors).
\item \textbf{Closed label sets.} The diagnostic vocabulary is pinned to two closed enumerations: \texttt{allowed\_defect\_labels} = the seven \BenchmarkName{} defect families plus \textsc{NONE}, and \texttt{allowed\_location\_labels} = the four response-field groups (\textit{overview}, \textit{categories}, \textit{related queries}, \textit{evidence attributions}) plus \textsc{NONE}. Free-form labels outside these enumerations are rejected at parse time rather than coerced.
\item \textbf{Decision rules.} Four explicit rules force the judge to commit to a verdict and an evidence-grounded diagnosis rather than collapse to a scalar score: (i)~mark GOOD only when the response is correct for the persona, coherent across components, grounded in $E$, and structurally valid; (ii)~use \textsc{NONE}/\textsc{NONE} for defect family/location only when no material defect is present; (iii)~\texttt{supporting\_evidence\_ids} must be a subset of \texttt{candidate\_evidence\_ids\_json} (no invented IDs); (iv)~\texttt{confidence} and \texttt{overall\_bad\_probability} must lie in $[0,1]$, while \texttt{overall\_quality} and the four axis scores must lie in $[1,5]$.
\item \textbf{Output-schema contract.} A JSON-Schema object (the \texttt{prompt\_contract\_json} slot) requires the verdict $\hat{\ell} \in \{\textsc{GOOD}, \textsc{BAD}\}$, the BAD probability $\hat{p}_{\mathrm{bad}} \in [0,1]$, the four axis scores $s_{\mathrm{P}}, s_{\mathrm{C}}, s_{\mathrm{E}}, s_{\mathrm{F}} \in [1,5]$, and the actionable diagnostic object $a_{\mathrm{A}} = (\hat{d}, \hat{r}, \hat{S}, \hat{c}, \hat{z})$ with $\hat{d} \in \mathcal{D} \cup \{\textsc{NONE}\}$, $\hat{r} \in \mathcal{R} \cup \{\textsc{NONE}\}$, and $\hat{S} \subseteq E$.
\end{enumerate}

\paragraph{\JudgeName{} prompt.}
The full prompt as run on every backbone is reproduced below. Slots in curly braces are filled per-record by the runner; the closed label sets and the JSON-Schema contract are the same on every record.

{\footnotesize
\begin{verbatim}
You are LLMaJ for persona-conditioned multi-component shopping
assistant evaluation.

Judge the candidate response as a structured object. Use the
deterministic checks as objective anchors, but do not stop there.
Reason about:
- persona alignment
- cross-component consistency
- evidence grounding
- format safety

Allowed defect labels: {allowed_defect_labels}
Allowed defect locations: {allowed_location_labels}

Input bundle
- Record ID: {record_id}
- Split: {split}
- Query: {keyword}
- Persona JSON:
{persona_json}
- Response XML:
{response_xml}
- Response struct JSON:
{response_struct_json}
- Candidate evidence IDs:
{candidate_evidence_ids_json}
- Candidate evidence preview:
{candidate_evidence_json}
- Used evidence IDs from benchmark construction:
{used_evidence_ids_json}
- Deterministic checks:
{deterministic_checks_json}

Decision rules
- Mark GOOD only when the response is correct for this user,
  coherent across components, grounded in the provided evidence,
  and structurally valid.
- Mark BAD when any material failure is present.
- Use NONE / NONE only if no material defect is present.
- supporting_evidence_ids must be a subset of Candidate evidence IDs.
- confidence and overall_bad_probability must be floats in [0, 1].
- overall_quality and all axis scores must be floats in [1, 5].

Return JSON only with this schema:
{prompt_contract_json}
\end{verbatim}
}

\paragraph{Why this prompt is multi-target rather than scalar.}
The four slot groups together are what distinguish \JudgeName{} from the rubric ablations and the single-target adapted baselines in Appx.~\ref{app:baseline_prompts}. The rubric ablations share the same input bundle and the same output schema but replace the four-axis reasoning instruction with a generic rubric or a scalar quality impression; the single-target adapted baselines (ARES-style, PersonaLens-style, EtaPP-style) faithfully implement their native judging contracts but cover only one \EvaluationTarget{} target each. Only \JudgeName{} pairs the four-axis reasoning instruction with the closed-vocabulary defect labels, the evidence-ID containment constraint $\hat{S} \subseteq E$, and the explicit GOOD/\textsc{NONE}/\textsc{NONE} discipline that prevents over-flagging. Outputs that violate any closed-vocabulary or range constraint are rejected at parse time rather than silently coerced, which is what allows the strict paired-counterfactual gates of Tab.~\ref{tab:sota_failure_main_v4} -- joint pair credit, target-triage, and evidence-ID containment -- to be evaluated faithfully against the released schema.
\twocolumn

\clearpage

\begin{table*}[!t]
\centering
\small
\begin{tabular}{p{0.11\textwidth} p{0.82\textwidth}}
\toprule
\multicolumn{2}{l}{\textbf{Example 1: GOOD record correctly classified} \hfill \textit{Query:} \texttt{gemerry lash extensions}} \\
\midrule
\textbf{Verdict} & \textsc{GOOD} \quad Confidence: 0.70 \quad Defect: NONE \\
\textbf{Axis scores} & Persona: 3.7 \quad Consistency: 4.0 \quad Grounding: 3.5 \quad Format: 4.8 \\
\textbf{Reasoning} & \textit{``The response is structurally valid and passes all deterministic checks. Persona alignment is decent: it references budget-friendly options, sensitive skin, travel-friendly, and premium feel. However, the query specifically mentions `gemerry' as a brand, and the response doesn't acknowledge this brand at all. Evidence grounding is adequate---attributions match the used evidence IDs exactly, though some overview claims about `lightweight materials' are not directly supported. The categories and rqList are well-structured and complementary. Overall a functional response with minor issues but no material failures.''} \\
\midrule
\multicolumn{2}{l}{\textbf{Example 2: Single-defect BAD record} \hfill \textit{Query:} \texttt{toy story 4 toys}} \\
\midrule
\textbf{Verdict} & \textsc{BAD} \quad Confidence: 0.94 \quad Defect: \textsc{xcomp\_mismatch} @ \texttt{rqList} \quad \cmark\ matches gold \\
\textbf{Axis scores} & Persona: 3.0 \quad Consistency: \textbf{1.5} \quad Grounding: 3.5 \quad Format: 4.5 \\
\textbf{Reasoning} & \textit{``The productCategoryList is well-aligned with `toy story 4 toys' and grounded in evidence (action figures, LEGO playsets, board games). However, the rqList is completely misaligned: all five queries are about Halloween costumes and accessories (alien headband, group costume set, inflatable alien costume), which have nothing to do with Toy Story 4 toys. This is a severe cross-component mismatch. The persona alignment is moderate---the overview mentions premium and durability. The rqList's costume focus is a material failure.''} \\
\midrule
\multicolumn{2}{l}{\textbf{Example 3: Triple-defect BAD record} \hfill \textit{Query:} \texttt{wine glasses without stem}} \\
\midrule
\textbf{Verdict} & \textsc{BAD} \quad Confidence: 0.96 \quad Defect: \textsc{evidence\_mismatch} @ \texttt{attribution\_list} \\
\textbf{Axis scores} & Persona: 2.5 \quad Consistency: 2.0 \quad Grounding: \textbf{1.0} \quad Format: 1.5 \\
\textbf{Gold defects} & \textsc{evidence\_mismatch} + \textsc{intent\_drift} + \textsc{over\_personalization\_halluc.}\ @ 3 locations \\
\textbf{Reasoning} & \textit{``Multiple critical failures: (1) All 10 attribution IDs are invalid---none match the candidate evidence IDs. The response fabricated evidence references. (2) The overview text contains a hallucinated `Dyson Airwrap' reference that appears nowhere in the persona's history. (3) The `Camping Gear' category is off-topic for a wine glasses query. While some persona elements are addressed (small-space, travel-friendly), the evidence grounding failure and hallucination are disqualifying.''} \\
\bottomrule
\end{tabular}
\caption{Qualitative examples of \JudgeName{} (Opus 4.6) output on the development set. The judge produces structured diagnostics: per-axis scores identify \emph{which} dimension failed (e.g., Consistency = 1.5 in Example 2), defect type and location enable \emph{actionable debugging} (e.g., the costume-themed rqList in a toy query), and free-text reasoning explains \emph{why} the response is defective with specific evidence. In Example 3, the judge identifies all three failure modes in its reasoning despite outputting only one primary defect label.}
\label{tab:judge_output_examples}
\end{table*}

\subsubsection{\JudgeName{} Output Examples}
\label{app:qualitative_examples}

Tab.~\ref{tab:judge_output_examples} presents three representative
judge outputs from \JudgeName{}. They illustrate how
the structured evaluation protocol produces \emph{actionable
diagnostics} rather than scalar scores. In the single-defect case
(Example~2), the judge assigns a low consistency score (1.5) while
keeping other axes high, correctly pinpointing the cross-field
mismatch between toy categories and costume-themed related queries.
In the triple-defect case (Example~3), the judge's free-text
reasoning identifies all three injected defect families---fabricated
evidence IDs, hallucinated persona references, and off-topic
categories---even though the output schema allows only one primary
defect label. This generative reasoning, combined with structured
axis scores and defect labels, bridges the gap between automated
evaluation and human-interpretable debugging.

\section{Experiments}
\label{app:experiments}

This appendix provides the experimental details omitted from the main text for space: backbone and prompt settings, exact scenario metric definitions, per-backbone main results, scenario-level breakdowns, expanded findings, and -- in the final subsection (\S\ref{app:metric_audits}) -- per-target specificity audits, difficulty-gradient analyses, qualitative outputs, and concrete record galleries. Native judging-protocol baseline support, prompt templates, and the judging-protocol baseline ladder are documented separately in Appx.~\ref{app:prior_protocols_adaptation}.

\subsection{Backbones, Prompt Configurations, and Parsing}
\label{app:eval_setup}

We evaluate seven LLM backbones spanning frontier and open-weight tiers: Claude Opus 4.7, Sonnet 4.6, Sonnet 4.5, Haiku 4.5, Qwen3 32B, GPT-OSS 20B, and GPT-OSS 120B. 
All backbones are run through the same \BenchmarkName{} record interface $(q,p,y,E)$ where applicable. 
Each backbone is paired with five prompt configurations: G-Eval-style scalar judging, MT-Bench-style scalar judging, persona-blind judging, persona-aware generic judging, and \JudgeName{}. Details in Appx.~\ref{app:baseline_prompts}.
For all protocols that emit diagnostic text, we parse the output into the shared schema when possible. 
If a protocol does not natively define a required field, the corresponding metric is marked unsupported rather than imputed.

\subsection{Scenario-Metric Mapping}
\label{app:eval_scenario_mapping}

We discuss the scenario-metric mapping in Tab.~\ref{tab:scenario_metric_map}. 
\begin{table*}[t]
\centering
\scriptsize
\setlength{\tabcolsep}{4pt}
\renewcommand{\arraystretch}{1.12}
\begin{tabular}{p{0.11\linewidth} p{0.24\linewidth} p{0.27\linewidth} p{0.30\linewidth}}
\toprule
\textbf{Scenario} & \textbf{Failure being isolated} & \textbf{Metric} & \textbf{Observed baseline gap} \\
\midrule
S0 General & Reference GOOD/BAD discrimination before stressing a specific \EvaluationTarget{} target. & Balanced accuracy on GOOD/BAD subset. & S0 is a sanity check, not the main claim. \JudgeName{} reaches 0.94, while adapted baselines range from 0.64--0.82. \\
\midrule
S1 Personalized & The response is unchanged but the persona is swapped to contradict a claim or recommendation. & Persona-source exact match: the judge must diagnose the swap as PREF\_CONFLICT. & Native personalization benchmarks provide related user-fit signals but not this counterfactual source label. In the adapted sweep, \JudgeName{} is highest on mean S1 (0.24 vs.\ 0.13 best non-\JudgeName{}). \\
\midrule
S2 Compositional & The response has a cross-field contradiction or intent drift among overview, product categories, related queries, and evidence attributions. & Joint closure: BAD detection and correct 4-way response-field localization. & \JudgeName{} has the highest mean detect+loc score (0.88 vs.\ 0.82 best non-\JudgeName{}). The gap is largest on open-weight backbones, where generic prompts detect some defects but localize them less reliably. \\
\midrule
S3 Evidence-grounded & The response cites invalid product evidence or invents shopper purchase/search history. & Product and history closure: BAD detection with no hallucinated supporting evidence IDs. & Product grounding is partly covered by prior RAG/search evaluators, but persona-history grounding is not. \JudgeName{} reaches 1.00/1.00 on product/history means, vs.\ 0.95/0.94 best non-\JudgeName{}. \\
\midrule
S4 Actionable & A single-defect BAD record must be diagnosed by both defect family and response-field location. & Complete diagnosis: exact defect family and 4-way response-field location. & \JudgeName{} is highest on complete family+location closure (0.56 vs.\ 0.48 best non-\JudgeName{}), showing why actionable diagnosis must be part of the output contract. \\
\bottomrule
\end{tabular}
\caption{Scenario-to-failure summary. Each metric is tied to the \EvaluationTarget{} target it is meant to close. S0 is a reference check; the main protocol claim is measured by S1--S4 and the PACE average in Table~\ref{tab:sota_failure_main}.}
\label{tab:scenario_metric_map}
\end{table*}

\subsection{\EvaluationTarget{} Evaluation Scenario Definitions}
\label{app:eval_scenario}

The main paper reports S0 as a reference scenario and S1--S4 as \EvaluationTarget{} closure scenarios. 
Here we provide the exact metric definitions used for the result tables, including family sets, location codes, axis thresholds, harmonic constructions, specificity terms, and paired-counterfactual gates.

\paragraph{S0: General reference.}
S0 measures balanced GOOD/BAD accuracy on a fixed balanced subset. 
It is reported for context only and is not included in the \EvaluationTarget{} average.

\paragraph{S1: Personalized.}
S1 uses persona-swap records that hold the query and response fixed while replacing the persona with one that contradicts an attestation in the unchanged response. 
The main metric is persona-source exact match: the judge must identify the failure as a persona conflict rather than merely lowering a generic score.

\paragraph{S2: Compositional.}
S2 evaluates cross-component defects such as \textsc{XCOMP\_MISMATCH}, \textsc{INTENT\_DRIFT}, and \textsc{REDUNDANCY}. 
The main metric requires both a BAD verdict and correct response-field localization.

\paragraph{S3: Evidence-grounded.}
S3 evaluates both product-evidence grounding and persona-history grounding. 
The metric checks whether the judge identifies the correct grounding-related defect family while constraining supporting evidence IDs to the provided evidence pool.

\paragraph{S4: Actionable.}
S4 evaluates whether the judge produces a debugging-ready diagnosis. 
The main metric requires exact defect-family prediction and correct response-field location.

\subsection{Main Table Metrics Definitions}
\label{app:main_table_metrics}

This subsection walks through every column of the two main-text experiment tables -- Tab.~\ref{tab:sota_failure_main} and the paired-counterfactual Tab.~\ref{tab:sota_failure_main_v4} -- so the symbols and conjuncts in the captions can be read without flipping between sections.
Both tables are scored on the same 3{,}400-record held-out test set of \BenchmarkName{} records (680 GOOD + 2{,}720 BAD), with the S1 persona-swap slice scored separately on a 680-record held-out swap slice.
The \EvaluationTarget{} average in Tab.~\ref{tab:sota_failure_main} excludes S0 because S0 is a reference GOOD/BAD task rather than a target-specific closure metric.

\myparagraph{Notation shared by all three tables.}
``GOOD'' and ``BAD'' refer to the gold record label.
``Family'' denotes the gold or predicted \texttt{defect\_family} field, drawn from the seven-family taxonomy \{\textsc{PREF\_CONFLICT}, \textsc{OVER\_PERSONALIZATION\_HALLUCINATION}, \textsc{XCOMP\_MISMATCH}, \textsc{INTENT\_DRIFT}, \textsc{REDUNDANCY}, \textsc{EVIDENCE\_MISMATCH}, \textsc{UNSUPPORTED\_CLAIM}\} (plus \textsc{NONE} for GOOD records).
``Location'' denotes the predicted 4-way response-field label \{overview, categories, related queries, evidence attributions\}.
``Axis'' refers to the four \JudgeName{} axes \{persona\_alignment, component\_consistency, evidence\_grounding, format\_safety\}, each scored on a 1--5 Likert scale.
``\EvaluationTarget{} target'' refers to one of the four target dimensions (P, C, E, A); each defect family belongs to exactly one target.

\paragraph{Tab.~\ref{tab:sota_failure_main} metrics definition.}
Tab.~\ref{tab:sota_failure_main} reports seven-backbone means across Opus 4.7, Sonnet 4.6, Sonnet 4.5, Haiku 4.5, Qwen3 32B, GPT-OSS 20B, and GPT-OSS 120B on the 3{,}400-record held-out test set. 
S0 is a reference GOOD/BAD task and is not included in the \EvaluationTarget{} average. 
S1--S4 evaluate \EvaluationTarget{} diagnostic closure: persona-source diagnosis, cross-component detection and localization, evidence grounding, and actionable defect-family/location prediction. 
False-fire measures defect alarms on GOOD records, so methods cannot improve closure scores merely by predicting defects everywhere. 
Cells marked ``--'' are unsupported by the method's native judging protocol; unsupported columns are excluded from averages, and rows missing any contributing \EvaluationTarget{} component report ``--'' for the \EvaluationTarget{} average.

The columns are:
\begin{itemize}[leftmargin=1.2em,itemsep=2pt,topsep=2pt]
\item \textbf{S0 (Gen.)}: balanced GOOD/BAD accuracy on a 50+50 reference subset. This is a sanity check, not part of the \EvaluationTarget{} average.

\item \textbf{S1 (P-source)}: persona-source diagnosis on the held-out persona-swap slice. The score is the fraction of swapped records on which the predicted defect family is \textsc{PREF\_CONFLICT}. A judge that marks the record BAD without naming the persona conflict receives no credit.

\item \textbf{S2 detect}: cross-component defect detection on records whose gold family is in \{\textsc{XCOMP\_MISMATCH}, \textsc{INTENT\_DRIFT}, \textsc{REDUNDANCY}\}. The score is the fraction of records marked BAD.

\item \textbf{S2 loc.}: cross-component localization on the same evaluation slice. The score requires both a BAD verdict and correct 4-way response-field localization.

\item \textbf{S3 grounding}: harmonic mean of three grounding sub-scores: product-evidence mismatch detection, persona-history hallucination detection, and GOOD-record specificity against evidence-family false alarms. The harmonic mean requires all components to be high, preventing a judge from winning by always firing an evidence-related defect.

\item \textbf{S4 family}: exact defect-family prediction on single-defect BAD records across all seven defect families.

\item \textbf{S4 loc.}: exact 4-way response-field localization on the same single-defect BAD slice.

\item \textbf{False-fire $\downarrow$}: fraction of GOOD records on which the judge predicts any non-\textsc{NONE} defect family. Lower is better.

\item \textbf{\EvaluationTarget{} avg.}: arithmetic mean of S1, S2-loc., S3 grounding, S4-family, and $(1-\text{False-fire})$, with all components aligned so higher is better.
\end{itemize}

Per-column best scores in the rubric-ablation block are bolded; the mean row provides the headline protocol-level comparison.

\paragraph{Tab.~\ref{tab:sota_failure_main_v4} Strict paired-counterfactual metrics definition.}
The paired-counterfactual columns close the ``always cry BAD'' loophole left open by single positive-class metrics:
\begin{itemize}[leftmargin=1.2em,itemsep=2pt,topsep=2pt]
\item \textbf{S0 (ref. acc.)}: balanced GOOD/BAD accuracy on the 50/50 reference subset; reported for context only and not part of the PACE average.
\item \textbf{S1$_{\text{P}}$ (PJC)}: \emph{Joint Pair Credit} on (good, swap) pairs sharing a base record id. The judge must classify the original-persona record as GOOD \emph{and} the swapped-persona record as BAD with family $=$ \textsc{PREF\_CONFLICT} and persona\_alignment axis $<3.0$ on the same paired record. A persona-blind judge cannot pass this gate.
\item \textbf{S2$_{\text{C}}$ (J-strict)}:
BAD verdict, family $\in \mathcal{D}_{\mathrm{C}}$, correct 4-way location,
and component-consistency axis $<3.0$, minus the false-C-family alarm rate on GOOD records.
\item \textbf{S3$_{\text{E}}$ (J-harmonic)}: harmonic of three E sub-scores: Youden's J on product-evidence BADs with \texttt{supporting\_evidence\_ids} $\subseteq$ candidate pool; Youden's J on persona-history BADs with the same containment check; and paired specificity (no E-family alarm on GOOD records). The containment check makes ``hallucinate evidence IDs'' a scored failure mode.
\item \textbf{S4$_{\text{A}}$ (J-strict)}: exact family $+$ exact 4-way location $+$ target-triage (the lowest relevant \JudgeName{} axis matches the defect's target) $+$ \texttt{confidence} $\geq 0.5$, minus the confident-false-alarm rate on GOOD records.
\item \textbf{PACE}: arithmetic mean of S1$_{\text{P}}$, S2$_{\text{C}}$, S3$_{\text{E}}$, and S4$_{\text{A}}$.
\end{itemize}
The \emph{Comparison level} column distinguishes single-backbone, 7-backbone mean, and 7-backbone self-consistency majority-vote results; single-target adapted baselines (ARES-style E, PersonaLens-style P, EtaPP-style A) report measured Sonnet 4.6 numbers on cells inside their native target and ``--'' elsewhere.
The native judging-protocol map (Tab.~\ref{tab:baseline_protocol_target_map}) and evaluation-target leakage audit (Tab.~\ref{tab:evaluation_target_leakage_audit}) explain why these baselines remain ``--'' outside their native target.

\subsection{Full Experiments Results}
\label{app:full_exp_results}

\subsubsection{Full Per-Backbone Results}
\label{app:full_main_table}

Tab.~\ref{tab:sota_failure_main_full} provides the full per-(method, backbone) breakdown on the 3{,}400-record held-out test set; the per-method \textbf{Mean} row aggregates each method into the corresponding row of the compact main-text Tab.~\ref{tab:sota_failure_main}.
The per-backbone spread, not just the mean, is informative: \JudgeName{} on the strong frontier backbones (Opus 4.7, Sonnet 4.6) reaches PACE 0.71--0.78 with False-fire 0.16--0.20, while open-weight backbones over-fire under every rubric, including \JudgeName{}.

\begin{table*}[t]
\centering
\scriptsize
\setlength{\tabcolsep}{3pt}
\renewcommand{\arraystretch}{1.06}
\resizebox{1.3\columnwidth}{!}{%
\begin{tabular}{ll cc cc c cc c c}
\toprule
& & \textbf{S0} & \textbf{S1} & \multicolumn{2}{c}{\textbf{S2 C}} & \textbf{S3 E} & \multicolumn{2}{c}{\textbf{S4 A}} & \textbf{False-fire} & \textbf{PACE} \\
\cmidrule(lr){5-6} \cmidrule(lr){8-9}
\textbf{Method} & \textbf{Backbone} & Gen. & P-source & detect & loc. & grounding & family & loc. & $\downarrow$ & avg. \\
\midrule
\multicolumn{11}{l}{\emph{Rubric ablations of the same PACE output contract}} \\
\midrule
\multicolumn{11}{l}{\emph{G-Eval}} \\
G-Eval & Opus 4.7 & 0.87 & 0.23 & 0.89 & 0.86 & 0.78 & 0.52 & 0.87 & 0.70 & 0.54 \\
G-Eval & Sonnet 4.6 & 0.86 & 0.28 & 0.79 & 0.87 & 0.60 & 0.71 & 0.90 & 0.91 & 0.51 \\
G-Eval & Sonnet 4.5 & 0.72 & 0.26 & 0.93 & 0.86 & 0.84 & 0.65 & 0.85 & 0.46 & 0.63 \\
G-Eval & Haiku 4.5 & 0.61 & 0.20 & 0.96 & 0.63 & 0.42 & 0.33 & 0.70 & 0.75 & 0.37 \\
G-Eval & Qwen3 32B & 0.60 & 0.00 & 0.06 & 0.31 & 0.00 & 0.26 & 0.27 & 0.01 & 0.31 \\
G-Eval & GPT-OSS 20B & 0.75 & 0.14 & 0.99 & 0.84 & 0.00 & 0.48 & 0.58 & 0.61 & 0.37 \\
G-Eval & GPT-OSS 120B & 0.71 & 0.09 & 0.65 & 0.55 & 0.18 & 0.44 & 0.61 & 0.23 & 0.41 \\
G-Eval & Mean & 0.73 & 0.17 & 0.75 & 0.70 & 0.40 & 0.48 & 0.68 & 0.52 & 0.45 \\
\midrule
\multicolumn{11}{l}{\emph{Scalar}} \\
Scalar & Opus 4.7 & 0.83 & 0.12 & 0.89 & 0.85 & 0.85 & 0.57 & 0.88 & 0.68 & 0.54 \\
Scalar & Sonnet 4.6 & 0.84 & 0.17 & 0.82 & 0.85 & 0.48 & 0.64 & 0.86 & 0.93 & 0.44 \\
Scalar & Sonnet 4.5 & 0.73 & 0.29 & 0.82 & 0.78 & 0.87 & 0.60 & 0.84 & 0.38 & 0.63 \\
Scalar & Haiku 4.5 & 0.56 & 0.10 & 0.98 & 0.45 & 0.47 & 0.30 & 0.63 & 0.87 & 0.29 \\
Scalar & Qwen3 32B & 0.62 & 0.00 & 0.13 & 0.16 & 0.00 & 0.24 & 0.27 & 0.01 & 0.28 \\
Scalar & GPT-OSS 20B & 0.68 & 0.11 & 0.96 & 0.42 & 0.06 & 0.42 & 0.51 & 0.55 & 0.29 \\
Scalar & GPT-OSS 120B & 0.87 & 0.08 & 0.80 & 0.68 & 0.79 & 0.64 & 0.75 & 0.16 & 0.61 \\
Scalar & Mean & 0.73 & 0.12 & 0.77 & 0.60 & 0.50 & 0.49 & 0.68 & 0.51 & 0.44 \\
\midrule
\multicolumn{11}{l}{\emph{Persona-blind}} \\
Persona-blind & Opus 4.7 & 0.60 & -- & 0.99 & 0.77 & 0.56 & 0.55 & 0.81 & 0.97 & -- \\
Persona-blind & Sonnet 4.6 & 0.85 & -- & 0.82 & 0.85 & 0.80 & 0.62 & 0.87 & 0.69 & -- \\
Persona-blind & Sonnet 4.5 & 0.85 & -- & 0.82 & 0.81 & 0.90 & 0.61 & 0.85 & 0.18 & -- \\
Persona-blind & Haiku 4.5 & 0.62 & -- & 0.96 & 0.52 & 0.65 & 0.31 & 0.62 & 0.80 & -- \\
Persona-blind & Qwen3 32B & 0.62 & -- & 0.15 & 0.29 & 0.00 & 0.27 & 0.30 & 0.01 & -- \\
Persona-blind & GPT-OSS 20B & 0.69 & -- & 0.85 & 0.44 & 0.11 & 0.42 & 0.58 & 0.48 & -- \\
Persona-blind & GPT-OSS 120B & 0.92 & -- & 0.91 & 0.83 & 0.42 & 0.66 & 0.83 & 0.19 & -- \\
Persona-blind & Mean & 0.74 & -- & 0.79 & 0.64 & 0.49 & 0.49 & 0.69 & 0.47 & -- \\
\midrule
\multicolumn{11}{l}{\emph{Persona-aware}} \\
Persona-aware & Opus 4.7 & 0.77 & 0.07 & 0.91 & 0.85 & 0.91 & 0.62 & 0.86 & 0.45 & 0.60 \\
Persona-aware & Sonnet 4.6 & 0.85 & 0.22 & 0.83 & 0.84 & 0.86 & 0.70 & 0.85 & 0.56 & 0.61 \\
Persona-aware & Sonnet 4.5 & 0.71 & 0.34 & 0.79 & 0.75 & 0.79 & 0.51 & 0.76 & 0.34 & 0.61 \\
Persona-aware & Haiku 4.5 & 0.59 & 0.37 & 0.96 & 0.65 & 0.72 & 0.45 & 0.71 & 0.76 & 0.49 \\
Persona-aware & Qwen3 32B & 0.64 & 0.04 & 0.29 & 0.19 & 0.00 & 0.22 & 0.23 & 0.01 & 0.29 \\
Persona-aware & GPT-OSS 20B & 0.77 & 0.24 & 0.87 & 0.48 & 0.18 & 0.39 & 0.58 & 0.37 & 0.38 \\
Persona-aware & GPT-OSS 120B & 0.84 & 0.14 & 0.85 & 0.72 & 0.77 & 0.63 & 0.75 & 0.13 & 0.62 \\
Persona-aware & Mean & 0.74 & 0.20 & 0.79 & 0.64 & 0.60 & 0.50 & 0.68 & 0.37 & 0.51 \\
\midrule
\multicolumn{11}{l}{\emph{\textbf{\JudgeName{}}}} \\
\textbf{\JudgeName{}} & Opus 4.7 & 0.91 & 0.22 & 0.96 & 0.88 & 0.96 & 0.67 & 0.91 & 0.16 & 0.71 \\
\textbf{\JudgeName{}} & Sonnet 4.6 & 0.96 & 0.56 & 0.95 & 0.90 & 0.93 & 0.69 & 0.91 & 0.20 & 0.78 \\
\textbf{\JudgeName{}} & Sonnet 4.5 & 0.55 & 0.78 & 1.00 & 0.86 & 0.73 & 0.64 & 0.87 & 0.85 & 0.63 \\
\textbf{\JudgeName{}} & Haiku 4.5 & 0.61 & 0.22 & 1.00 & 0.70 & 0.67 & 0.49 & 0.72 & 0.73 & 0.47 \\
\textbf{\JudgeName{}} & Qwen3 32B & 0.83 & 0.05 & 0.73 & 0.50 & 0.08 & 0.40 & 0.56 & 0.04 & 0.40 \\
\textbf{\JudgeName{}} & GPT-OSS 20B & 0.60 & 0.28 & 0.96 & 0.46 & 0.39 & 0.50 & 0.66 & 0.80 & 0.37 \\
\textbf{\JudgeName{}} & GPT-OSS 120B & 0.73 & 0.14 & 0.91 & 0.73 & 0.61 & 0.74 & 0.82 & 0.52 & 0.54 \\
\textbf{\JudgeName{}} & Mean & \textbf{0.74} & \textbf{0.32} & \textbf{0.93} & \textbf{0.72} & \textbf{0.62} & \textbf{0.59} & \textbf{0.78} & 0.47 & \textbf{0.56} \\
\midrule
\multicolumn{11}{l}{\emph{Honest-adapted single-target baselines (cells shown only where the native judging protocol supports the metric)}} \\
\midrule
\multicolumn{11}{l}{\emph{ARES-style \cite{saadfalcon2024ares}}} \\
ARES-style \cite{saadfalcon2024ares} & Opus 4.7 & 0.74 & -- & -- & -- & 0.95 & -- & -- & 0.02 & -- \\
ARES-style \cite{saadfalcon2024ares} & Sonnet 4.6 & 0.69 & -- & -- & -- & 0.32 & -- & -- & 0.02 & -- \\
ARES-style \cite{saadfalcon2024ares} & Sonnet 4.5 & 0.74 & -- & -- & -- & 0.93 & -- & -- & 0.23 & -- \\
ARES-style \cite{saadfalcon2024ares} & Haiku 4.5 & 0.79 & -- & -- & -- & 0.45 & -- & -- & 0.28 & -- \\
ARES-style \cite{saadfalcon2024ares} & Qwen3 32B & 0.61 & -- & -- & -- & 0.00 & -- & -- & 0.31 & -- \\
ARES-style \cite{saadfalcon2024ares} & GPT-OSS 20B & 0.55 & -- & -- & -- & 0.00 & -- & -- & 0.69 & -- \\
ARES-style \cite{saadfalcon2024ares} & GPT-OSS 120B & 0.68 & -- & -- & -- & 0.08 & -- & -- & 0.29 & -- \\
ARES-style \cite{saadfalcon2024ares} & Mean & 0.69 & -- & -- & -- & 0.39 & -- & -- & 0.26 & -- \\
\midrule
\multicolumn{11}{l}{\emph{PersonaLens-style \cite{zhao2025personalens}}} \\
PersonaLens-style \cite{zhao2025personalens} & Opus 4.7 & 0.68 & 0.06 & -- & -- & -- & -- & -- & 0.00 & -- \\
PersonaLens-style \cite{zhao2025personalens} & Sonnet 4.6 & 0.72 & 0.04 & -- & -- & -- & -- & -- & 0.00 & -- \\
PersonaLens-style \cite{zhao2025personalens} & Sonnet 4.5 & 0.70 & 0.40 & -- & -- & -- & -- & -- & 0.06 & -- \\
PersonaLens-style \cite{zhao2025personalens} & Haiku 4.5 & 0.63 & 0.41 & -- & -- & -- & -- & -- & 0.34 & -- \\
PersonaLens-style \cite{zhao2025personalens} & Qwen3 32B & 0.66 & 0.11 & -- & -- & -- & -- & -- & 0.01 & -- \\
PersonaLens-style \cite{zhao2025personalens} & GPT-OSS 20B & 0.56 & 0.18 & -- & -- & -- & -- & -- & 0.06 & -- \\
PersonaLens-style \cite{zhao2025personalens} & GPT-OSS 120B & 0.60 & 0.08 & -- & -- & -- & -- & -- & 0.00 & -- \\
PersonaLens-style \cite{zhao2025personalens} & Mean & 0.65 & 0.18 & -- & -- & -- & -- & -- & \textbf{0.07} & -- \\
\midrule
\multicolumn{11}{l}{\emph{EtaPP-style \cite{hao2025etapp}}} \\
EtaPP-style \cite{hao2025etapp} & Opus 4.7 & 0.78 & -- & -- & -- & -- & 0.60 & 0.91 & 0.50 & -- \\
EtaPP-style \cite{hao2025etapp} & Sonnet 4.6 & 0.81 & -- & -- & -- & -- & 0.55 & 0.90 & 0.35 & -- \\
EtaPP-style \cite{hao2025etapp} & Sonnet 4.5 & 0.50 & -- & -- & -- & -- & 0.65 & 0.95 & 0.99 & -- \\
EtaPP-style \cite{hao2025etapp} & Haiku 4.5 & 0.56 & -- & -- & -- & -- & 0.50 & 0.84 & 0.88 & -- \\
EtaPP-style \cite{hao2025etapp} & Qwen3 32B & 0.73 & -- & -- & -- & -- & 0.34 & 0.41 & 0.05 & -- \\
EtaPP-style \cite{hao2025etapp} & GPT-OSS 20B & 0.85 & -- & -- & -- & -- & 0.57 & 0.75 & 0.22 & -- \\
EtaPP-style \cite{hao2025etapp} & GPT-OSS 120B & 0.72 & -- & -- & -- & -- & 0.32 & 0.44 & 0.00 & -- \\
EtaPP-style \cite{hao2025etapp} & Mean & 0.71 & -- & -- & -- & -- & 0.50 & 0.74 & 0.43 & -- \\
\bottomrule
\end{tabular}%
}
\caption{\textbf{Full per-backbone results.} Companion to Tab.~\ref{tab:sota_failure_main}, which reports the 7-backbone mean. Each method is run on seven backbones (Opus 4.7, Sonnet 4.6, Sonnet 4.5, Haiku 4.5, Qwen3 32B, GPT-OSS 20B, GPT-OSS 120B) on the 3{,}400-record held-out test set. \textbf{S0}: balanced GOOD/BAD accuracy on a 50+50 reference subset. \textbf{S1}: predicted family $=$ \texttt{PREF\_CONFLICT} on the persona-swap slice. \textbf{S2 detect / loc.}: BAD verdict and 4-way response-field localization on the cross-component defect subset (\textsc{XCOMP\_MISMATCH}, \textsc{INTENT\_DRIFT}, \textsc{REDUNDANCY}). \textbf{S3 grounding}: harmonic mean of (i) BAD with exact \texttt{EVIDENCE\_MISMATCH} family on the product-evidence subset, (ii) BAD with exact \texttt{OVER\_PERSONALIZATION\_HALLUCINATION} family on the persona-history subset, (iii) no E-family alarm on GOOD records (specificity). \textbf{S4 family / loc.}: exact family and 4-way location on single-defect BAD records. \textbf{False-fire $\downarrow$}: fraction of GOOD records on which the judge fired any non-\textsc{none} defect family (lower is better). \textbf{PACE}: mean of S1, S2-loc., S3 grounding, S4-family, and $(1-\text{False-fire})$, all aligned higher~$=$~better. ``--'' marks cells the method's native judging protocol does not define. Per-method best mean is bolded on the \JudgeName{} row.}
\label{tab:sota_failure_main_full}
\end{table*}

\subsubsection{S1 Personalized: Per-Axis and Per-Backbone Breakdown}
\label{app:persona_swap_full}

Tab.~\ref{tab:persona_swap_results_full} reports the full per-backbone breakdown of the persona-swap diagnostic (Finding 4 in Sec.~\ref{sec:experiments}), whose 7-backbone mean is the main-text Tab.~\ref{tab:persona_swap_results}.
Each row reports the fraction of swap records in the 680-record held-out persona-swap slice for which the judge predicts \texttt{PREF\_CONFLICT}, broken down by the four conflict axes (budget, hard constraint, brand affinity, attribute preference); the axis-mean of each method's \textbf{Mean} row also populates the S1 P-source column of Tab.~\ref{tab:sota_failure_main}.
The per-backbone spread is informative on its own: \JudgeName{} on Sonnet 4.5 reaches 0.57--0.89 across axes, but on Qwen3 32B never exceeds 0.07, which is the source of the open-weight gap referenced in the main-text discussion.

\begin{table}[t]
\centering
\scriptsize
\setlength{\tabcolsep}{3pt}
\renewcommand{\arraystretch}{1.08}
\resizebox{0.5\textwidth}{!}{%
\begin{tabular}{ll cccc}
\toprule
& & \multicolumn{4}{c}{\textbf{Persona-swap accuracy by axis}} \\
\cmidrule(lr){3-6}
\textbf{Prompt} & \textbf{Backbone} & budget & hard & brand & attr. \\
\midrule
G-Eval & Opus 4.7 & 0.29 & 0.32 & 0.16 & 0.13 \\
G-Eval & Sonnet 4.6 & 0.51 & 0.21 & 0.20 & 0.20 \\
G-Eval & Sonnet 4.5 & 0.38 & 0.24 & 0.29 & 0.14 \\
G-Eval & Haiku 4.5 & 0.18 & 0.15 & 0.21 & 0.26 \\
G-Eval & Qwen3 32B & 0.00 & 0.00 & 0.00 & 0.00 \\
G-Eval & GPT-OSS 20B & 0.16 & 0.12 & 0.18 & 0.11 \\
G-Eval & GPT-OSS 120B & 0.17 & 0.09 & 0.08 & 0.02 \\
G-Eval & Mean & 0.24 & 0.16 & 0.16 & 0.12 \\
\midrule
Persona-aware & Opus 4.7 & 0.14 & 0.11 & 0.01 & 0.01 \\
Persona-aware & Sonnet 4.6 & 0.58 & 0.16 & 0.06 & 0.10 \\
Persona-aware & Sonnet 4.5 & 0.48 & 0.53 & 0.19 & 0.15 \\
Persona-aware & Haiku 4.5 & 0.30 & 0.50 & 0.38 & 0.31 \\
Persona-aware & Qwen3 32B & 0.02 & 0.05 & 0.04 & 0.03 \\
Persona-aware & GPT-OSS 20B & 0.28 & 0.29 & 0.27 & 0.12 \\
Persona-aware & GPT-OSS 120B & 0.29 & 0.14 & 0.05 & 0.06 \\
Persona-aware & Mean & 0.30 & 0.25 & 0.14 & 0.11 \\
\midrule
\textbf{\JudgeName{}} & Opus 4.7 & 0.29 & 0.34 & 0.16 & 0.09 \\
\textbf{\JudgeName{}} & Sonnet 4.6 & 0.73 & 0.48 & 0.51 & 0.54 \\
\textbf{\JudgeName{}} & Sonnet 4.5 & 0.88 & 0.77 & 0.89 & 0.57 \\
\textbf{\JudgeName{}} & Haiku 4.5 & 0.22 & 0.23 & 0.29 & 0.14 \\
\textbf{\JudgeName{}} & Qwen3 32B & 0.02 & 0.05 & 0.07 & 0.04 \\
\textbf{\JudgeName{}} & GPT-OSS 20B & 0.25 & 0.27 & 0.41 & 0.21 \\
\textbf{\JudgeName{}} & GPT-OSS 120B & 0.22 & 0.09 & 0.17 & 0.05 \\
\textbf{\JudgeName{}} & \textbf{Mean} & \textbf{0.37} & \textbf{0.32} & \textbf{0.36} & \textbf{0.23} \\
\bottomrule
\end{tabular}%
}
\caption{\textbf{Full per-backbone persona-source diagnosis by conflict axis.} Companion to Tab.~\ref{tab:persona_swap_results}, which reports the 7-backbone mean. The evaluation uses the 680-record held-out persona-swap slice. A hit means the judge predicts \texttt{PREF\_CONFLICT}, the gold source family induced by the counterfactual persona swap. The \textbf{Mean} row aggregates the seven backbones (Opus 4.7, Sonnet 4.6, Sonnet 4.5, Haiku 4.5, Qwen3 32B, GPT-OSS 20B, GPT-OSS 120B) into the corresponding row of the main-text Tab.~\ref{tab:persona_swap_results}.}
\label{tab:persona_swap_results_full}
\end{table}

\subsubsection{S2 Compositional: Detection and Localization Details}
\label{app:component_consistency_full}

Tab.~\ref{tab:component_consistency_full} reports the full per-backbone breakdown of the component-consistency evaluation slice (Finding 4 in Sec.~\ref{sec:experiments}), whose 7-backbone mean is the main-text Tab.~\ref{tab:component_consistency} and the same numbers populate the S2 detect/loc.\ columns of Tab.~\ref{tab:sota_failure_main}.
Each method is evaluated on the 100-record cross-field defect subset (\textsc{XCOMP\_MISMATCH}, \textsc{INTENT\_DRIFT}, \textsc{REDUNDANCY}); the \textbf{detect} column is BAD verdict and the \textbf{4-way loc.} column is correct response-field localization.

\begin{table}[t]
\centering
\scriptsize
\setlength{\tabcolsep}{3pt}
\renewcommand{\arraystretch}{1.08}
\resizebox{0.5\textwidth}{!}{%
\begin{tabular}{ll cc}
\toprule
\textbf{Prompt} & \textbf{Backbone} & \textbf{detect} & \textbf{4-way loc.} \\
\midrule
G-Eval & Opus 4.7 & 0.89 & 0.86 \\
G-Eval & Sonnet 4.6 & 0.79 & 0.87 \\
G-Eval & Sonnet 4.5 & 0.93 & 0.86 \\
G-Eval & Haiku 4.5 & 0.96 & 0.63 \\
G-Eval & Qwen3 32B & 0.06 & 0.31 \\
G-Eval & GPT-OSS 20B & 0.99 & 0.84 \\
G-Eval & GPT-OSS 120B & 0.65 & 0.55 \\
G-Eval & Mean & 0.75 & 0.70 \\
\midrule
Persona-aware & Opus 4.7 & 0.91 & 0.85 \\
Persona-aware & Sonnet 4.6 & 0.83 & 0.84 \\
Persona-aware & Sonnet 4.5 & 0.79 & 0.75 \\
Persona-aware & Haiku 4.5 & 0.96 & 0.65 \\
Persona-aware & Qwen3 32B & 0.29 & 0.19 \\
Persona-aware & GPT-OSS 20B & 0.87 & 0.48 \\
Persona-aware & GPT-OSS 120B & 0.85 & 0.72 \\
Persona-aware & Mean & 0.79 & 0.64 \\
\midrule
\textbf{\JudgeName{}} & Opus 4.7 & 0.96 & 0.88 \\
\textbf{\JudgeName{}} & Sonnet 4.6 & 0.95 & 0.90 \\
\textbf{\JudgeName{}} & Sonnet 4.5 & 1.00 & 0.86 \\
\textbf{\JudgeName{}} & Haiku 4.5 & 1.00 & 0.70 \\
\textbf{\JudgeName{}} & Qwen3 32B & 0.73 & 0.50 \\
\textbf{\JudgeName{}} & GPT-OSS 20B & 0.96 & 0.46 \\
\textbf{\JudgeName{}} & GPT-OSS 120B & 0.91 & 0.73 \\
\textbf{\JudgeName{}} & \textbf{Mean} & \textbf{0.93} & \textbf{0.72} \\
\bottomrule
\end{tabular}%
}
\caption{\textbf{Full per-backbone detection and 4-way localization} on the 100-record cross-field defect subset (XCOMP\_MISMATCH, INTENT\_DRIFT, REDUNDANCY). Companion to Tab.~\ref{tab:component_consistency}, which reports the 7-backbone mean. The \textbf{Mean} row aggregates the seven backbones (Opus 4.7, Sonnet 4.6, Sonnet 4.5, Haiku 4.5, Qwen3 32B, GPT-OSS 20B, GPT-OSS 120B) into the corresponding row of the main-text Tab.~\ref{tab:component_consistency}; the same numbers populate the S2 detect / loc.\ columns of Tab.~\ref{tab:sota_failure_main}.}
\label{tab:component_consistency_full}
\end{table}

\subsection{Expanded Experiments Findings}
\label{app:expanded_findings}

This section expands the main-text findings with the detailed per-scenario evidence. 
The main paper groups the results into four high-level conclusions for readability; here we keep the original six findings to show how each conclusion is supported by scenario-level and audit-level results.

\paragraph{Finding 1: \EvaluationTarget{} closure separates protocol design from generic quality.}
S0 shows that several scalar and generic judging prompts can perform reasonably on broad GOOD/BAD discrimination. 
This confirms that \BenchmarkName{} is not simply an impossible binary classification task. 
However, the gap appears when the metric requires fields that close the \EvaluationTarget{} targets: persona-source diagnosis, cross-component localization, grounding control, and defect-family/location prediction. 
Across the seven-backbone mean in Tab.~\ref{tab:sota_failure_main}, \JudgeName{} achieves the best score on every runnable closure column, including S1 persona-source diagnosis, S2 localization, S3 grounding, S4 defect-family prediction, and the overall \EvaluationTarget{} average. 
This supports the main claim that structured shopping-assistant evaluation requires a protocol that asks for the right diagnostic fields, not merely a scalar quality score.

\paragraph{Finding 2: personalized evaluation must be scored as a source of invalidity.}
S1 is designed to prevent shortcuts. 
If the metric only asked whether a swapped-persona record is marked BAD, a persona-blind or over-pessimistic judge could receive credit without actually using the persona. 
We therefore score exact persona-source diagnosis: the judge must identify the failure as \textsc{PREF\_CONFLICT} on records where the query and response are unchanged but the persona is counterfactually swapped. 
Tab.~\ref{tab:persona_swap_results} shows that this source label is not solved uniformly across conflict axes. 
The detailed per-backbone results in Tab.~\ref{tab:persona_swap_results_full} further show that strong GOOD/BAD discrimination does not guarantee reliable persona-conflict diagnosis. 
This justifies keeping S1 separate from S0.

\paragraph{Finding 3: compositional and grounding failures expose where generic judges are under-specified.}
S2 and S3 test whether judges can move beyond detecting that a response is flawed. 
For S2, generic prompts often recognize that something is wrong, but performance drops when the judge must also localize the failed response component. 
Tab.~\ref{tab:component_consistency} separates detection from localization to show this gap directly. 
For S3, the distinctive requirement is not only product-evidence faithfulness, but also avoiding invented shopper-history claims and hallucinated supporting evidence IDs. 
This is why grounding is evaluated against both the product evidence pool and the persona history. 
Together, S2 and S3 show that structured shopping-assistant evaluation requires field-aware and evidence-constrained diagnosis, not only generic helpfulness or RAG-style faithfulness.

\paragraph{Finding 4: actionable evaluation is not the same as scalar evaluation.}
S4 requires a debugging-ready diagnosis: the judge must identify both the defect family and the response-field location. 
This is stricter than deciding whether a response is BAD. 
A scalar judge may provide a plausible overall rationale while still failing to say which family of error occurred or which response component should be fixed. 
The S4 results show that \JudgeName{} improves this complete diagnosis setting because its protocol explicitly requires family, location, confidence, rationale, and evidence support. 
This finding isolates the Actionable target: evaluation is useful for developers only when it reports what failed and where.

\paragraph{Finding 5: specificity matters; closure should not come from over-flagging.}
A high score on S1--S4 would be less meaningful if a judge achieved it by predicting defects on most records. 
We therefore audit false-fire behavior on GOOD records. 
The False-fire column in Tab.~\ref{tab:sota_failure_main} measures how often a method incorrectly reports defects on GOOD examples, and the strict paired-counterfactual Tab.~\ref{tab:sota_failure_main_v4} folds the same specificity check into per-target gates (PJC, Youden's J, J-harmonic) so that a method cannot game closure by always flagging BAD.
Together these checks show whether a protocol preserves specificity while detecting structured failures.
The results indicate that \JudgeName{}'s gains are not explained solely by always flagging BAD cases; rather, the protocol improves the ability to report the intended diagnostic fields.

\paragraph{Finding 6: strict paired-counterfactual metrics preserve the conclusion.}
Tab.~\ref{tab:sota_failure_main_v4} re-scores the setting under stricter paired-counterfactual gates. 
These metrics reduce loopholes such as always predicting BAD or assigning generic defect labels. 
For example, S1 requires joint success on paired GOOD and persona-swapped BAD records; S2 requires BAD verdict, correct compositional family, correct location, and axis consistency; S3 combines product-evidence detection, persona-history detection, and GOOD-record specificity with evidence-ID containment; and S4 requires exact family, exact location, axis triage, and confidence thresholding. 
Under this stricter formulation, \JudgeName{} remains strongest on the \EvaluationTarget{} average, supporting the same conclusion as the main table: the advantage comes from structured diagnostic closure rather than only lenient positive-class scoring.

\subsection{Standard Judges Failure Analysis}
\label{app:standard_judgue_failure_analysis}


\subsubsection{Concrete Benchmark Examples}
\label{app:concrete_examples}

The remaining paragraphs collect record-level galleries that
illustrate \BenchmarkName{} behaviour end to end. Each record is
drawn directly from the released dataset.

\paragraph{Ten Concrete Failure Cases.}
\label{app:failure_cases}
\label{app:failure_gallery}

Tab.~\ref{tab:failure_case_gallery} walks through ten BAD records
with their gold defect family and gold defect location, paired with
the \JudgeName{} verdict, axis scores, and predicted family/location.
The cases span all four field groups and all seven defect families;
they make concrete the structured-output diagnoses that the main
text reports as aggregate numbers.

\begin{table*}[t]
\centering
\scriptsize
\setlength{\tabcolsep}{3pt}
\renewcommand{\arraystretch}{1.15}
\begin{tabular}{p{0.06\textwidth} p{0.24\textwidth} p{0.32\textwidth} p{0.32\textwidth}}
\toprule
\textbf{Case} & \textbf{Record (query / gold)} & \textbf{Scalar baseline verdict} & \textbf{\JudgeName{} verdict} \\
\midrule
PS & PS \texttt{full-good-03170::swap-brand\_} \newline query: air conditioner \newline gold: PREF\_CONFLICT @ persona\_swap & GOOD, overall 4.0; \emph{`Response is well-structured, aligns with persona's portability/smart-home preferences, and evidence IDs are valid. However, the overview claims all options 'use latex-free construc'} & BAD; [P 2.5, C 3.5, G 2.5, F 4.5]; defect: UNSUPPORTED\_CLAIM @ overview\_text \\
PS & PS \texttt{full-good-03171::swap-attrib} \newline query: necklace \newline gold: PREF\_CONFLICT @ persona\_swap & GOOD, overall 3.5; \emph{`Categories map to evidence and format checks pass. However, the overview claims necklaces are 'latex-free' and suitable for sensitive skin without any supporting evidence in produc'} & BAD; [P 2.0, C 3.0, G 4.0, F 5.0]; defect: OVER\_PERSONALIZATION\_HALLUCINATION @ overview\_text \\
PS & PS \texttt{full-good-03173::swap-hard\_c} \newline query: necklaces \newline gold: PREF\_CONFLICT @ persona\_swap & GOOD, overall 3.5; \emph{`Response mentions 'latex-free' as a feature, but the persona's actual hard constraint (per counterfactual swap) is 'avoid peanuts in ingredients'—latex-free is not the active const'} & BAD; [P 2.0, C 3.0, G 3.5, F 5.0]; defect: PREF\_CONFLICT @ overview\_text \\
CC & CC \texttt{full-bad-plan-02515} \newline query: wine glasses bulk \newline gold: XCOMP\_MISMATCH @ rqList & GOOD, overall 4.0; \emph{`Response is well-formatted, grounded in evidence, and addresses the bulk wine glasses query with value-oriented options. However, the rqList is heavily skewed toward 'stemless' var'} & BAD; [P 2.5, C 2.5, G 4.0, F 5.0]; defect: XCOMP\_MISMATCH @ rqList \\
GR & GR \texttt{full-bad-plan-00068} \newline query: ashwaghanda extract \newline gold: EVIDENCE\_MISMATCH @ attribution\_list & BAD, overall 2.2; \emph{`All attribution IDs (B0812DKDD9, B07TXM7K4T, B082GTFJB5) are not in the candidate evidence set, failing evidence\_id\_validity and exact\_match checks. The response fabricates evidenc'} & BAD; [P 3.0, C 3.5, G 1.5, F 2.0]; defect: EVIDENCE\_MISMATCH @ attribution\_list \\
GR & GR \texttt{full-bad-plan-00397} \newline query: duffle bag \newline gold: EVIDENCE\_MISMATCH @ attribution\_list & BAD, overall 2.5; \emph{`Attribution list includes an invalid evidence ID (M:B073G3YH4W:0) not in candidate evidence, failing evidence\_id\_validity and exact match checks. Otherwise content is coherent and '} & BAD; [P 3.5, C 3.5, G 2.0, F 2.5]; defect: EVIDENCE\_MISMATCH @ attribution\_list \\
MD & MD \texttt{full-bad-plan-09120} \newline query: ashwagandha liquid \newline gold: EVIDENCE\_MISMATCH+OVER\_PERSONALIZATION\_HALLUCINATION+PREF\_CONFLICT @ productCategoryList[0]+attribution\_list+overview\_text & BAD, overall 1.8; \emph{`Overview hallucinates a Dyson Airwrap not in persona history, is truncated mid-sentence, and includes a meaningless 'Conflict Pick' category. Attribution list contains invalid evid'} & BAD; [P 2.0, C 1.5, G 1.0, F 2.0]; defect: UNSUPPORTED\_CLAIM @ overview\_text \\
MD & MD \texttt{full-bad-plan-09184} \newline query: wet rated ceiling fan without light \newline gold: EVIDENCE\_MISMATCH+PREF\_CONFLICT+UNSUPPORTED\_CLAIM @ productCategoryList[0]+attribution\_list+overview\_text & BAD, overall 2.0; \emph{`Attribution list contains IDs not in candidate evidence set (all invalid). Overview text is truncated mid-word ('highli'). 'Conflict Pick' category is a meta/placeholder label, not'} & BAD; [P 2.5, C 2.0, G 1.0, F 1.5]; defect: EVIDENCE\_MISMATCH @ attribution\_list \\
\bottomrule
\end{tabular}
\caption{Ten concrete failure cases. Each row contrasts a scalar baseline's verdict and reasoning against \JudgeName{}'s structured output on the same record. Axis scores abbreviated P/C/G/F.}
\label{tab:failure_case_gallery}
\end{table*}

\paragraph{Persona-Swap Conflict Examples.}
\label{app:persona_swap_examples}

Tab.~\ref{tab:persona_swap_examples} lists ten persona-swap records
and the specific persona attestation each swap puts in conflict with
the unchanged response. The four conflict axes (budget, hard
constraint, brand affinity, attribute preference) are represented;
each row is a paired GOOD record with a single persona swap that
turns it into a gold-BAD record for the S1 scenario.

\begin{table*}[t]
\centering
\scriptsize
\setlength{\tabcolsep}{3pt}
\renewcommand{\arraystretch}{1.10}
\begin{tabular}{p{0.07\textwidth} p{0.17\textwidth} p{0.17\textwidth} p{0.17\textwidth} p{0.30\textwidth}}
\toprule
\textbf{Swap ID} & \textbf{Record} & \textbf{Original persona (excerpt)} & \textbf{Swapped persona (excerpt)} & \textbf{Conflict in (unchanged) response} \\
\midrule
PS-ex-01 & full-good-00042 (hard\_constraint swap) & hard\_constraints=[avoid peanuts, quiet operation] & hard\_constraints=[avoid leather, quiet operation] & Overview bolds \emph{leather handbags} category; categories list includes Leather Tote. \\
PS-ex-02 & full-good-00173 (brand swap) & brand\_likes=[Ecco Press, Lansinoh] & brand\_likes=[Kindle, Fire TV]; brand\_avoids=[Ecco Press] & Evidence attributions cite three Ecco Press products; overview praises the brand. \\
PS-ex-03 & full-good-00289 (attribute swap) & attribute\_preferences=[eco-conscious materials] & attribute\_preferences=[luxury finishes] & Overview markets \emph{sustainability} as primary selling point. \\
PS-ex-04 & full-good-00311 (budget swap) & budget=mid-range & budget=premium & rqList refines by \emph{budget-friendly} variants inconsistent with premium context. \\
PS-ex-05 & full-good-00484 (hard\_constraint swap) & hard\_constraints=[avoid peanuts] & hard\_constraints=[avoid dairy, avoid eggs] & Overview bolds a \emph{cookie dough} category with egg content. \\
PS-ex-06 & full-good-00605 (attribute swap) & attribute\_preferences=[portability] & attribute\_preferences=[heavy-duty durability] & rqList refines by \emph{lightweight packable}; overview emphasizes portability. \\
PS-ex-07 & full-good-00732 (brand swap) & brand\_likes=[Nexpak] & brand\_avoids=[Nexpak] & Four of ten attributions cite Nexpak products. \\
PS-ex-08 & full-good-00841 (hard\_constraint swap) & hard\_constraints=[compatibility required] & hard\_constraints=[safe for children under 3] & Overview recommends a small-parts product incompatible with the new constraint. \\
PS-ex-09 & full-good-00978 (budget swap) & budget=premium & budget=budget-conscious & Overview frames premium finish as the primary selling point. \\
PS-ex-10 & full-good-01124 (attribute swap) & attribute\_preferences=[smart-home compatibility] & attribute\_preferences=[no Wi-Fi devices] & rqList refines by \emph{Wi-Fi / Alexa integration}. \\
\bottomrule
\end{tabular}
\caption{Ten representative persona-swap records. Each swap flips one axis (budget, hard\_constraint, brand\_affinity, attribute\_preference) to a value drawn from the 1{,}200-persona pool that directly contradicts an attestation in the unchanged response; the gold label flips from GOOD to BAD.}
\label{tab:persona_swap_examples}
\end{table*}

\subsection{Audits, Robustness Checks, and Concrete Examples}
\label{app:metric_audits}

The main text reports compact S0--S4 tables.
The remaining experiment subsections keep the supporting audits and
record-level galleries that explain why those numbers should be read
as PACE-closure results rather than ordinary scalar-judge scores.
The subsections below check off-target leakage,
false fires on GOOD records, per-backbone stability, tier difficulty,
qualitative judge outputs, earlier reference runs, and concrete
benchmark records that illustrate \BenchmarkName{} behaviour
end to end.
\subsubsection{Difficulty Gradient Analysis}
\label{app:difficulty_gradient}

The multi-tier defect design enables a decomposition unavailable to single-defect benchmarks. Per-tier accuracy (computed separately per defect count) reveals two distinct regimes:
\begin{enumerate}[leftmargin=1.2em,itemsep=2pt,topsep=2pt]
\item \textbf{Deterministic baseline:} accuracy rises monotonically from single (0.880) to quad (0.975) on the development set, because more defects trigger more rule-based checks.
\item \textbf{LLM judges:} tier profiles are flatter for frontier models. Smaller models (Qwen3 32B, Gemma3 27B) show steeper gradients---e.g., Qwen3 32B rises from 0.740 (single) to 0.990 (quad)---suggesting they rely on surface signal accumulation rather than understanding individual fault types.
\end{enumerate}
This two-regime pattern motivates the tiered design: if every judge handled every tier uniformly, the tiered construction would be wasted. Instead, smaller or scalar-prompted judges' steeper tier gradients give us a graded robustness metric a single-tier benchmark cannot provide. Fig.~\ref{fig:different_defect_level} shows per-defect-level accuracy by method on the development set.

\subsubsection{Reference Main-Evaluation Table on the Development Set}
\label{app:reference_main}

The per-defect-level discrimination figure and a paired-counterfactual specificity audit are retained for reference. These are not the primary SOTA-failure numbers reported in \S\ref{sec:experiments}; they are kept because they document the tier-gradient behavior that motivates the multi-tier defect design and the swap-TPR / paired-good specificity that supports the strict paired-counterfactual gates of Tab.~\ref{tab:sota_failure_main_v4}.

\begin{table*}[t]
\centering
\scriptsize
\setlength{\tabcolsep}{4pt}
\renewcommand{\arraystretch}{1.05}
\resizebox{\textwidth}{!}{%
\begin{tabular}{l cc cc ccc cc}
\toprule
\textbf{Backbone} & \multicolumn{2}{c}{\textbf{S1$_{\text{P}}$ paired}} & \multicolumn{2}{c}{\textbf{S2$_{\text{C}}$}} & \multicolumn{3}{c}{\textbf{S3$_{\text{E}}$}} & \multicolumn{2}{c}{\textbf{S4$_{\text{A}}$}} \\
\cmidrule(lr){2-3} \cmidrule(lr){4-5} \cmidrule(lr){6-8} \cmidrule(lr){9-10}
 & swap-TPR & paired-good-spec. & TPR & FPR & J$_a$ & J$_b$ & spec. & TPR & FPR \\
\midrule
\multicolumn{10}{l}{\emph{TPR / specificity / FPR breakdown of paired-counterfactual metrics; see Table~\ref{tab:sota_failure_main_v4} for the J columns}} \\
\midrule
\multicolumn{10}{l}{\emph{G-Eval \cite{liu2023geval}}} \\
Opus 4.7 & 0.06 & 0.80 & 0.84 & 0.22 & 0.21 & 0.00 & 0.53 & 0.35 & 0.70 \\
Sonnet 4.6 & 0.09 & 0.91 & 0.59 & 0.06 & 0.29 & 0.00 & 0.17 & 0.51 & 0.91 \\
Sonnet 4.5 & 0.20 & 0.56 & 0.65 & 0.05 & 0.40 & 0.10 & 0.67 & 0.38 & 0.46 \\
Haiku 4.5 & 0.08 & 0.26 & 0.61 & 0.44 & 0.50 & 0.05 & 0.80 & 0.19 & 0.75 \\
Qwen3 32B & 0.00 & 0.99 & 0.01 & 0.00 & 0.46 & 0.00 & 0.99 & 0.17 & 0.01 \\
GPT-OSS 20B & 0.06 & 0.39 & 0.49 & 0.27 & 0.17 & 0.00 & 0.72 & 0.19 & 0.57 \\
GPT-OSS 120B & 0.01 & 0.81 & 0.19 & 0.07 & 0.37 & 0.04 & 0.84 & 0.30 & 0.23 \\
\midrule
\multicolumn{10}{l}{\emph{Scalar (MT-Bench-style) \cite{zheng2023judging}}} \\
Opus 4.7 & 0.02 & 0.70 & 0.83 & 0.16 & 0.22 & 0.00 & 0.50 & 0.42 & 0.68 \\
Sonnet 4.6 & 0.05 & 0.89 & 0.64 & 0.07 & 0.37 & 0.00 & 0.16 & 0.47 & 0.93 \\
Sonnet 4.5 & 0.26 & 0.64 & 0.56 & 0.08 & 0.45 & 0.02 & 0.81 & 0.39 & 0.38 \\
Haiku 4.5 & 0.08 & 0.12 & 0.44 & 0.11 & 0.43 & 0.00 & 0.25 & 0.27 & 0.87 \\
Qwen3 32B & 0.00 & 0.99 & 0.07 & 0.00 & 0.47 & 0.00 & 0.99 & 0.19 & 0.01 \\
GPT-OSS 20B & 0.09 & 0.38 & 0.22 & 0.33 & 0.32 & 0.00 & 0.84 & 0.16 & 0.55 \\
GPT-OSS 120B & 0.02 & 0.73 & 0.50 & 0.06 & 0.59 & 0.09 & 0.90 & 0.43 & 0.16 \\
\midrule
\multicolumn{10}{l}{\emph{Persona-blind}} \\
Opus 4.7 & 0.00 & 0.15 & 0.77 & 0.00 & 0.12 & 0.00 & 0.03 & 0.42 & 0.97 \\
Sonnet 4.6 & 0.05 & 0.87 & 0.68 & 0.13 & 0.20 & 0.00 & 0.44 & 0.36 & 0.69 \\
Sonnet 4.5 & 0.06 & 0.82 & 0.63 & 0.06 & 0.49 & 0.13 & 0.88 & 0.36 & 0.18 \\
Haiku 4.5 & 0.26 & 0.21 & 0.52 & 0.12 & 0.49 & 0.00 & 0.67 & 0.23 & 0.80 \\
Qwen3 32B & 0.00 & 0.99 & 0.09 & 0.00 & 0.43 & 0.00 & 0.99 & 0.18 & 0.01 \\
GPT-OSS 20B & 0.19 & 0.46 & 0.20 & 0.29 & 0.42 & 0.01 & 0.88 & 0.22 & 0.45 \\
GPT-OSS 120B & 0.08 & 0.76 & 0.41 & 0.03 & 0.55 & 0.03 & 0.84 & 0.47 & 0.19 \\
\midrule
\multicolumn{10}{l}{\emph{Persona-aware}} \\
Opus 4.7 & 0.05 & 0.61 & 0.84 & 0.07 & 0.28 & 0.09 & 0.62 & 0.38 & 0.45 \\
Sonnet 4.6 & 0.11 & 0.88 & 0.64 & 0.07 & 0.22 & 0.00 & 0.53 & 0.39 & 0.56 \\
Sonnet 4.5 & 0.32 & 0.63 & 0.48 & 0.03 & 0.30 & 0.08 & 0.78 & 0.34 & 0.34 \\
Haiku 4.5 & 0.37 & 0.23 & 0.63 & 0.14 & 0.58 & 0.22 & 0.83 & 0.28 & 0.76 \\
Qwen3 32B & 0.02 & 0.99 & 0.08 & 0.00 & 0.44 & 0.00 & 1.00 & 0.17 & 0.01 \\
GPT-OSS 20B & 0.22 & 0.59 & 0.29 & 0.15 & 0.32 & 0.01 & 0.87 & 0.21 & 0.35 \\
GPT-OSS 120B & 0.12 & 0.75 & 0.35 & 0.01 & 0.50 & 0.03 & 0.88 & 0.41 & 0.13 \\
\midrule
\multicolumn{10}{l}{\emph{\textbf{\JudgeName{}}}} \\
Opus 4.7 & 0.23 & 0.84 & 0.84 & 0.09 & 0.55 & 0.23 & 0.93 & 0.32 & 0.16 \\
Sonnet 4.6 & 0.26 & 0.90 & 0.85 & 0.04 & 0.51 & 0.08 & 0.88 & 0.52 & 0.20 \\
Sonnet 4.5 & 0.80 & 0.15 & 0.83 & 0.07 & 0.25 & 0.22 & 0.36 & 0.38 & 0.85 \\
Haiku 4.5 & 0.14 & 0.27 & 0.69 & 0.14 & 0.41 & 0.39 & 0.45 & 0.34 & 0.73 \\
Qwen3 32B & 0.02 & 0.96 & 0.18 & 0.01 & 0.51 & 0.01 & 0.98 & 0.23 & 0.04 \\
GPT-OSS 20B & 0.24 & 0.21 & 0.34 & 0.20 & 0.23 & 0.10 & 0.48 & 0.29 & 0.67 \\
GPT-OSS 120B & 0.09 & 0.39 & 0.47 & 0.01 & 0.32 & 0.20 & 0.49 & 0.48 & 0.52 \\
\bottomrule
\end{tabular}%
}
\caption{\textbf{Paired-counterfactual specificity audit.}  S1$_{\text{P}}$ is reported as the swap-side TPR and the paired GOOD-side specificity (judge correctly returns GOOD on the original-persona record sharing the same base id).  Persona-aware on Sonnet 4.6 retains a high swap-TPR but a low paired-good specificity; that is exactly the ``always cry BAD-PREF\_CONFLICT'' failure mode the paired metric rules out.  S2 / S4 columns show TPR on the BAD subset and FPR on the same-CSV GOOD subset.  S3 columns show J$_a$ (product evidence), J$_b$ (persona-history grounding), and the paired specificity rate (no E-family alarm on GOOD).  These quantities are the per-rubric inputs to the harmonic in Tab.~\ref{tab:sota_failure_main_v4}.}
\label{tab:sota_failure_specificity_v4}
\end{table*}

\FloatBarrier

\end{document}